\documentclass[runningheads]{llncs}

\usepackage{eccv}

\usepackage{eccvabbrv}

\usepackage{graphicx}
\usepackage{amsmath}
\usepackage{amssymb}
\usepackage{booktabs,multirow}
\usepackage{enumitem}
\usepackage{tabularx}
\usepackage{rotating}
\usepackage{tikz}

\usepackage[dvipsnames]{xcolor}

\usepackage{colortbl}
\usepackage{makecell}

\definecolor{myblue}{HTML}{1086ed}

\newcommand{\bluecircled}[1]{%
  \begin{tikzpicture}[baseline=(char.base)]
    \node[shape=circle, draw=myblue, fill=myblue, text=white, inner sep=1pt] (char) {#1};
  \end{tikzpicture}%
}

\definecolor{OliveGreen}{rgb}{0,0.6,0}
\definecolor{SoftRed}{rgb}{1,0.2,0.2}
\definecolor{PastePink}{RGB}{253, 223, 236}

\newcommand{\rotateT}[1]{\begin{turn}{90}\footnotesize{#1}\end{turn}}
\newcommand{\tsx}[1]{\scriptsize{#1}}
\newcommand{\tsb}[1]{\scriptsize{\textbf{#1}}}

\newcommand{\circledc}{%
  \mathbin{\tikz[baseline=(char.base)]{%
    \node[shape=circle,draw,inner sep=1pt] (char) {$c$};}}
}

\usepackage{hyperref}

\usepackage{orcidlink}

\begin{document}

\title{OLAF: A Plug-and-Play Framework for Enhanced Multi-object Multi-part Scene Parsing} 

\titlerunning{OLAF: Framework For Scene Parsing at Part-Level}

\author{Pranav Gupta\inst{1}\and
Rishubh Singh\inst{2,3} \and
Pradeep Shenoy\inst{3} \and
Ravi Kiran Sarvadevabhatla\inst{1}}

\authorrunning{P.~Gupta et al.}

\institute{ IIIT Hyderabad \and Swiss Federal Institute of Technology (EPFL) \and
Google Research \\
\email{\{pranav.gu@research.,ravi.kiran@\}iiit.ac.in , 
rishubh.singh@epfl.ch,
shenoypradeep@google.com}}

\maketitle

\begin{abstract}
  Multi-object multi-part scene segmentation is a challenging task whose complexity scales exponentially with part granularity and number of scene objects. To address the task, we propose a plug-and-play approach termed OLAF. First, we augment the input (RGB) with channels containing object-based structural cues (fg/bg mask, boundary edge mask). We propose a weight adaptation technique which enables regular (RGB) pre-trained models to process the augmented (5-channel) input in a stable manner during optimization. In addition, we introduce an encoder module termed LDF to provide low-level dense feature guidance. This assists segmentation, particularly for smaller parts. OLAF enables significant mIoU gains of $\mathbf{3.3}$ (Pascal-Parts-58), $\mathbf{3.5}$ (Pascal-Parts-108) over the SOTA model. On the most challenging variant (Pascal-Parts-201), the gain is $\mathbf{4.0}$. Experimentally, we show that OLAF's broad applicability enables gains across  multiple architectures (CNN, U-Net, Transformer) and datasets.  The code is available at \href{https://olafseg.github.io/}{olafseg.github.io}
\end{abstract}

\section{Introduction}
\label{sec:intro}

\begin{figure}[t]
  \centering
   \includegraphics[width=\linewidth, height=6cm]{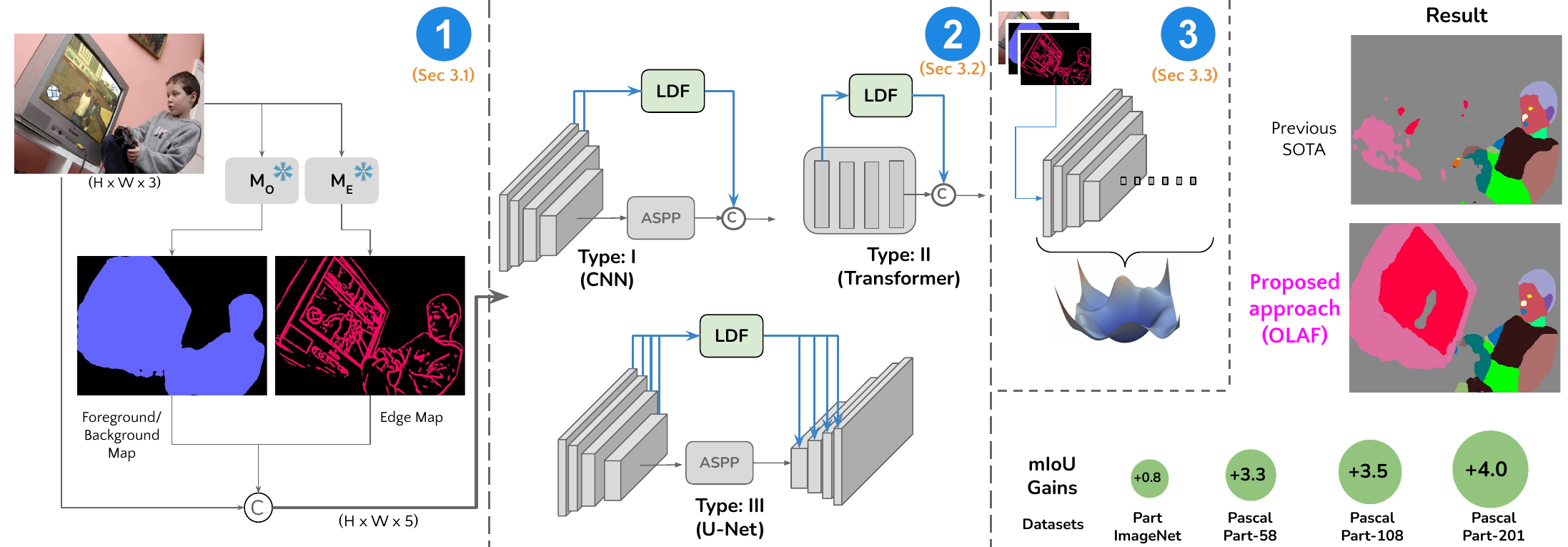}
      \caption{The recipe for OLAF, our plug-and-play framework for enhanced multi-object multi-part scene parsing: \protect\bluecircled{1} Augment RGB input with object-based channels (fg/bg, boundary edges) obtained from frozen pre-trained models ($M_O,M_E$) \protect\bluecircled{2} Use  Low-level Dense Feature guidance from segmentation encoder (LDF, shaded green) \protect\bluecircled{3} Employ targeted weight adaptation for stable optimization with augmented input. We show that following this recipe leads to significant gains (up to $\mathbf{4.0}$ mIoU) across multiple architectures and across multiple challenging datasets.}  
    \label{fig:intro}
\end{figure}

\begin{figure}[t]
  \centering
   \includegraphics[width=\linewidth, height=6cm]{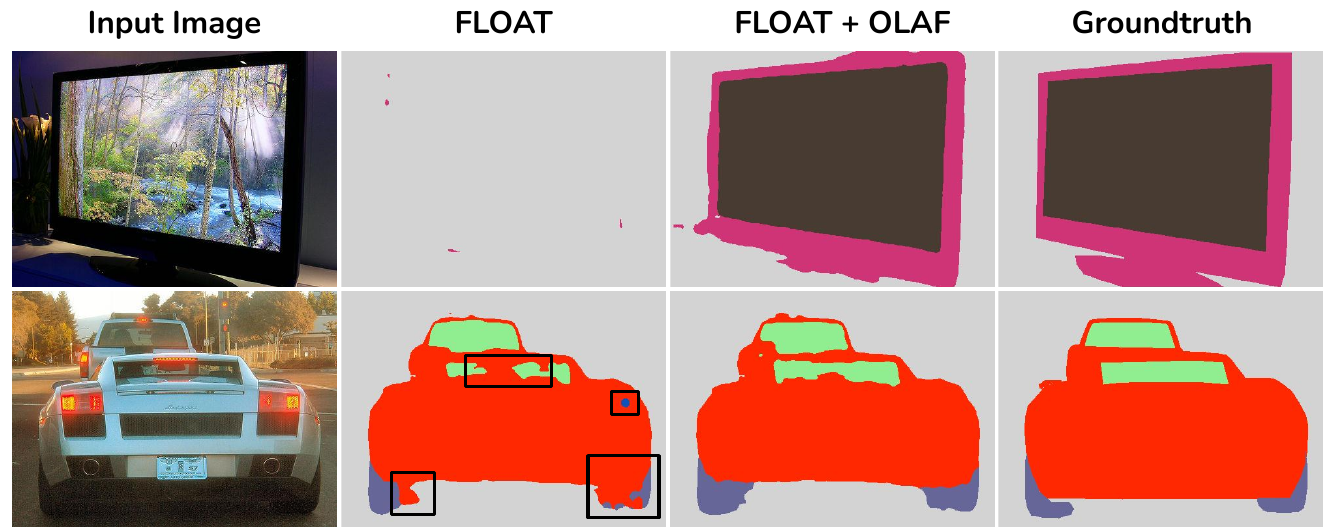}
      \caption{The segmentation results for state-of-the-art approach FLOAT~\cite{float} and its \textcolor{orange}{limitations} can be seen in the second column. In the first row, FLOAT completely fails to identify \textit{TV Frame} and \textit{TV Screen}. In the second row, FLOAT fails to capture the edge partition between \textit{Car-Body}, \textit{Car-Tire} and also between \textit{Car-Body}, \textit{Car-Window}. The third column shows results by incorporating our plug-and-play approach OLAF into FLOAT, leading to significantly improved object and part segmentation results.}  
    \label{fig:float_limitations}
\end{figure}

Multi-object multi-part segmentation is a challenging task that involves simultaneously segmenting multiple objects in an image while also segmenting their individual parts. The task goes beyond conventional object segmentation~\cite{icnet, objectseg_instance,objectseg_interactive,objectseg_large,objectseg_rvos,objectseg_unsupervised, segnet, pspnet} and aims to enable multi-granular scene understanding. The availability of granular semantic detail is crucial for applications in robotics~\cite{robotics_1,robotics_2}, visual question answering~\cite{vqa}, object interaction and modeling~\cite{shape_composite, shapeglot} and other domains~\cite{intro_1, intro_2, intro_3, intro_4} where understanding the scene in terms of objects and their constituent parts is crucial.

Related approaches primarily address simpler variants such as single-object part parsing~\cite{ppp_human,ppp_semantic,vehile_1,vehicle_2,animal_semantic,animal_joint} or part parsing for objects with fewer or visibly larger parts~\cite{compositor}. Some recent methods~\cite{float,gmnet,bsanet, co-rank} have been developed to specifically tackle the more complex task of multi-object multi-part parsing. However, these suffer from three significant limitations:

\textcolor{orange}{\textit{Limitation 1:}} Foreground (union of object regions) is often incorrectly segmented, impacting the constituent part segmentation (\Cref{fig:float_limitations}, first row). \textcolor{orange}{\textit{Limitation 2:}} Crucial boundary details between objects and parts are not captured accurately (\Cref{fig:float_limitations}, second row). \textcolor{orange}{\textit{Limitation 3}:} Small and thin parts especially fail to be segmented (\Cref{fig:ldf_qualitative}, \Cref{fig:qualitative_fig}). 

To address \textcolor{orange}{\textit{Limitation 1 \& 2}}, we first obtain a plausible boundary edge mask using a pre-trained network. We use another pre-trained network to obtain 
preliminary object segmentation and combine the object label channels to obtain a binary foreground mask. These masks are included as additional channels to constitute the 5-channel (3 RGB + 2 masks) input to the reference segmentation network (see \bluecircled{1} in \Cref{fig:intro}). The masks provide an inductive bias and guide the model to focus on relevant parts from the onset of training. We also propose a weight adaptation technique that enables the pre-trained segmentation encoder to process the new 5-channel input without destabilizing optimization (\bluecircled{3} in \Cref{fig:intro}).

To address \textcolor{orange}{\textit{Limitation 3}} (i.e. small and thin parts), we introduce an encoder module termed Low-level Dense Feature (LDF)  - see \bluecircled{2} in \Cref{fig:intro}. This module, in conjunction with augmented input representation, provides low-level dense feature guidance enabling better segmentation, especially for small/thin parts. 

To summarize our contributions:
\begin{itemize}[leftmargin=*]
    \item \textbf{Input Augmentation}: We introduce an augmented 5-channel input representation with auxiliary channels containing object and boundary cues.
    \item \textbf{Weight Adaptation Technique}: We introduce a targeted weight-adaptation training procedure that ensures stable optimization of pre-trained backbones on the augmented (5-channel) input.
    \item \textbf{Low-Level Dense Feature Guidance (LDF)}: We propose a generic encoder module called LDF which provides valuable low-level dense feature guidance, especially for small part segmentation. 
    \item \textbf{Performance Boost}: OLAF achieves significant mIoU improvements, surpassing state-of-the-art by $\mathbf{3.3}$ on Pascal-Parts-58, $\mathbf{3.5}$ on Pascal-Parts-108, and $\mathbf{4.0}$ on Pascal-Parts-201.
    \item \textbf{Generalizability}: We show that OLAF enhances performance across multiple representative segmentation families (CNN, U-Net, Transformer) and multiple datasets (Pascal-Parts 58/108/201 and PartImageNet), suggesting broad applicability as a plug-and-play framework. 

\end{itemize}

\section{Related Work}
\label{sec:related_work}

\textbf{Single-Object Multi-Part Segmentation} considers a single object and its constituent parts. Most works explore segmentation of non-rigid object categories such as person~\cite{ppp_distillation,ppp_human,ppp_hybrid,ppp_instance,ppp_look,ppp_mutual,ppp_semantic,ppp_semantic_,ppp_weakly}, some animals~\cite{animal_joint, animal_semantic,animal_semantic_} and rigid object categories (e.g. vehicles~\cite{vehicle_2,vehicle_embedding,vehicle_learning,vehile_1}). Some works have also examined open-world part segmentation~\cite{open_world,wei2023ov}. However, the single object condition is restrictive and not representative of general scenes which contain multiple objects from distinct categories and associated occlusions.

\textbf{Multi-Object Multi-Part Segmentation} has recently got increased attention due to its complexity and importance in downstream applications. There have been multiple different approaches to tackle this hard problem \cite{float, bsanet, gmnet, co-rank}. The initial work of Zhao et al.~\cite{bsanet} and the works of Micheli et al.~\cite{gmnet, edge_gmnet} leverage object and boundary awareness through auxiliary tasks and model design changes. The current state of the art approach (FLOAT~\cite{float}) employs label-space factorization to reduce the number of output heads. Typically, existing approaches do not attempt segmentation of very small/thin parts although annotations are available~\cite{float}. Beyond the popular datasets (Pascal-Parts-58), our approach enables improved performance for the harder variants (Pascal-Parts-108, Pascal-Parts-201). 

\textbf{Object level guidance} and \textbf{Boundary/edge awareness} is typically present as an auxiliary task or in terms of guidance from an object network's features~\cite{bsanet,gmnet, edge_gmnet,factseg,auxilary_dhfnet,auxilary_foreground,auxilary_unsupervised,open_world}. For multi-object multi-part segmentation, Zhao et al.~\cite{bsanet} add an auxiliary task of predicting object segments from the learned part segmentation representation. Michieli et al.~\cite{gmnet} use the features from the last layer of an object segmentation network as guidance to the part segmentation decoder. In contrast, our work OLAF adds object segmentation and edge information directly as additional channels to the input which is observed to be more beneficial. 


\textbf{Low-level Feature Guidance} has been used in previous works~\cite{ldf_hypercolumns,ldf_bisenet, ldf_full, ldf_attention} to enhance the performance of segmentation by leveraging low-level visual (spatial) cues. While some works incorporate skip-connections~\cite{aspp, ldf_unet}, others utilize downsampling strategies along with skip-connections to obtain sufficient receptive field for context capturing~\cite{ldf_bisenet, ldf_full}. While this strategy generally works well, it may not be suitable for tasks such as part segmentation because the information in low-level features is too coarse. In particular, downsampling compromises details of small or thin parts. By contrast, our approach strives to efficiently exploit low-level cues in the most beneficial manner while also capturing the semantics of small or thin parts.

\section{Methodology of OLAF}
\label{sec:methodology_of_olaf}

\begin{figure}[!t]
  \centering
    \includegraphics[ width=\linewidth, height=7.2cm]{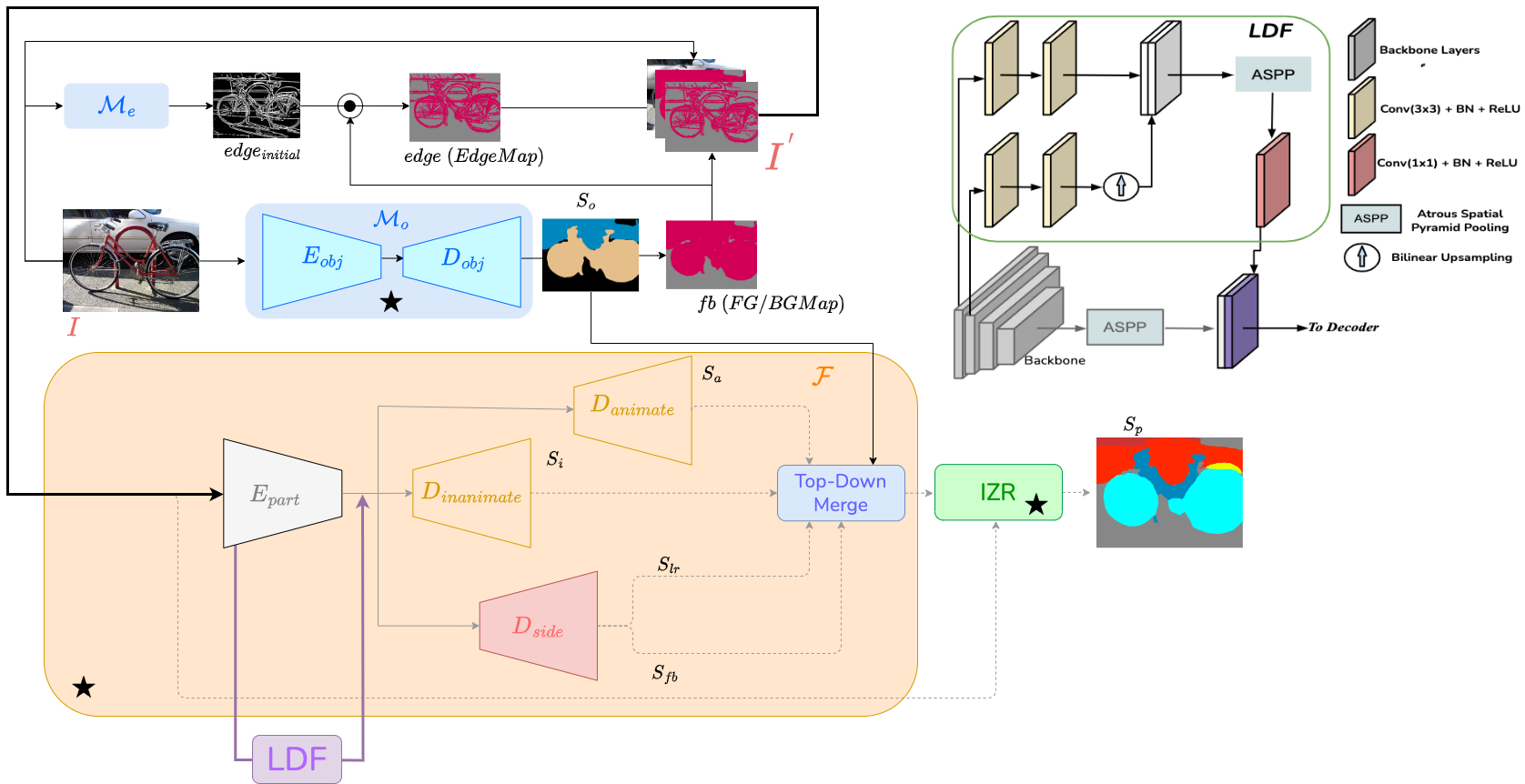}
    \caption{Illustration of OLAF's architectural integration with FLOAT~\cite{float}(Sec.~\ref{sec:methodology_of_olaf}). FLOAT's components are tagged with $\bigstar$. The object masks from output $S_o$ of object segmentation network $\mathcal{M}_o$ are merged to obtain the foreground map $fg$. The output of edge generation network $\mathcal{M}_e$ is thresholded and filtered using $fg$ to obtain edge map $edge$. The obtained maps are stacked with input image $I$ to obtain the $5$-channel input $I^{'}$ for the part segmentation network $\mathcal{F}$. The interface for LDF (Sec.~\ref{sec:lldfe}) with encoder $E_{part}$ and its architecture (top right) are also shown. A similar integration of OLAF also exists for U-Net style and Transformer style  architectures.}
    \label{fig:ldf_arc}
\end{figure}

OLAF introduces structural modifications at two key aspects of the standard segmentation pipeline (see Fig.~\ref{fig:ldf_arc}). The first change occurs at input stage, where we enrich RGB input with auxiliary channels containing object-based structural cues, including foreground/background masks and boundary edge masks. The second structural adjustment takes place within the encoder. We introduce a dedicated module termed LDF which provides low-level dense feature guidance to benefit the segmentation of smaller parts. In addition to these structural enhancements, we introduce a weight adaptation technique which ensures that pre-trained RGB (3-channel) backbones seamlessly adapt to augmented (5-channel) input during optimization. In this section, we provide a detailed explanation of these crucial components.

\subsection{Foreground and Edge masks as boundary cues}
\label{sec:fgbgedge}

The inclusion of foreground (union of object regions) guides the part predictions to lie within object boundaries. To obtain the corresponding channel mask, we use a state-of-the-art object segmentation network~\cite{aspp} to obtain object predictions. These are transformed into a binary foreground/background mask.

\[
fb(x, y) = \begin{cases}
   1, & \text{if } P(x, y) \in C \text{ and } P(x, y) \neq 0 \\
   0, & \text{otherwise}
\end{cases}
\]
where $C$ denotes the predicted object class and pixel location is $(x,y)$. 

Edges play a crucial role in delineating boundaries between different objects, parts and recognizing intricate details within the scene. To obtain a collection of such edges, we use the Holistically-Nested Edge Detection (HED)~\cite{hed} model. The raw output \( edge_{\text{initial}} \) from HED lies in the range $[0,1]$ and contains edges from background as well. To filter out these background edges, we employ the fg/bg mask as follows:
\[
\textit{edge} = \mathbb{I}[\textit{edge}_{\textit{initial}} > 0 ] \odot \textit{fb}
\]
where \textit{edge} represents foreground edges and \( \odot \) denotes element-wise multiplication and $\mathbb{I}$ is the indicator function. Note that $edge$ mask is binary, i.e. pixel location where there is an $edge$ has pixel value $1$ otherwise $0$.

To prepare the input, we append the foreground and edge masks as separate channels to the input image $I$. More precisely, the original input image $I$ with dimensions $H \times W \times 3$ is transformed into a modified input  $I'$ with dimensions $H \times W \times 5$ (see \bluecircled{1} in \Cref{fig:intro}).

Conventional segmentation approaches typically include auxiliary tasks to learn foreground/background~\cite{factseg} and edges during training~\cite{bsanet}. However, directly including foreground/background and edges as part of the input can be thought of as a structural inductive bias for the task. These masks provide strong boundary cues throughout the optimization process. In addition, they eliminate the issue of irregular gradient flow arising from ad-hoc scaling of task-related losses~\cite{dery2021auxiliary} in existing (RGB input only) approaches. 

\begin{figure*}[!t]
  \centering
    \includegraphics[ width=\linewidth,height=6cm]{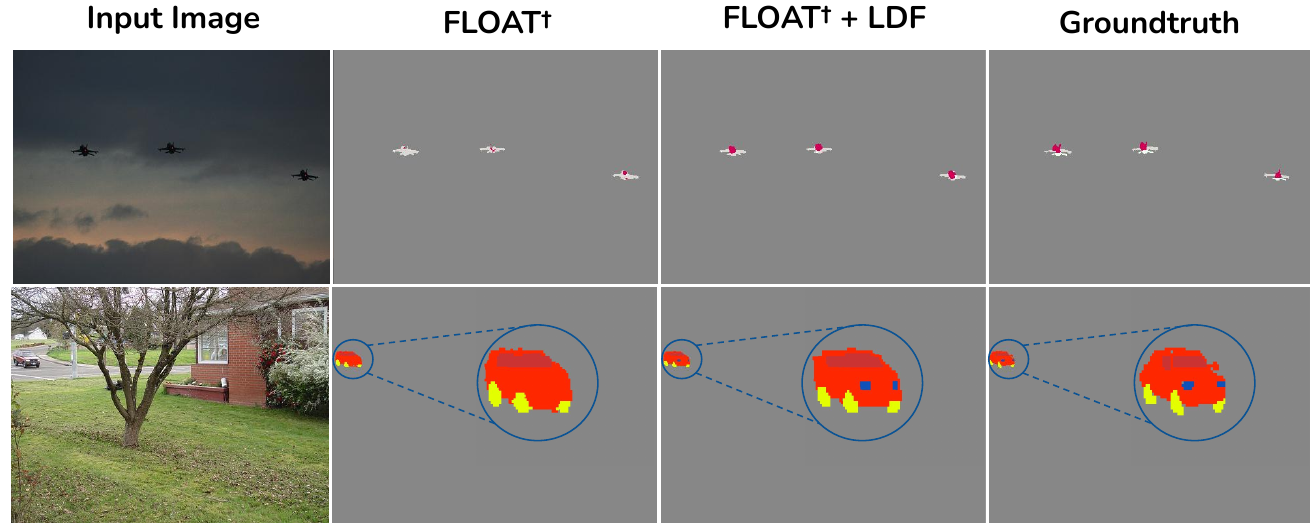}
    \caption{LDF (\Cref{sec:lldfe}) consistently improves the performance of small/thin parts. As shown in Row I, FLOAT~\cite{float} with LDF successfully predicts \textcolor{red}{Aeroplane-Body} while FLOAT fails to do so adequately. Similar results can be seen in Row II, where LDF successfully predicts \textcolor{blue}{Car-Light} which FLOAT completely misses.}
    \label{fig:ldf_qualitative}
\end{figure*}

\subsection{Low-level Dense Feature Extractor (LDF)}
\label{sec:lldfe}

Typically, encoders in segmentation architectures~\cite{aspp,pspnet,ldf_bisenet,segnet,typical_lednet} process image representations in a downsampled feature space (often $1/8^{th}$ or $1/16^{th}$ size of input image). However, aggressive downsampling and intermediate pooling operations lead to the loss of fine details and small entity instances in an image. This effect is more pronounced for parts since they cover relatively smaller pixel areas compared to scene objects.

To address this issue, some architectures either employ skip-connections~\cite{aspp} or directly concatenate output from early backbone blocks with the decoder~\cite{densenet}. However, despite these methods allowing flow of low level features to the decoder, they still fail to segment small/thin parts. This is because information from  early stages of the encoder is too coarse and lacks contextual information for accurate segmentation of such parts.

To overcome these challenges, we introduce the Low-level Dense Feature Extractor (LDF) module. LDF leverages early blocks of the backbone network, where low-level information associated with small/thin parts are more prominent. To capture dense features of these small/thin parts, LDF includes (a) convolutional layers to enhance the features extracted from the initial stages of the backbone (b) an upsampling layer to maintain consistent feature map size (c) Atrous Spatial Pyramid Pooling (ASPP)~\cite{aspp} to capture contextual information at multiple scales (see \Cref{fig:ldf_arc}). This enables the model to extract dense low-level features at various spatial resolutions and consider context at different scales, including context relevant to small/thin parts. LDF can be formalized as:
\begin{align*}
    feat(x_{1}, x_{2}) &= Conv_{3\times3}(x_{1}) \circledc UP(Conv_{3\times3}(x_{2})) \\
    LDF(x_{1}, x_{2}) &= Conv_{1\times1}(ASPP(feat(x_{1},x_{2})))
\end{align*}
where $x_{1}$ and $x_{2}$ are the features from the first and second block of the backbone, $Conv_{3\times3}$ represents a $3\times3$ convolution with \textit{stride = 1} and \textit{padding = 1} to avoid dimension reduction, $Conv_{1\times1}$ represents $1\times1$ convolution. $ASPP(.)$ represents Atrous Spatial Pyramid Pooling\cite{aspp} and $\circledc$ represents the concatenation operation. $UP(.)$ represents applying an Upsample Convolution layer which applies an upsampling step with a scale factor, followed by a $Conv_{1\times1}$, batch normalization, and a ReLU activation.

Similar to input channel augmentation (\Cref{sec:fgbgedge}), LDF can also be viewed as imposing a structural inductive bias for smaller and thin parts. LDF enables noticeable improvements in performance for such parts (see \Cref{fig:ldf_qualitative},\Cref{tab:ablation_table}).

\subsection{Weight Adaptation}
\label{subsec:weight-adaptation}

Standard pre-trained segmentation models support 3-channel RGB images as input. To efficiently adapt existing models for our modified 5 channel input, we employ a simple but effective technique. As the initial step, for filters in the first layer, we average their weights across the channel dimension. The result is used to initialize the weights of the filters corresponding to the two newly included input channels (i.e. fg/bg and edges). To ensure stable training, the optimization contains a warm-up phase consisting of $n_{warm}$ epochs. Compared to alternative schemes (\Cref{sec:weightadaptationschemes}), this approach prevents weight updates that lead to instability or divergence during the initial stages of training

\section{Experiments}

\subsection{Datasets}


\noindent \textbf{Pascal-Part Dataset Variants:} The Pascal-Parts dataset contains 4,998 training samples and 5,105 samples for testing~\cite{voc}. The dataset contains part level annotation for the 20 Pascal VOC2010 semantic object classes (including the background class). We follow Singh et al.~\cite{float} and evaluate OLAF on three variants of Pascal-Part, which are Pascal-Part-58, Pascal-Part-108 and Pascal-Part-201 in increasing order of complexity. Pascal-Part-58 contains 58 part classes, mostly focusing on larger parts of objects such as heads, torsos, legs for animals, and components such as body, wheels for non-living objects. Pascal-Part-108 is more challenging, featuring 108 part classes. It includes smaller parts such as eyes, necks, feet for animals, and roofs, doors for non-living objects. Pascal-Part-201~\cite{float} is the most challenging version among the Pascal-Part dataset variants. It includes 201 part classes and introduces additional part attributes `left', `right', `front', `back', `upper', `lower', along with minor parts (e.g. `eyebrows'), which are not present in the other variants (58/108). 

\noindent \textbf{PartImageNet~\cite{partimagenet}:} This is a large-scale dataset containing 24,095 images. 
We follow the official training/validation split and evaluate performance on the publicly available validation set.

\subsection{Evaluation Metrics}

We use the standard mean Intersection over Union (mIoU) score as a performance measure. 
mIoU tends to be influenced more by the contributions of ``larger'' part instances. Therefore, we also report   sqIoU~\cite{float}, designed specifically for fairer balance between small and larger parts. 
For comparison with existing works~\cite{compositor}, we report average pixel accuracy as a metric for PartImageNet~\cite{partimagenet}.

\subsection{Training Details}

\begin{table}[!t]
  \centering   
  \caption{Pascal-Part-201 results (mIoU/mAvg - top table and sqIoU/sqAvg - bottom table). ``+ O'': augmented with OLAF, ``\textdagger'': with ViT-H backbone.
  }
  \resizebox{\textwidth}{!}{  
  \begin{tabular}{@{}l|ccccccccccccccccccccc|cc@{}}
    \toprule
    {Model} &  \rotateT{bgr} & \rotateT{aero} & \rotateT{bike} & \rotateT{bird} & \rotateT{boat} & \rotateT{bottle} & \rotateT{bus} & \rotateT{car} & \rotateT{cat} & \rotateT{chair} & \rotateT{cow} & \rotateT{table} & \rotateT{dog} & \rotateT{horse} & \rotateT{mbike} & \rotateT{person} & \rotateT{plant} & \rotateT{sheep} & \rotateT{sofa} & \rotateT{train} & \rotateT{TV} & \tsx{\textbf{mIOU}} & \tsx{\textbf{mAvg}} \\
    \midrule
    {DeepLabv3~\cite{aspp}} & \tsx{91.0} & \tsx{31.6} & \tsb{47.7} & \tsx{24.3} & \tsx{56.7} & \tsx{46.4} & \tsx{31.0} & \tsx{36.7} & \tsx{24.2} & \tsx{35.6} & \tsx{17.5} & \tsx{38.6} & \tsx{27.3} & \tsx{20.7} & \tsx{38.0} & \tsx{26.9} & \tsx{50.8} & \tsx{13.3} & \tsx{42.1} & \tsx{14.7} & \tsx{57.6} & \tsx{26.3} & \tsx{36.8} \\
     {\textbf{Deeplabv3 + O}} & \tsb{93.6} & \tsb{35.3} & \tsx{42.9} & \tsb{27.2} & \tsb{65.5} & \tsb{51.0} & \tsb{33.3} & \tsb{38.4} & \tsb{25.8} & \tsb{42.1} & \tsb{20.5} & \tsb{46.9} & \tsb{29.4} & \tsb{22.6} & \tsb{39.3} & \tsb{28.9} & \tsb{56.1} & \tsb{15.5} & \tsb{46.8} & \tsb{17.2} & \tsb{65.2} & \tsb{28.6} & \tsb{40.2}\\
    \midrule
    \footnotesize{GMNet\cite{gmnet}}  & \tsx{90.8} & \tsx{26.6} & \tsx{33.1} & \tsx{21.2} & \tsx{55.0} & \tsx{43.5} & \tsx{24.6} & \tsx{27.5} & \tsx{21.7} & \tsx{35.5} & \tsx{15.1} & \tsx{40.3} & \tsx{25.0} & \tsx{17.5} & \tsx{31.9} & \tsx{21.9} & \tsx{44.2} & \tsx{11.9} & \tsx{43.3} & \tsx{14.0} & \tsx{53.2} & \tsx{22.5} & \tsx{33.2}\\
    \footnotesize{\textbf{GMNet + O}} & \tsb{93.1} & \tsb{28.5} & \tsb{35.9} & \tsb{23.5} & \tsb{64.5} & \tsb{47.0} & \tsb{26.7} & \tsb{29.5} & \tsb{26.4} & \tsb{41.9} & \tsb{19.0} & \tsb{47.9} & \tsb{29.1} & \tsb{20.3} & \tsb{34.0} & \tsb{26.3} & \tsb{49.8} & \tsb{16.9} & \tsb{54.8} & \tsb{15.9} & \tsb{60.5} & \tsb{26.0} & \tsb{37.7}  \\
    \midrule
    \footnotesize{BSANet\cite{bsanet}} & \tsx{91.2} & \tsx{34.6} & \tsx{41.7} & \tsx{27.9} & \tsx{61.2} & \tsx{51.7} & \tsx{34.1} & \tsx{38.1} & \tsx{26.1} & \tsx{35.4} & \tsx{24.0} & \tsx{43.6} & \tsx{28.4} & \tsx{23.0} & \tsx{37.4} & \tsx{27.7} & \tsx{54.7} & \tsx{14.3} & \tsx{40.4} & \tsx{17.8} & \tsx{59.4} & \tsx{28.5} & \tsx{38.7}\\
    \footnotesize{\textbf{BSANet + O}} & \tsb{92.9} & \tsb{35.9} & \tsb{43.1} & \tsb{29.9} & \tsb{69.8} & \tsb{53.3} & \tsb{35.8} & \tsb{39.8} & \tsb{28.9} & \tsb{39.8} & \tsb{26.6} & \tsb{49.2} & \tsb{31.2} & \tsb{24.9} & \tsb{38.9} & \tsb{31.1} & \tsb{59.5} & \tsb{18.7} & \tsb{49.9} & \tsb{19.0} & \tsb{65.8} & \tsb{31.1} & \tsb{42.1}\\
    \midrule
    \footnotesize{FLOAT\cite{float}} & \tsx{92.5} & \tsx{36.7} & \tsx{42.6} & \tsx{34.4} & \tsx{75.3} & \tsx{51.4} & \tsx{35.8} & \tsx{42.0} & \tsx{37.8} & \tsx{59.6} & \tsx{35.5} & \tsx{58.2} & \tsx{41.0} & \tsx{34.0} & \tsx{40.2} & \tsx{40.8} & \tsx{52.2} & \tsx{28.5} & \tsx{69.0} & \tsx{15.1} & \tsx{56.1} & \tsx{36.9} & \tsx{46.6}\\
    \footnotesize{\textbf{FLOAT + O}} & \tsb{93.3} & \tsb{38.7} & \tsb{45.1} & \tsb{37.4} & \tsb{76.7} & \tsb{54.1} & \tsb{38.7} & \tsb{46.2} & \tsb{41.4} & \tsb{59.7} & \tsb{38.9} & \tsb{58.3} & \tsb{43.1} & \tsb{35.2} & \tsb{42.3} & \tsb{44.6} & \tsb{61.0} & \tsb{31.5} & \tsb{69.1} & \tsb{16.4} & \tsb{69.3} & \tsb{39.8} & \tsb{49.6} \\
    \midrule
    \footnotesize{FLOAT\textsuperscript{\textdagger}} & \tsx{93.2}	& \tsx{36.9}	& \tsx{44.5}	& \tsx{36.8}	& \tsx{76.4}	& \tsx{32.1}	& \tsx{36.3}	& \tsx{44.2}	& \tsx{39.3}	& \tsx{59.3}	& \tsx{36.8}	& \tsx{58.1}	& \tsx{42.7}	& \tsx{34.1}	& \tsx{41.2}	& \tsx{43.2}	& \tsx{56.7}	& \tsx{30.2}	& \tsx{68.9}	& \tsx{15.2}	& \tsx{59.4} & \tsx{37.7}	& \tsx{46.9}\\
\footnotesize{\textbf{FLOAT\textsuperscript{\textdagger} + O}} & \tsb{93.7}	& \tsb{39.9}	& \tsb{45.6}	& \tsb{38.3}	& \tsb{77.1}	& \tsb{55.3}	& \tsb{39.3}	& \tsb{47.9}	& \tsb{42.2}	& \tsb{60.7}	& \tsb{39.4}	& \tsb{59.6}	& \tsb{44.2}	& \tsb{37.3}	& \tsb{44.2}	& \tsb{47.1}	& \tsb{63.5}	& \tsb{32.2}	& \tsb{69.6}	& \tsb{16.9}	& \tsb{70.9}	& \tsb{40.9}	& \tsb{50.7}\\
    \bottomrule
    \toprule
     &  \rotateT{bgr} & \rotateT{aero} & \rotateT{bike} & \rotateT{bird} & \rotateT{boat} & \rotateT{bottle} & \rotateT{bus} & \rotateT{car} & \rotateT{cat} & \rotateT{chair} & \rotateT{cow} & \rotateT{table} & \rotateT{dog} & \rotateT{horse} & \rotateT{mbike} & \rotateT{person} & \rotateT{plant} & \rotateT{sheep} & \rotateT{sofa} & \rotateT{train} & \rotateT{TV} & \tsx{\textbf{sqIOU}} & \tsx{\textbf{sqAvg}} \\
    \midrule
    \footnotesize{Deeplabv3~\cite{aspp}} & \tsx{89.6} & \tsx{28.9} & \tsb{39.3} & \tsx{17.1} & \tsx{57.4} & \tsx{32.3} & \tsx{27.1} & \tsx{26.0} & \tsx{20.5} & \tsx{39.8} & \tsx{14.8} & \tsx{34.7} & \tsx{22.7} & \tsx{17.2} & \tsx{31.5} & \tsx{19.2} & \tsx{34.9} & \tsx{10.8} & \tsx{52.6} & \tsx{14.4} & \tsx{53.8} & \tsx{21.5} & \tsx{32.6} \\
    \footnotesize{\textbf{Deeplabv3 + O}} & \tsb{93.2} & \tsb{33.4} & \tsx{36.5} & \tsb{21.0} & \tsb{67.2} & \tsb{38.0} & \tsb{30.0} & \tsb{28.8} & \tsb{23.2} & \tsb{47.3} & \tsb{18.8} & \tsb{44.1} & \tsb{25.8} & \tsb{19.9} & \tsb{33.5} & \tsb{22.1} & \tsb{41.2} & \tsb{13.7} & \tsb{58.3} & \tsb{17.9} & \tsb{62.4} & \tsb{24.8} & \tsb{37.0}\\
    \midrule
    \footnotesize{GMNet\cite{gmnet}}  & \tsx{89.4} & \tsx{20.7} & \tsx{23.5} & \tsx{12.6} & \tsx{53.1} & \tsx{25.8} & \tsx{19.3} & \tsx{17.2} & \tsx{18.1} & \tsx{38.2} & \tsx{11.2} & \tsx{35.2} & \tsx{15.9} & \tsx{14.2} & \tsx{25.4} & \tsx{13.8} & \tsx{26.9} & \tsx{8.5} & \tsx{52.0} & \tsx{13.8} & \tsx{46.9} & \tsx{16.9} & \tsx{27.7}\\
    \footnotesize{\textbf{GMNet + O}} & \tsb{91.4} & \tsb{23.8} & \tsb{27.3} & \tsb{16.5} & \tsb{63.6} & \tsb{30.2} & \tsb{23.7} & \tsb{20.9} & \tsb{22.7} & \tsb{45.7} & \tsb{15.8} & \tsb{43.6} & \tsb{23.0} & \tsb{17.4} & \tsb{28.5} & \tsb{19.1} & \tsb{33.5} & \tsb{14.9} & \tsb{64.5} & \tsb{17.3} & \tsb{56.0} & \tsb{21.4} & \tsb{33.3}  \\
    \midrule
    \footnotesize{BSANet\cite{bsanet}}  & \tsx{89.9} & \tsx{30.7} & \tsx{33.5} & \tsx{18.6} & \tsx{60.2} & \tsx{31.2} & \tsx{29.2} & \tsx{26.4} & \tsx{21.2} & \tsx{37.8} & \tsx{17.5} & \tsx{38.0} & \tsx{22.3} & \tsx{17.8} & \tsx{31.2} & \tsx{18.2} & \tsx{33.6} & \tsx{10.8} & \tsx{47.2} & \tsx{17.5} & \tsx{55.4} & \tsx{22.1} & \tsx{32.8}\\
    \footnotesize{\textbf{BSANet + O}} & \tsb{91.4} & \tsb{32.9} & \tsb{35.7} & \tsb{21.6} & \tsb{69.8} & \tsb{34.2} & \tsb{31.8} & \tsb{29.1} & \tsb{25.0} & \tsb{43.2} & \tsb{21.0} & \tsb{44.6} & \tsb{26.1} & \tsb{20.7} & \tsb{33.3} & \tsb{22.5} & \tsb{39.3} & \tsb{16.2} & \tsb{57.7} & \tsb{19.6} & \tsb{62.8} & \tsb{25.7} & \tsb{37.1}  \\
    \midrule
    \footnotesize{FLOAT\cite{float}} & \tsx{90.8} & \tsx{32.5} & \tsx{35.8} & \tsx{24.5} & \tsx{63.9} & \tsx{36.1} & \tsx{30.4} & \tsx{29.9} & \tsx{33.0} & \tsx{50.8} & \tsx{28.1} & \tsx{47.6} & \tsx{35.6} & \tsx{26.1} & \tsx{33.6} & \tsx{29.9} & \tsx{34.5} & \tsx{20.6} & \tsx{69.0} & \tsx{13.6} & \tsx{56.8} & \tsx{29.5} & \tsx{39.2} \\
    \footnotesize{\textbf{FLOAT + O}} & 
\tsb{91.5} & 
\tsb{34.9} & 
\tsb{39.0} & 
\tsb{27.2} & 
\tsb{65.1} & 
\tsb{39.1} & 
\tsb{34.6} & 
\tsb{34.0} & 
\tsb{36.3} & 
\tsb{50.9} & 
\tsb{32.5} & 
\tsb{47.7} & 
\tsb{37.7} & 
\tsb{30.0} & 
\tsb{35.9} & 
\tsb{33.1} & 
\tsb{38.7} & 
\tsb{22.9} & 
\tsb{69.2} & 
\tsb{13.8} & 
\tsb{65.2} & 
\tsb{32.5} & 
\tsb{41.9} \\
\midrule
     \footnotesize{FLOAT\textsuperscript{\textdagger}} & \tsx{91.4}	& \tsx{32.6}	& \tsx{37.5}	& \tsx{26.4}	& \tsx{65.2}	& \tsx{16.9}	& \tsx{40.2}	& \tsx{32.4}	& \tsx{34.2}	& \tsx{50.6}	& \tsx{29.6}	& \tsx{47.5}	& \tsx{37.3}	& \tsx{26.4}	& \tsx{34.5}	& \tsx{32.6}	& \tsx{39.3}	& \tsx{22.6}	& \tsx{68.8}	& \tsx{13.9}	& \tsx{60.5}	& \tsx{30.8}	& \tsx{40.0}\\
     
    \footnotesize{\textbf{FLOAT\textsuperscript{\textdagger} + O}}  
& \tsb{92.3}	& \tsb{36.4}	& \tsb{39.7}	& \tsb{28.5}	& \tsb{65.6}	& \tsb{40.6}	& \tsb{35.5}	& \tsb{35.7}	& \tsb{37.4}	& \tsb{52.2}	& \tsb{33.4}	& \tsb{49.3}	& \tsb{38.9}	& \tsb{32.3}	& \tsb{37.9}	& \tsb{35.8}	& \tsb{41.6}	& \tsb{23.7}	& \tsb{69.7}	& \tsb{14.4}	& \tsb{66.9}	& \tsb{34.3}	& \tsb{43.2} \\

    \bottomrule
  \end{tabular}
 }
  \label{tab:results201}
  \vspace{-3mm}
\end{table}

To show the effectiveness of our approach for multi-object multi-part scene parsing, we apply the recipe for OLAF (see \Cref{fig:intro}) on BSANet~\cite{bsanet}, GMNet~\cite{gmnet}, Deeplabv3~\cite{aspp} and the current state-of-the-art FLOAT~\cite{float}. For training these models, we consider all the Pascal-Part dataset variants. We also apply OLAF to DeepLabV3+~\cite{aspp}, Segformer~\cite{segformer}  trained on PartImageNet~\cite{partimagenet} dataset. We also experiment with different backbones - ResNet-50~\cite{resnet}, Swin Transformer~\cite{swin} and MiT-B2~\cite{segformer}. We apply the same hyperparameters, training strategy, augmentations and pretrained backbones used in the respective methods\footnote{The training details can be accessed from the respective papers.}. For weight adaptation (\Cref{subsec:weight-adaptation}), $n_{warm}$ is set to $5$. All the experiments are conducted on clusters with NVIDIA A100 GPUs.

\section{Experimental Results}

\subsection{Pascal-Part-58, 108 and 201}

\Cref{tab:results201} presents performance metrics for the most challenging variant of the Pascal-Part dataset, Pascal-Part-201. When applying OLAF to  state-of-the-art model FLOAT~\cite{float}, we observe significant improvements: a $\mathbf{4.0}$ increase in $\mathbf{mIoU}$ and a $\mathbf{4.8}$ increase in $\mathbf{sqIoU}$ compared to FLOAT. This improvement is particularly noteworthy given that (a) Pascal-Part-201 is characterized by numerous small parts (b) object categories in this dataset have subtle intra and inter-category part label variations (e.g. `left front leg'/`right front leg'  in \texttt{horse}, \texttt{cow} etc. and `right leg' in \texttt{person}).

\begin{table}[!t]
  \centering  
  \begin{minipage}[b]{0.5\textwidth}
   \caption{Pascal-Part-58\&108 segmentation results. `+ O':augmented with our approach OLAF, `\textdagger': with ViT-H backbone.}
    \centering
    \resizebox{\textwidth}{!}{
      \begin{tabular}{lcccc|cccc}
        \toprule
        \multirow{2}{*}{Method} & \multicolumn{4}{c|}{58} & \multicolumn{4}{c}{108} \\
        \cline{2-9}
        & mIOU & mAvg & sqIOU & sqAvg & mIOU & mAvg & sqIOU & sqAvg \\
        \midrule
        Deeplabv3~\cite{aspp} & 54.3 & 55.4 & 46.0 & 48.4 & 41.3 & 43.6 & 32.2 & 36.1 \\
        \textbf{Deeplabv3 + O} & \textbf{59.0} & \textbf{61.6} & \textbf{52.1} & \textbf{55.6} & \textbf{46.4} & \textbf{51.5} & \textbf{39.2} & \textbf{45.2} \\
        \midrule
        BSANet\cite{bsanet} & 58.2 & 58.9 & 49.3 & 51.5 & 45.9 & 48.4 & 36.6 & 41.0 \\
        \textbf{BSANet + O} & \textbf{59.8} & \textbf{61.7} & \textbf{51.9} & \textbf{55.3} & \textbf{47.1} & \textbf{50.3} & \textbf{38.7} & \textbf{43.5} \\
        \midrule
        GMNet\cite{gmnet} & 59.0 & 61.8 & 49.4 & 54.3 & 45.8 & 50.5 & 35.8 & 41.9 \\
        \textbf{GMNet + O} & \textbf{60.2} & \textbf{63.4} & \textbf{51.6} & \textbf{55.5} & \textbf{47.2} & \textbf{52.1} & \textbf{38.5} & \textbf{44.8} \\
        \midrule
        HIPIE\cite{wang2023hierarchical} & 63.8 &67.1	& 57.2	& 60.7 & - & - & - & - \\
        \midrule
        FLOAT\cite{float} & 61.0 & 64.2 & 54.2 & 57.1 & 48.0 & 53.0 & 40.5 & 45.6 \\
        \textbf{FLOAT + O} & \textbf{62.7} & \textbf{66.1} & \textbf{55.4} & \textbf{58.5} & \textbf{50.3} & \textbf{55.3} & \textbf{43.4} & \textbf{48.4} \\
        \midrule
        FLOAT\textsuperscript{\textdagger} & 62.1	& 65.5	& 55.8	& 58.9 & 48.9	& 54.2	& 41.6	& 46.7 \\
        \textbf{FLOAT\textsuperscript{\textdagger} + O} & \textbf{64.3}	& \textbf{68}	& \textbf{57.7}	& \textbf{60.8} & \textbf{51.5}	&\textbf{56.9}	& \textbf{45.0}	& \textbf{49.9} \\
        \bottomrule
      \end{tabular}
    }
    \label{tab:side_by_side_results58_108}
  \end{minipage}%
  \vspace{1pt}
  \begin{minipage}[b]{0.4\textwidth}
    \centering
    \caption{Results on validation set of the PartImageNet~\cite{partimagenet} dataset. The backbone for each approach is specified separately. ``+ O'': augmented with our proposed method OLAF. 
    }
    \resizebox{\textwidth}{!}{
      \begin{tabular}{lccccc}
        \toprule
        \footnotesize{Method} & \footnotesize{Backbone} & \footnotesize{mIOU} & \footnotesize{mAcc} \\
        \midrule
        \footnotesize{Deeplabv3+\cite{dlv3_plus}} & \tsx{ResNet-50\cite{resnet}} & \tsx{57.53} & \tsx{71.07}  \\
        \footnotesize{Compositor\cite{compositor}} & \tsx{ResNet-50\cite{resnet}} & \tsx{61.44} & \tsx{73.41} \\
        \footnotesize{\textbf{Deeplabv3+ + O}} & \tsx{ResNet-50\cite{resnet}} & \tsb{61.71} & \tsb{74.26}  \\
        \midrule
        \footnotesize{Segformer\cite{segformer}} & \tsx{MiT-B2\cite{segformer}} & \tsx{60.52} & \tsx{71.62}  \\
        \footnotesize{Compositor\cite{compositor}} & \tsx{Swin-T\cite{swin}} & \tsx{64.64} & \tsx{78.31} \\
        \footnotesize{\textbf{Segformer + O}} & \tsx{MiT-B2\cite{segformer}} & \tsb{65.46} & \tsb{79.10}  \\
        \bottomrule
      \end{tabular}
    }
    \label{tab:results_partimagenet}
  \end{minipage}
\end{table}

In both Pascal-Part-58 and Pascal-Part-108 (see 
 \Cref{tab:side_by_side_results58_108}), our approach consistently outperforms the baselines. Specifically, FLOAT with OLAF exhibits improvements of $\mathbf{3.3}$ in $\mathbf{mIoU}$ and $\mathbf{3.5}$ in $\mathbf{sqIoU}$. Similarly, in Pascal-Part-108, FLOAT achieves substantial improvements: $\mathbf{3.5}$ in $\mathbf{mIoU}$ and $\mathbf{4.5}$ in $\mathbf{sqIoU}$.

\subsection{PartImageNet}

For PartImageNet~\cite{partimagenet} (\Cref{tab:results_partimagenet}), OLAF augmented DeepLabV3+~\cite{dlv3_plus} outperforms both DeepLabV3+ and state-of-the-art Compositor~\cite{compositor} with large improvements in mean accuracy. The results  suggest that performance improvement is prominent for more recent, modern backbones (Swin-T~\cite{swin}, MiT-B2~\cite{segformer}). The results also suggest that OLAF's methodology generalizes well to enable gains across datasets and architectural frameworks.


    
  


\subsection{Ablation Studies}
\label{sec:ablation}

We conduct extensive ablation experiments with current state-of-the-art model FLOAT~\cite{float} trained on Pascal-Part-201 (\Cref{tab:ablation_table}).

 \begin{table}[!t]
   \centering
   \caption{Ablation experiments on PASCAL-Part-201 (\Cref{sec:ablation}). We use  FLOAT\textsuperscript{\textdagger} as the base model.}
   \resizebox{\linewidth}{!}{
     \begin{tabular}{@{}c|ccc|ccccc@{}}
       \toprule
        & LDF & Edge-Map & Fg/Bg-Map & mIoU & mAvg & sqIoU & sqAvg & mIoU$_{small}$\\
      \midrule
      \multirow{7}{*}{\begin{tabular}[c]{@{}l@{}}Input  Channel Presence and \\
      Architectural Changes
      \end{tabular}} & $-$ & $-$ & $-$ & 
      37.7	& 46.9	& 30.8	& 40.0 & 24.0\\ 
      & \checkmark &&& 38.8	& 48.1	& 31.8	& 41.0 & 25.7\\
      & & \checkmark && 38.9	& 48.2	& 32.2	& 41.3 & 24.5\\
      &&& \checkmark & 39.1	& 48.3	& 32.0	& 41.2 & 24.6\\   
      && \checkmark & \checkmark & 39.2	& 48.3	& 32.2	& 41.4 & 24.8\\          
      \midrule
       \textbf{OLAF} & \checkmark & \checkmark & \checkmark & \textbf{40.9}  & \textbf{50.5} & \textbf{34.3}  & \textbf{43.2} & \textbf{26.9} \\
      \midrule
      fg map baseline & \multicolumn{3}{c|}{Segment Anything (SAM)~\cite{kirillov2023segment}} &   40.5 & 50.2 & 34.0 & 42.8 & 26.4\\
      \midrule
      \multirow{2}{*}{Edge Map baselines} & \multicolumn{3}{c|}{EDTER~\cite{Pu_2022_CVPR}} &   39.5	& 49.1	& 33.1	& 41.9 & 24.9\\
      & \multicolumn{3}{c|}{Canny~\cite{canny_edge}} & 39.0 & 48.8 & 32.6 & 41.6 & 24.3\\
      \midrule
      \multirow{2}{*}{Added depth map baselines} & \multicolumn{3}{c|}{Marigold~\cite{ke2023repurposing}} &  40.7	& 50.2	& 34.2	& 43.1 & 25.1\\
      & \multicolumn{3}{c|}{Depth Anything~\cite{depthanything}} & 40.8 & 50.4 & 34.2 & 43.2 & 25.2\\
      \midrule     
      \multirow{2}{*}{\begin{tabular}[c]{@{}l@{}} Optimization \\
      (Input Layer Weight Adaptation)\end{tabular}} & \multicolumn{3}{c|}{Random-5} & 35.2	& 44.2	& 28.4	& 37.4 & 19.7\\
      & \multicolumn{3}{c|}{Average-RGB-5\cite{wang2016temporal}} & 36.3	& 45.5	& 29.2	& 38.9 & 20.5\\
      & \multicolumn{3}{c|}{Adapt-n-Freeze\cite{reddit_computervision}} & 38.2	& 47.1	& 31.3	& 39.5 & 21.7\\
      & \multicolumn{3}{c|}{Random-2} & 40.2	& 49.6	& 33.0	& 41.6 & 24.3\\
       \bottomrule
     \end{tabular}
   } 
   \label{tab:ablation_table}
 \end{table}

\noindent \textbf{Input channel presence and architectural changes:} Considering the LDF module alone, there is an initial improvement in mIoU from 37.7 to 38.8. The introduction of edge and foreground/background cues improves the mIoU to 39.2. However, the most substantial gains are observed when all components are combined, achieving an mIoU of 40.9. 

\begin{figure*}[!t]
  \centering
    \includegraphics[ width=\linewidth, height=7cm]{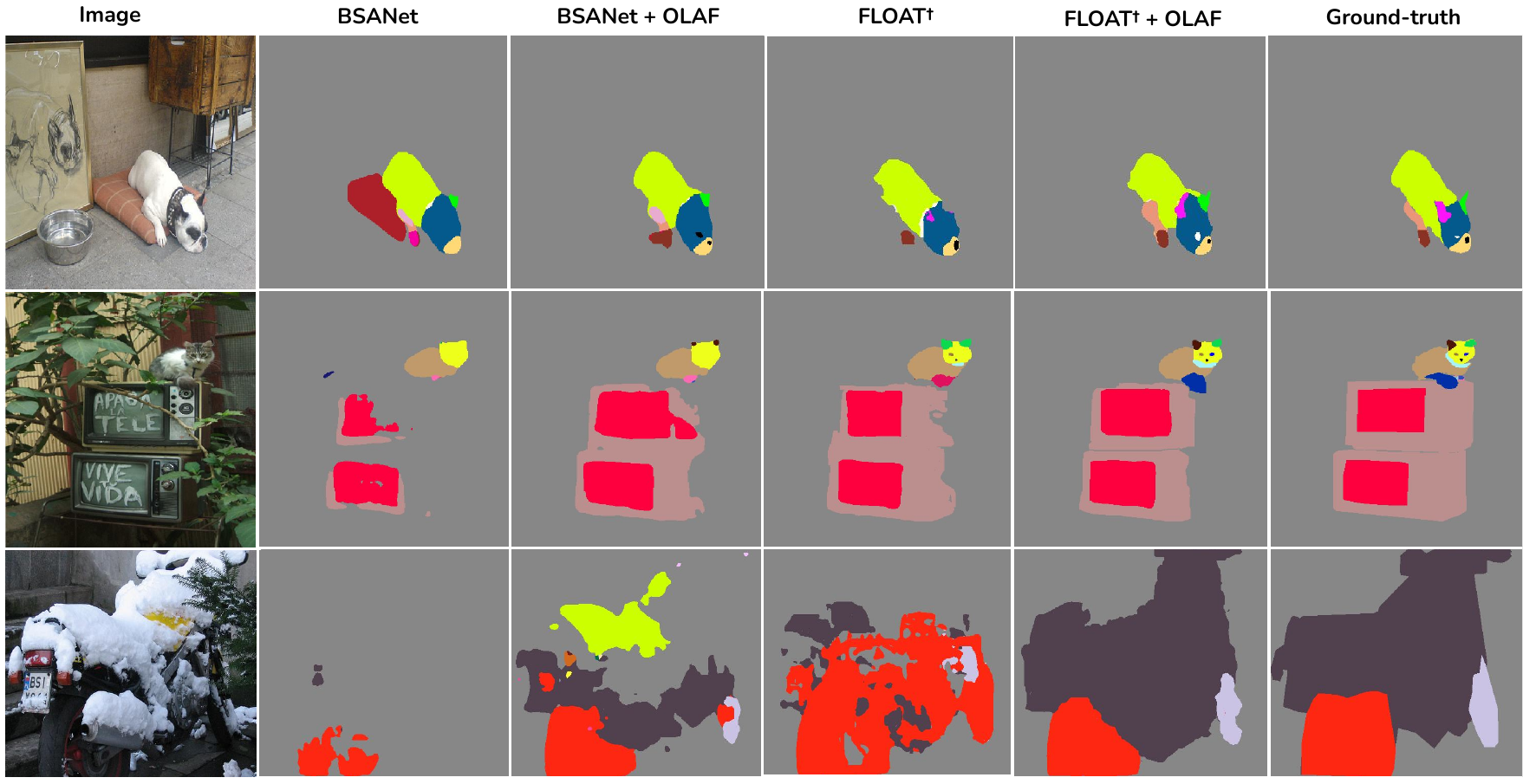}
    \caption{Qualitative comparison on Pascal-Part-201. OLAF consistently improves the performance of previous methods (BSANet~\cite{bsanet}, FLOAT~\cite{float}). This is especially seen for small parts as shown in Row 1 (eyes, ears, nose and right-front-leg), Row 2 (eye, ears, nose, mouth and tail) and occluded parts as shown in Row 3 (parts of the \texttt{motorbike}). 
}    \label{fig:qualitative_fig}
\end{figure*}

Similar to the convention from COCO~\cite{lin2014microsoft} for objects, we define `small' parts as those smaller than $25\times25$ pixels by area. We report the corresponding measure ($mIoU_{small}$) in \Cref{tab:ablation_table} (last column). Clearly, the inclusion of LDF enables noticeable gains for small parts compared to input channels' inclusion.

\subsubsection{Input channel baselines:}

As a baseline for foreground map input channel, we used the foreground map obtained by combining object mask outputs from Segment Anything (SAM)~\cite{kirillov2023segment} but this led to a slightly lower performance. 
Empirically, we observed that SAM does not accurately segment certain object categories such as \texttt{potted plant}, and tends to omit tiny parts of certain classes (e.g. tail of \texttt{airplane}, \texttt{bird}, \texttt{cat}, \texttt{cow}, \texttt{dog}, \texttt{horse}, \texttt{sheep}). 

For edge map, we explored Canny~\cite{canny_edge} and EDTER~\cite{edter} as alternatives. Compared to our default choice (HED~\cite{hed}), these choices fail to strike the right balance in terms of boundary edge density -- Canny maps contain too few semantically crucial edges while EDTER maps tend to be too dense. See Supplementary for examples illustrating these observations.

Intrigued by the performance enhancement capabilities of  additional input channels, we considered adding depth maps obtained from monocular depth estimation approaches (Marigold~\cite{ke2023repurposing}, Depth Anything~\cite{depthanything}) as additional input channels. In effect, this increased the number of channels to 6. The weight adaptation procedure was applied as described earlier (Sec.~\ref{subsec:weight-adaptation}). But the inclusion of depth map did not lead to performance gains. A likely reason could be that depth cues are likely more useful for differentiating objects, particularly when the scene has a large depth of field. But for intra-object parts, depth might not vary much. Consequently, depth map might not be as effective for aiding part segmentation.


\subsubsection{Weight Adaptation:} 
\label{sec:weightadaptationschemes}

To examine the effect of our weight adaptation procedure at input layer (\Cref{subsec:weight-adaptation}), we conducted experiments with alternate schemes as described below.

\begin{itemize}[leftmargin=*]
\item \textit{Random-5}: Retain weights of the entire backbone except for the input layer. The input layer dimension is reconfigured from 3 channel input to 5 channel input and weights for this layer's filters are initialized randomly.

\item  \textit{Random-2}: Retain weights of the entire backbone, including those for the 3 channels in the original (RGB) input layer. The weights for two newly added channels in the input layer are initialized randomly.

\item  \textit{Average-RGB-5}~\cite{wang2016temporal}: Average the channel-wise weights of the original (RGB) network's input layer. Initialize all 5 channels with this average. 

\item \textit{Adapt-n-Freeze}~\cite{reddit_computervision}: First, include a convolution layer to match augmented input (5 filters) and then include $1 \times 1$ filter so that output (3 channels) is compatible with base (RGB) network. Freeze the base network and train the adapter layers (Conv2D, $1 \times 1$). Then unfreeze and train the entire layer augmented base network together. 
\end{itemize}

\begin{figure*}[!t]
  \centering
    \includegraphics[ width=\linewidth, height=7cm]{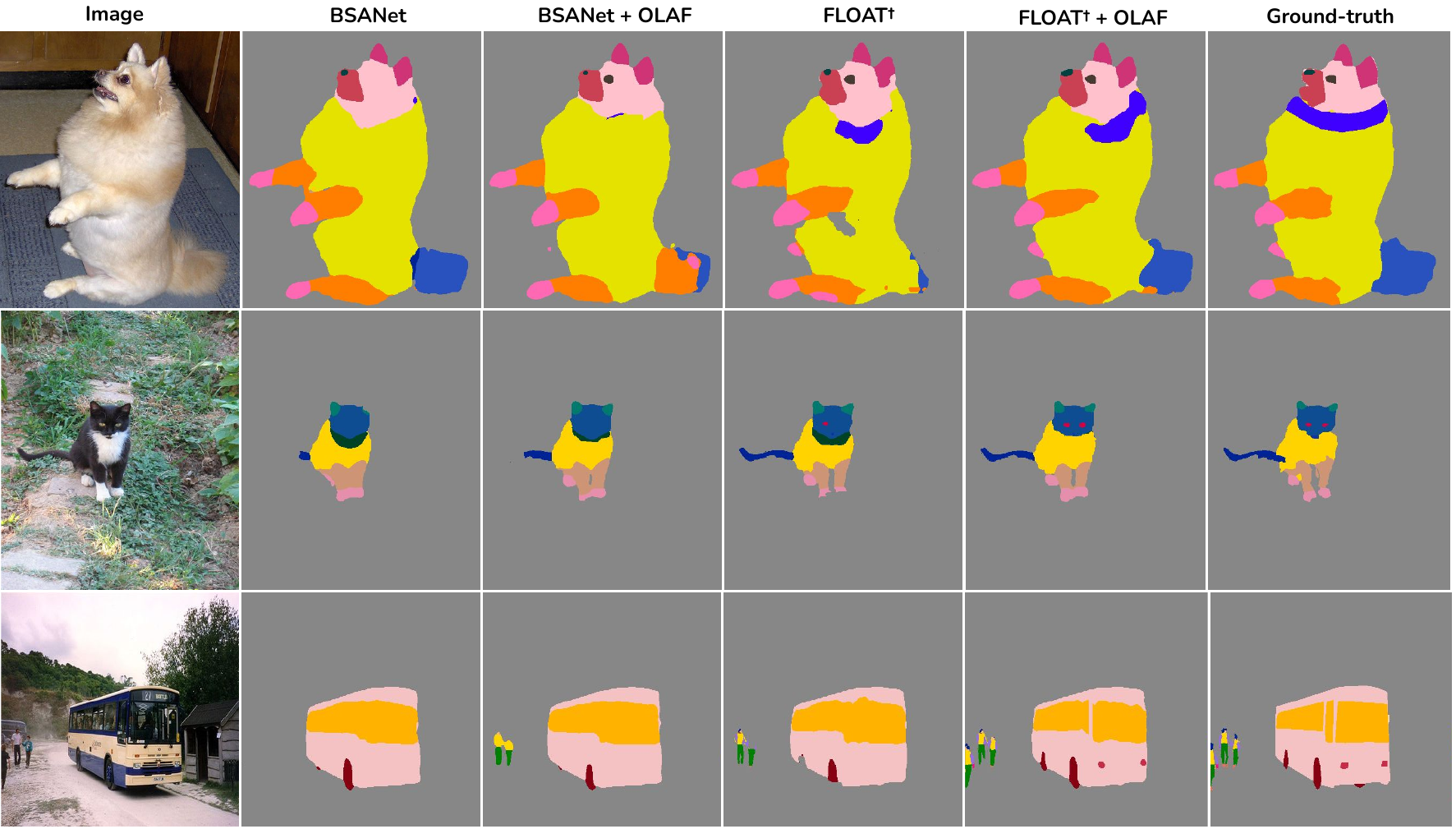}
    \caption{Qualitative comparison on Pascal-Part-108. OLAF consistently improves the performance of previous methods (BSANet~\cite{bsanet}, FLOAT~\cite{float}). This is especially seen for small parts as shown in Row 1 (eyes and tail), Row 2 (eyes, ears, front-paw and tail) and Row 3 (wheel, headlight and very tiny parts of humans). 
}    \label{fig:qualitative_fig108}
\end{figure*}

To ensure fair and consistent comparison, all methods used a warm-up phase consisting of $n_{warm}=5$ epochs.  As shown in \Cref{tab:ablation_table}, our proposed approach provides the best performance. In practice, we found that \textit{Random-5} method resulted in unstable learning. 
This instability arises from initializing the input layer with random weights, causing erratic gradient flow and impacting the pretrained weights of subsequent layers in the backbone. While \textit{Random-2} had stable optimization, the performance suffers from erratic gradients due to random initialization. The other adaptation schemes (\textit{Average-RGB-5}~\cite{wang2016temporal}, \textit{Adapt-n-Freeze}~\cite{reddit_computervision}) also had similar instability issues. In contrast, our weight adaptation approach ensures stable optimization and distinctly improved performance.

Overall, the ablation experiments suggest that all the ingredients of our recipe — object-based channels, LDF, targeted weight adaptation — are crucial and contribute to the enhanced performance of OLAF.

\subsection{Qualitative Results}

\begin{figure*}[!t]
  \centering
    \includegraphics[width=\linewidth, height=5.7cm]{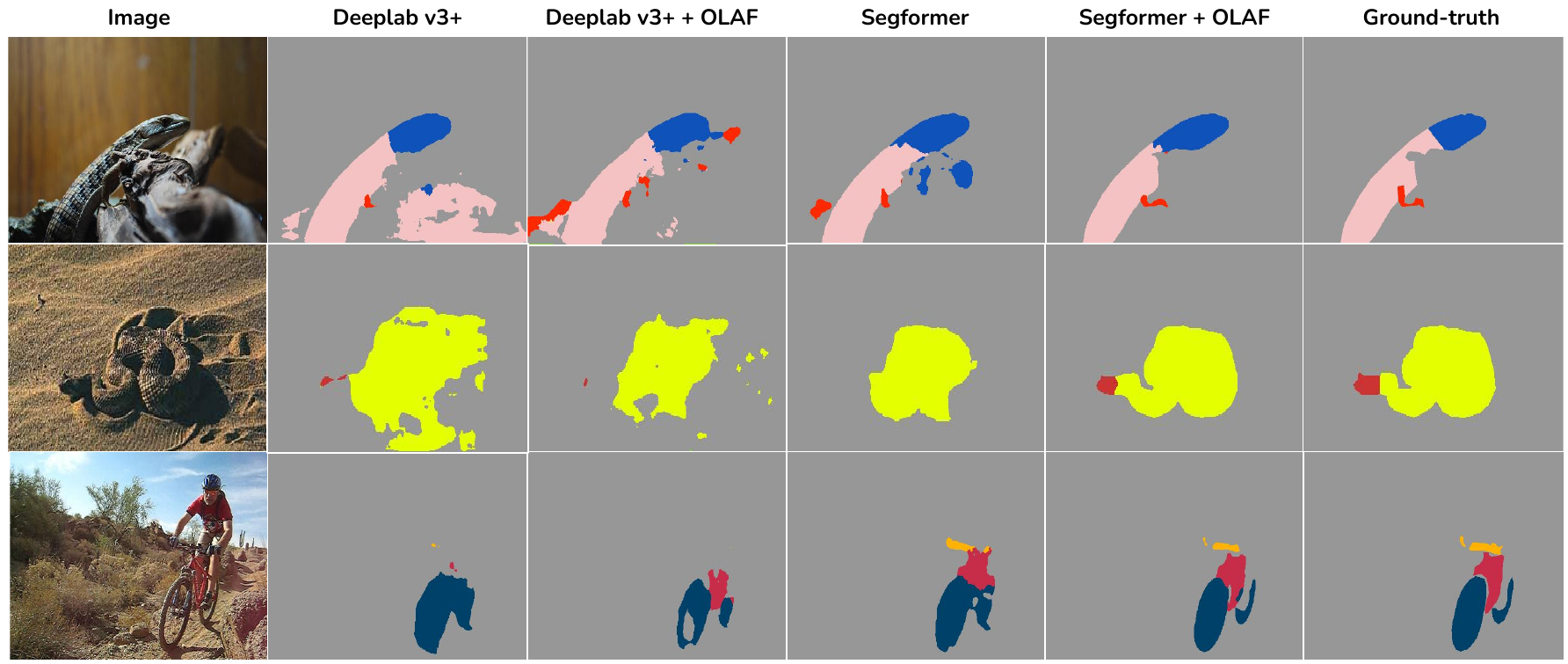}
    \caption{Qualitative comparison for images from PartImageNet~\cite{partimagenet}. OLAF consistently improves the segmentation quality, especially for the harder small parts of objects.}
    \label{fig:part_imgnet_qualitative_fig}
\end{figure*}

As seen in \Cref{fig:qualitative_fig}, OLAF consistently improves performance for BSANet~\cite{bsanet} and FLOAT~\cite{float}. It particularly improves the segmentation of small objects (\texttt{cat} in Row 2), small parts as shown in Row 1 (`right front leg', `right eye' and `left/right ear' of \texttt{dog}) and Row 2 (parts in face region of \texttt{cat}) and occluded parts as shown in Row 3 (parts of \texttt{motorbike} especially the body). A similar trend can also be seen for Pascal-Part-108 (\Cref{fig:qualitative_fig108}) and PartImageNet dataset in \Cref{fig:part_imgnet_qualitative_fig}. A limitation of OLAF stems from its dependence on the additional input channels (\Cref{sec:fgbgedge}). Poor-quality channel masks can affect segmentation results -- see Supplementary for examples.

\subsection{Computational Metrics}

As \Cref{tab:compute_table} shows, inclusion of OLAF leads to a modest rise in trainable parameters for BSANet (20\%), GMNet (8\%), FLOAT (14\%) and FLOAT\textsuperscript{\textdagger} (1.5\%). The training time per epoch shows a slight uptick (5\% - 10\%) while inference time increases by 0.26s for state of the art network. Overall, there is a pragmatic balance between OLAF's performance gain and increase in computational demand.

\begin{table}[!t]
  \centering  
  \begin{minipage}[t]{0.5\textwidth}
  \caption{Compute metrics for various methods with inclusion of OLAF ('+ O') and without (baseline) on Pascal-Parts-58. The batch size is 16.}
  \resizebox{\textwidth}{!}{
  \begin{tabular}{@{}lccccc@{}}
    \toprule
    \footnotesize{Method}  & \footnotesize{Trainable} & \footnotesize{Train Time} & \footnotesize{Test Time} \\
    &  \footnotesize{Param(M)} & \footnotesize(mins/epoch) &\footnotesize(secs/image) & \\
    \midrule
    \footnotesize{BSANet~\cite{bsanet}}  & $\;$\tsx{63.9} & \tsx{31.3} & \tsx{0.46} \\
    \footnotesize{\textbf{BSANet + O}} & $\;$\tsx{76.8} & \tsx{32.9} & \tsx{0.81} \\
    \midrule
    \footnotesize{GMNet~\cite{gmnet}} & \tsx{123.4} & \tsx{16.6} & \tsx{0.49} \\
    \footnotesize{\textbf{GMNet + O}} & \tsx{133.9} & \tsx{17.4} & \tsx{0.77} \\
    \midrule
    \footnotesize{FLOAT~\cite{float}} & $\;$\tsx{76.2} & \tsx{18.3} & \tsx{0.98} \\
    \footnotesize{\textbf{FLOAT + O}} & $\;$\tsx{86.7} & \tsx{20.1} & \tsx{1.36} \\   
    \midrule
    \footnotesize{FLOAT\textsuperscript{\textdagger}} & $\;$\tsx{674.7} & \tsx{100.8} & \tsx{8.33} \\
    \footnotesize{\textbf{FLOAT\textsuperscript{\textdagger} + O}} & $\;$\tsx{685.2} & \tsx{101.1} & \tsx{8.59} \\
    \bottomrule
  \end{tabular}
  }
  \label{tab:compute_table}
  \end{minipage}
  \vspace{1pt}
  \begin{minipage}[t]{0.45\textwidth}
  \caption{FLOAT~\cite{float} with different train and inference time resolutions for Pascal-Parts-58. The underlined value is the default setup for FLOAT. OLAF's results are included for reference.}
   \resizebox{\textwidth}{!}{
  \begin{tabular}{@{}lccc@{}}
    \toprule
    \footnotesize Method & \footnotesize Train & \footnotesize Inference & \footnotesize mIoU  \\
    \midrule
    \multirow{3}{*}{\footnotesize FLOAT~\cite{float}} & \multirow{3}{*}{\footnotesize 513 x 513}  & \footnotesize 513 x 513& \footnotesize 60.8 \\
    &  & \footnotesize \underline{770 x 770} & \footnotesize 61.0 \\
    &  & \footnotesize 1024 x 1024 & \footnotesize 57.4 \\
    \midrule
    \multirow{2}{*}{\footnotesize FLOAT~\cite{float}} & \multirow{2}{*}{\footnotesize 770 x 770}  & \footnotesize 770 x 770 & \footnotesize 60.5 \\
    &  & \footnotesize 1024 x 1024 & \footnotesize 57.6 \\
    \midrule
    \footnotesize \textbf{FLOAT+O} & \footnotesize $\mathbf{513 \times 513}$  & \footnotesize $\mathbf{770 \times 770}$ & \footnotesize $\mathbf{62.7}$ \\
    \bottomrule
  \end{tabular}  
  \label{tab:res}
  }
  \end{minipage}
\end{table}

\subsection{Effect of input resolution} 

During inference, all baselines conventionally operate on a higher resolution input ($770 \times 770$) compared to the resolution during training ($513 \times 513$). 
We examined the effect of higher resolution during training and inference on FLOAT~\cite{float} (the SOTA baseline). The results (Table~\ref{tab:res}) show that (i) for inference, there is a limit to the gain achieved by increasing resolution (ii) training with higher (than default) resolution does not necessarily provide a stronger baseline model. Moreover, a higher resolution significantly increases run time and memory requirements. The findings reemphasize the effectiveness of OLAF's plug-and-play design for enhancing performance without requiring an increase in default input resolution.

\section{Conclusion}


OLAF is a broadly applicable plug-and-play approach for enhancing multi-object multi-part scene parsing. OLAF's recipe consists of (i) \textit{augmenting RGB input with object-based channels} (fg/bg, boundary edges). This acts as a structural inductive bias and guides the model to focus on relevant parts throughout optimization (ii) \textit{using lightweight yet efficient low-level dense feature guidance (LDF)}. This acts as an inductive bias for small and thin parts. (iii) \textit{targeted weight-adaptation} for stable optimization with augmented input. 

Our approach shows the benefit of efficiently infusing targeted inductive biases into existing models. OLAF also addresses multiple limitations of existing methods. OLAF consistently improves segmentation performance, especially for small and thin parts, across a broad spectrum of challenging datasets and architectures. We expect OLAF's lightweight and modular enhancements to also benefit other computer vision tasks such as panoptic part segmentation~\cite{panoptic_1, panoptic_3, panotic_2, panoptic_4}.


%
%
\bibliographystyle{splncs04}
\bibliography{main}
\end{document}


\maketitle
\tableofcontents

\section{Incorporation of LDF on Different Architecture Types}

\subsection{U-Net}

\begin{figure}[h]
\centering
\includegraphics[width=0.5\linewidth]{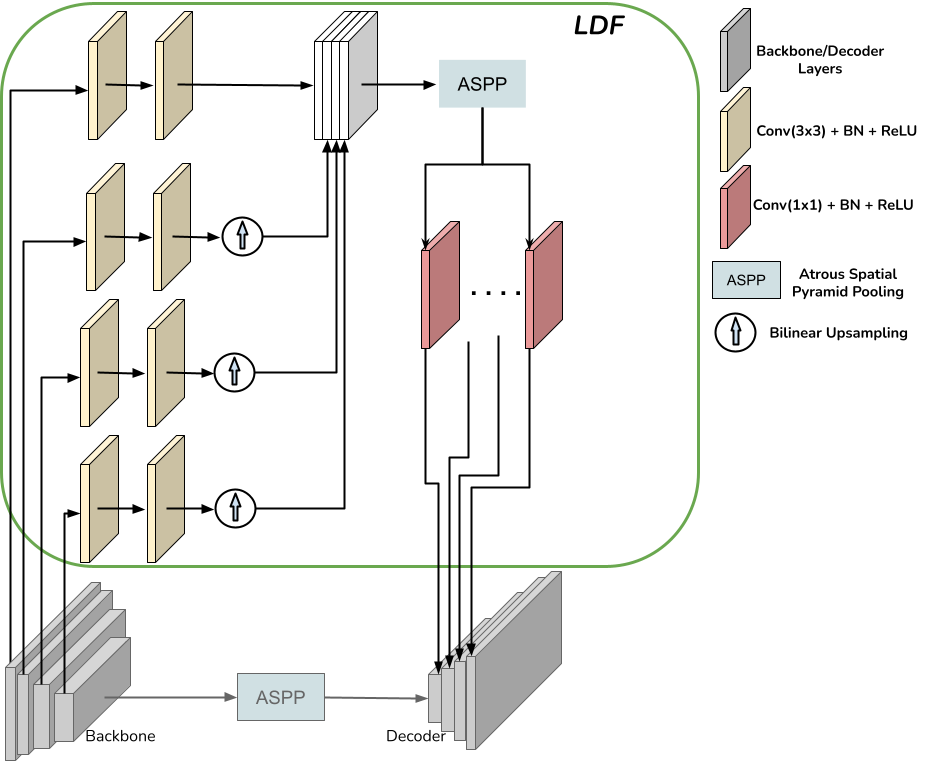}
\caption{Incorporation of "LDF" in U-Net-like architectures}
\label{fig:ldf_bsanet}
\end{figure}

In U-Net-style architecture, LDF is embedded within each encoder layer, and its output is passed to the respective decoder layers (see Fig.~\ref{fig:ldf_bsanet}).

\subsection{Transformer}

\begin{figure}[h]
  \centering
\includegraphics[width=0.5\linewidth]{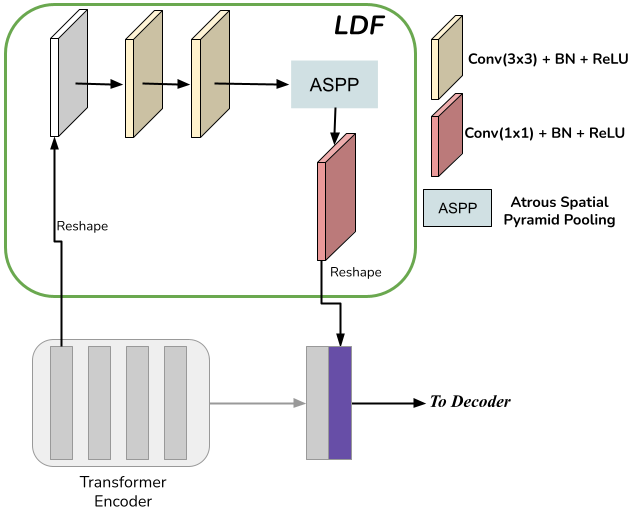}
\caption{Incorporation of "LDF" in Transformer-like architectures}
\label{fig:transformer}
\end{figure}

In Transformer-style architecture, we apply LDF solely to the first encoder block. Its output is directly merged with the encoder output and then passed to the decoder (see Fig.~\ref{fig:transformer}).

\section{Extraction of fg/bg and edge masks}

We use pre-trained Deeplabv3+ to derive object segmentation masks, subsequently transformed into fg/bg masks (refer to \textbf{Sec 3.1} in the main paper). Additionally, for edge masks, we employ pretrained Holistically-Nested Edge Detection (HED) (refer to \textbf{Sec 3.1} in the main paper). \\
\textbf{\textit{Note: Both networks (Deeplabv3+ and HED) remain frozen during OLAF training.}}

\begin{figure*}[t]
  \centering
\includegraphics[width=\textwidth, height=0.8\textheight]{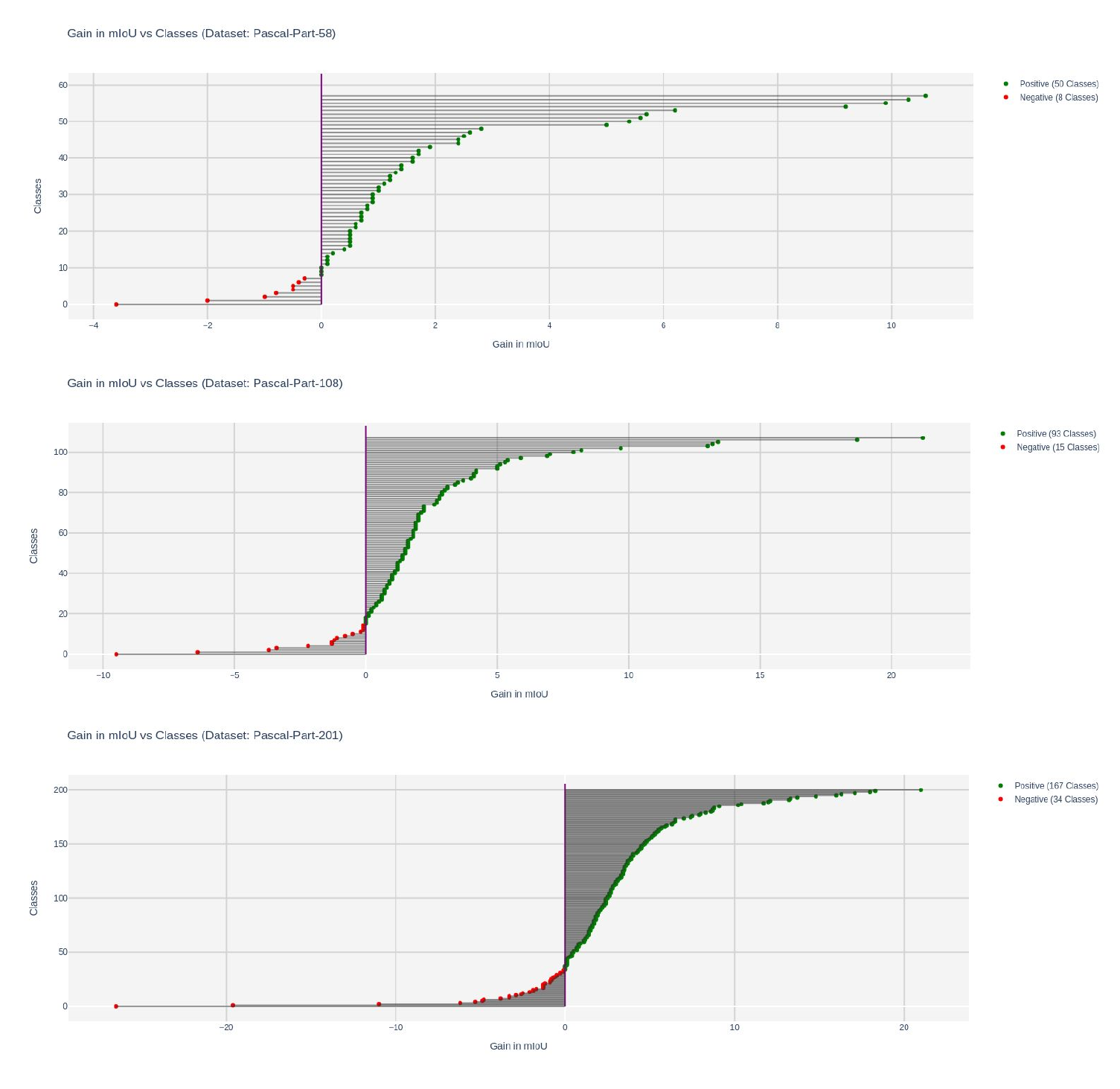}
\caption{Gain in mIoU vs Classes across all Pascal-Part datasets. Choice of model is FLOAT (current SOTA across all Pascal-Part datasets) }
\label{fig:gain_vs_miou}
\end{figure*}

\section{Gain Analysis}

In this section, we present the increase in mIoU in each class across all variants of Pascal-Part datasets. As depicted in Fig.~\ref{fig:gain_vs_miou}, FLOAT paired with OLAF consistently enhances overall performance. Across all dataset variants, OLAF notably enhances the performance of \textbf{83\% to 86\%} of the classes. An intriguing finding reveals that in pascal-part-201, OLAF notably raises the mIoU by \textbf{over 10\%} for 17 classes, and for some, the improvement is \textbf{over 20\%}.

\section{OLAF's Robustness to Mask Noise}
\begin{figure}[H]
  \centering
  \includegraphics[width=0.9\textwidth, height=0.4\textheight]{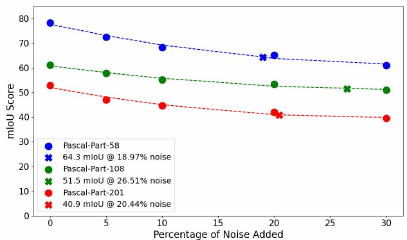}
  \caption{\textbf{Robustness of our approach to mask noise:} The plot depicts mIoU scores as a function of noise added to the ground-truth binary fg/bg and edge masks of different datasets. Noise levels of 0\%, 5\%, 10\%, 20\% and 30\% were introduced by randomly flipping the corresponding percentage of pixel labels in the masks. Note that 0\% noise level corresponds to using ground-truth masks during training and inference. OLAF achieves mIoU scores of 64.3 (part-58), 51.5 (part-108) and 40.9 (part-201), using HED edges which correspond to noise levels of 18.97, 26.51 and 20.44, respectively (represented by \textbf{\texttt{x}}'s in the figure).}
  \label{fig:miou_vs_noise}
\end{figure}

To examine the effect of mask noise on performance, we systematically add various amounts of random noise to the ground-truth masks during training and inference. From Figure~\ref{fig:miou_vs_noise}, we see that performance does not degrade too drastically with increase in noise. We also see that the current quality of masks used in OLAF corresponds to $18.9\% - 26.5\%$ mask noise.Note that OLAF achieves significant performance gains (up to $\mathbf{4.0}$ mIoU) despite this level of segmentation and edge detection noise. More promisingly, the results in Figure~\ref{fig:miou_vs_noise} suggest that \textbf{better masks from future improvements in semantic edge detection and segmentation approaches by the research community would lead to even better results}.

\section{Observations and Limitations}

\subsection{Observations}
\begin{enumerate}
\item Models incorporated with our 3-step recipe \textbf{(OLAF)} produce better results.
\item OLAF improves segmentation of small/thin parts \textbf{(LDF)}.
\item  OLAF helps segmenting missing foreground areas (union of part/obj pixels) \textbf{(Foreground/background as auxiliary channel in input)}.
\item OLAF helps in delineating boundaries between parts \textbf{(Edge as auxiliary channel in input)}.

\end{enumerate}

\begin{figure}[H]
  \centering
  \includegraphics[width=0.9\textwidth, height=0.7\textheight]{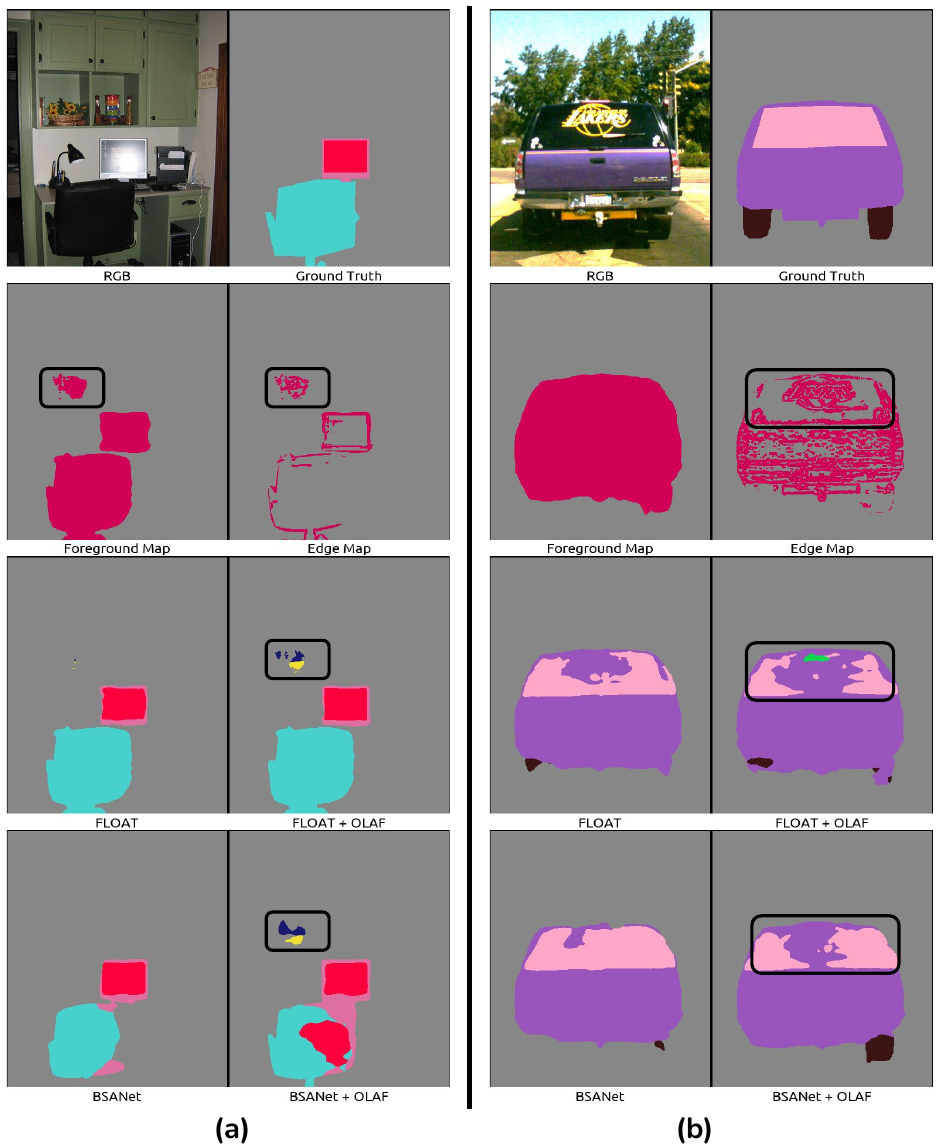}
  \caption{Results depicting the limitations of OLAF}
  \label{fig:olaf_limitations}
\end{figure}

\subsection{Limitations}
A limitation of OLAF stems from its dependence on the additional input channels. Poor-quality channel masks can affect segmentation results.  As illustrated in Fig.~\ref{fig:olaf_limitations}\textcolor{red}{(a)}, the noise present in fg/bg mask also propagated to the final results. A similar limitation can be seen in Fig.~\ref{fig:olaf_limitations}\textcolor{red}{(b)}, where noisy edge mask affected the final results. 

\section{Illustration of OLAF's mechanism along with existing SOTA (FLOAT)}

\begin{figure}[H]
  \centering
  \includegraphics[width=\textwidth]{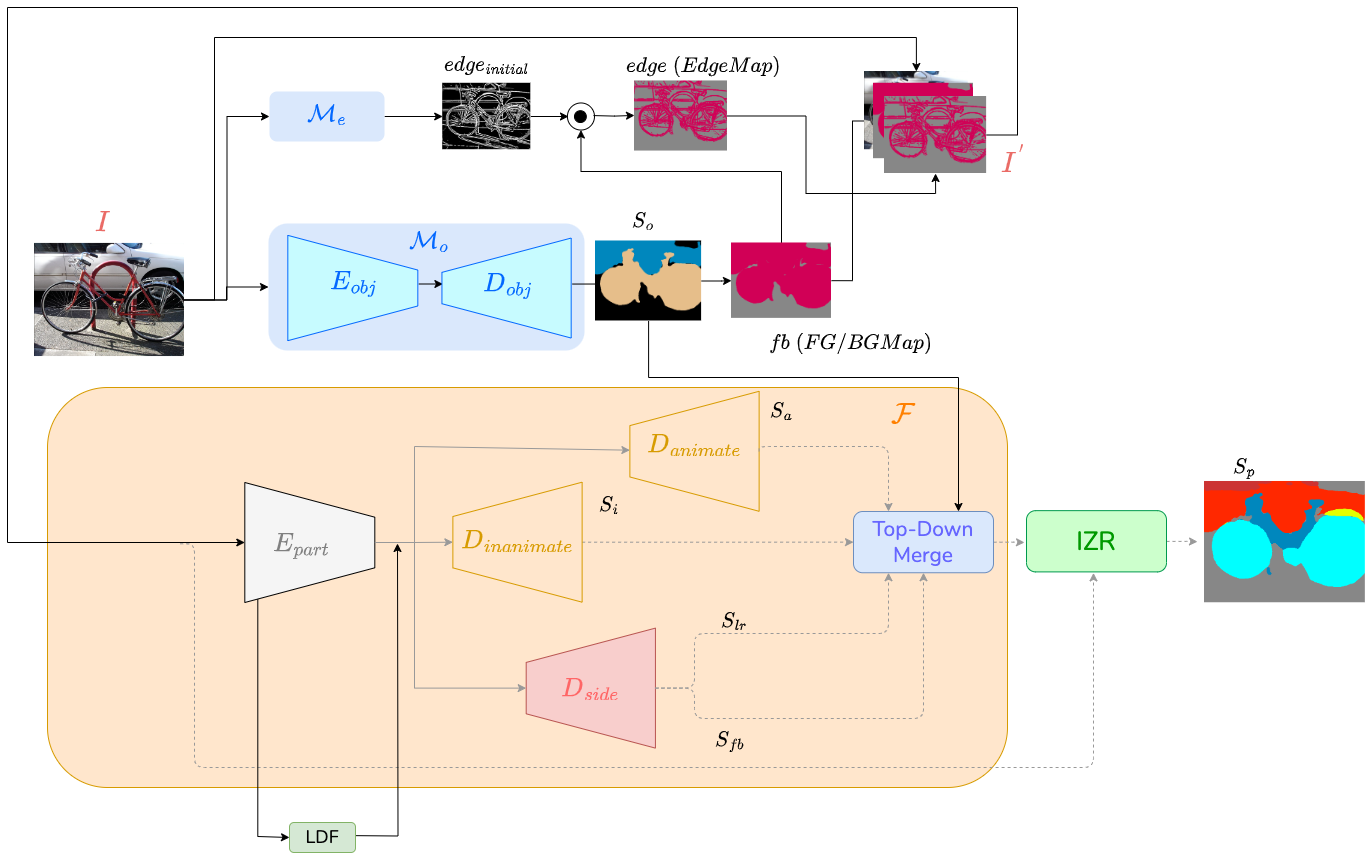}
  \caption{An illustration of OLAF's operations and how it interfaces with FLOAT~\cite{float} - the current SOTA architecture. FLOAT's core module is denoted as $\mathcal{F}$ - see ~\cite{float} for additional details. Refer to main paper for additional details on OLAF's modules and outputs. $I$ is input 3-channel RGB image and $I^{'}$ is the augmented 5-channel input to FLOAT.}
  \label{fig:olaf_float}
\end{figure}

\section{Edge Choices}

\begin{figure}[H]
  \centering
  \includegraphics[width=\textwidth]{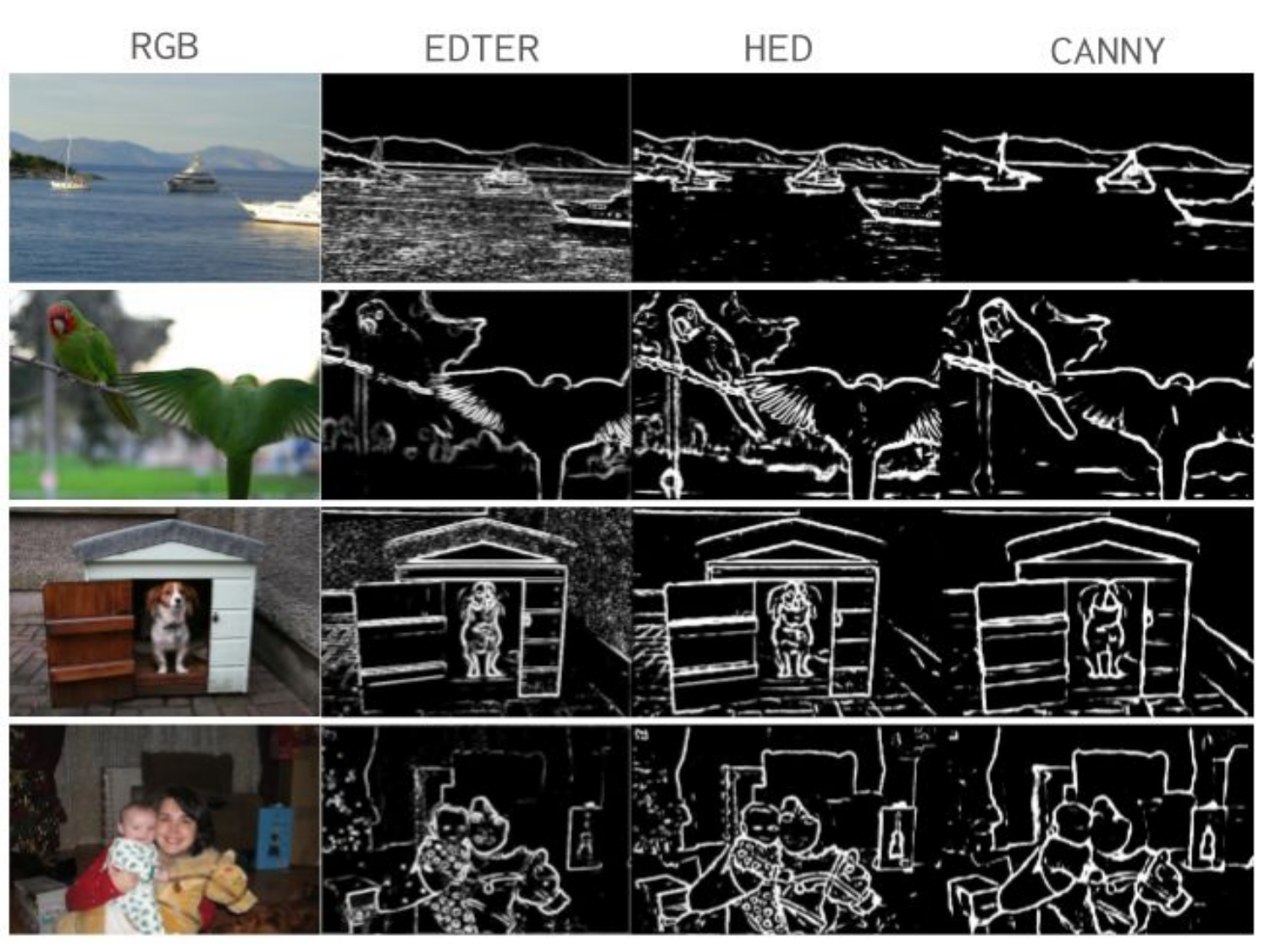}
  \caption{An illustration of various edge detection methods we have experimented with. Canny edge maps lack many semantically important edges, while EDTER maps tend to be overly dense. Holistically-Nested Edge Detection (HED) represents the optimal balance we discovered, achieving neither excessive density nor sparseness while preserving semantically meaningful cues.}
  \label{fig:olaf_float}
\end{figure}

\section{Qualitative comparison on fg/bg mask}

\begin{figure}[H]
  \centering
  \includegraphics[width=\textwidth]{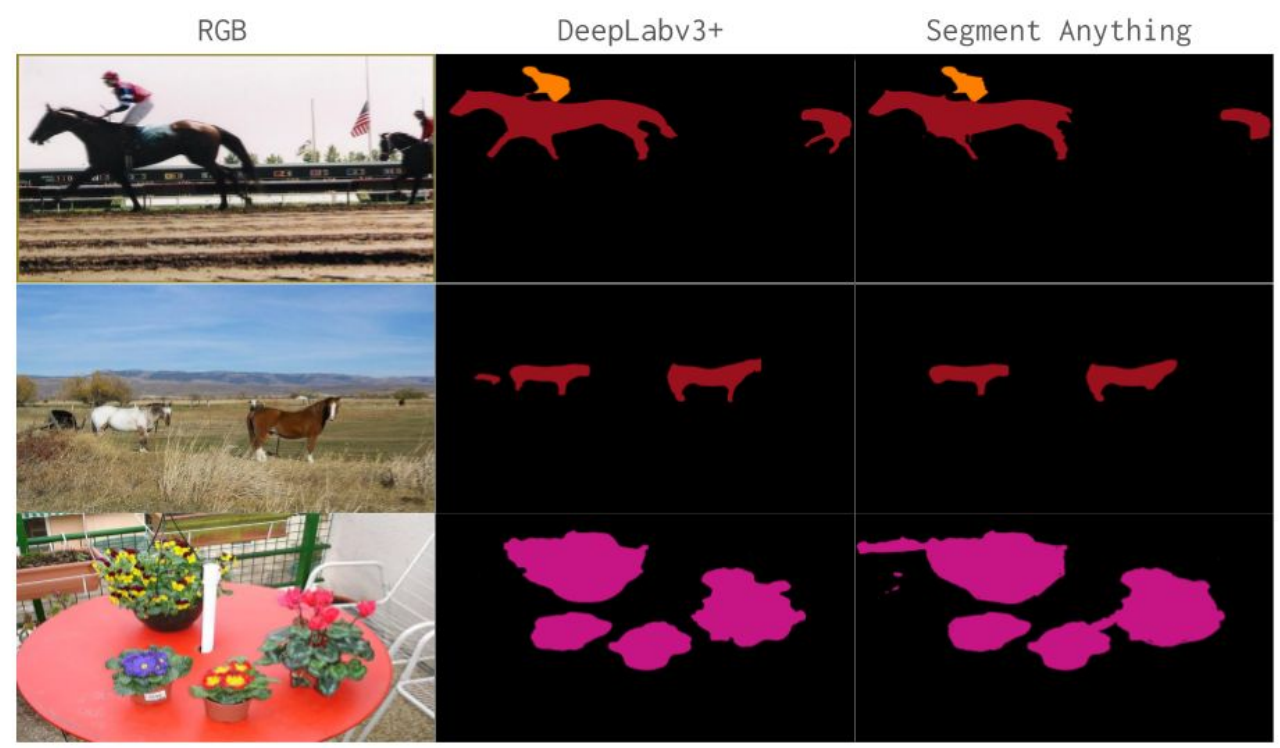}
  \caption{An illustration of scenarios where the Segment Anything model struggles, ultimately resulting in low-quality foreground/background masks. As depicted in the figure, SAM fails to accurately segment certain object categories, such as potted plants, and tends to overlook small parts such as the tails, legs, hand and other similar classes.}
  \label{fig:olaf_float}
\end{figure}

\section{Qualitative Results}

\subsection{Pascal-Part-58}

\begin{figure}[H]
  \centering
  \includegraphics[width=0.9\textwidth, height=0.8\textheight]{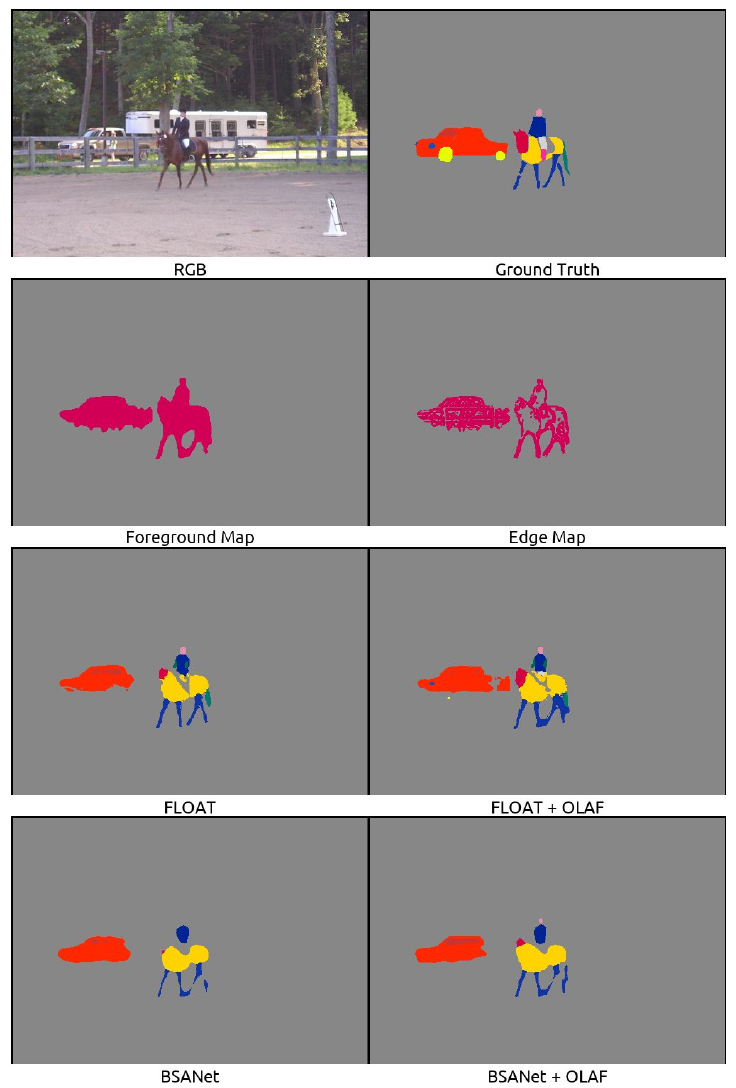}
  \caption{}
  
\end{figure}

\clearpage

\begin{figure}
  \centering
  \includegraphics[width=\textwidth, height=0.9\textheight]{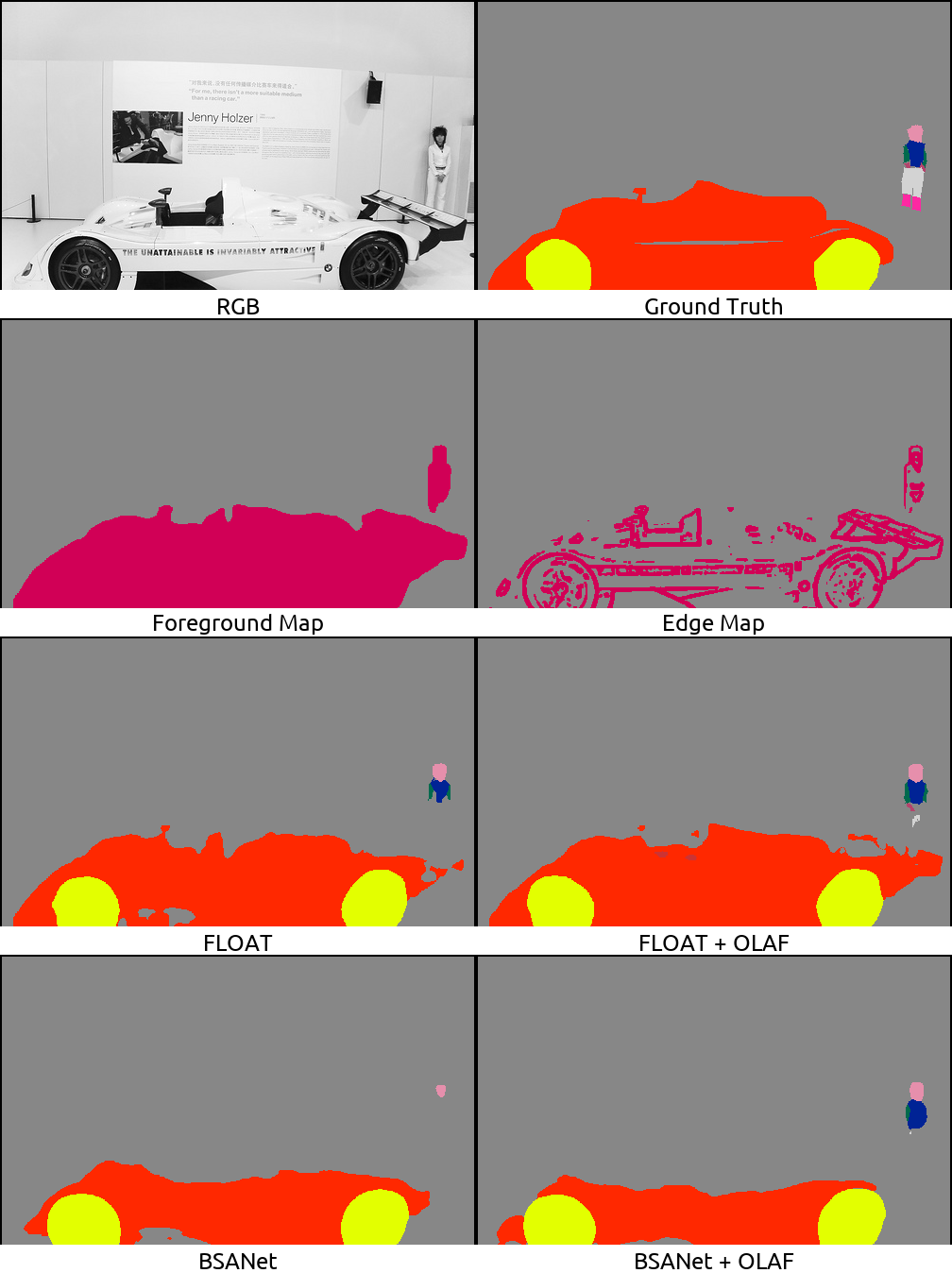}
  \caption{}
\end{figure}

\clearpage

\begin{figure}
  \centering
  \includegraphics[width=\textwidth, height=\textheight]{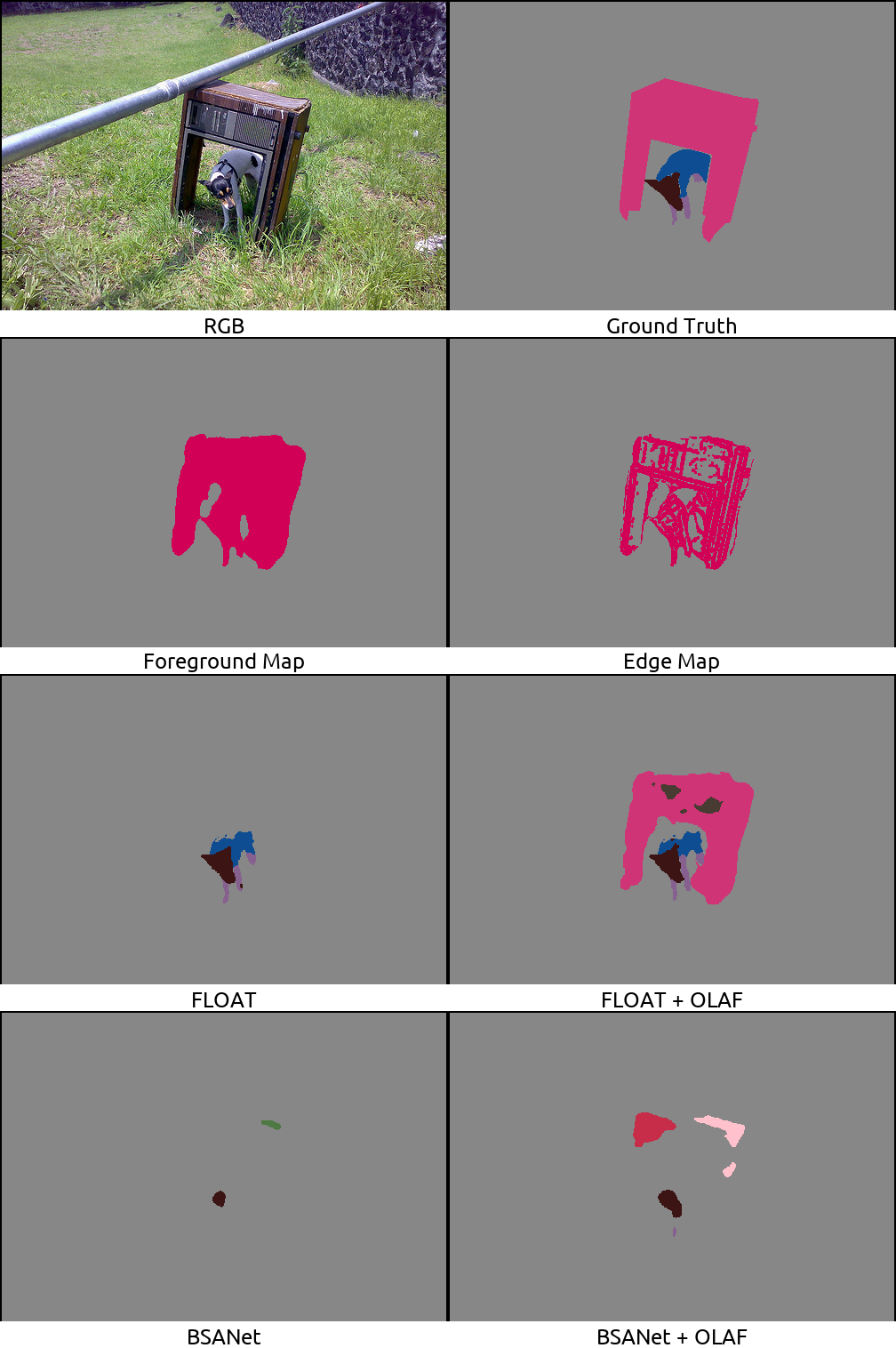}
  \caption{}
\end{figure}

\clearpage

\begin{figure}
  \centering
  \includegraphics[width=\textwidth, height=\textheight]{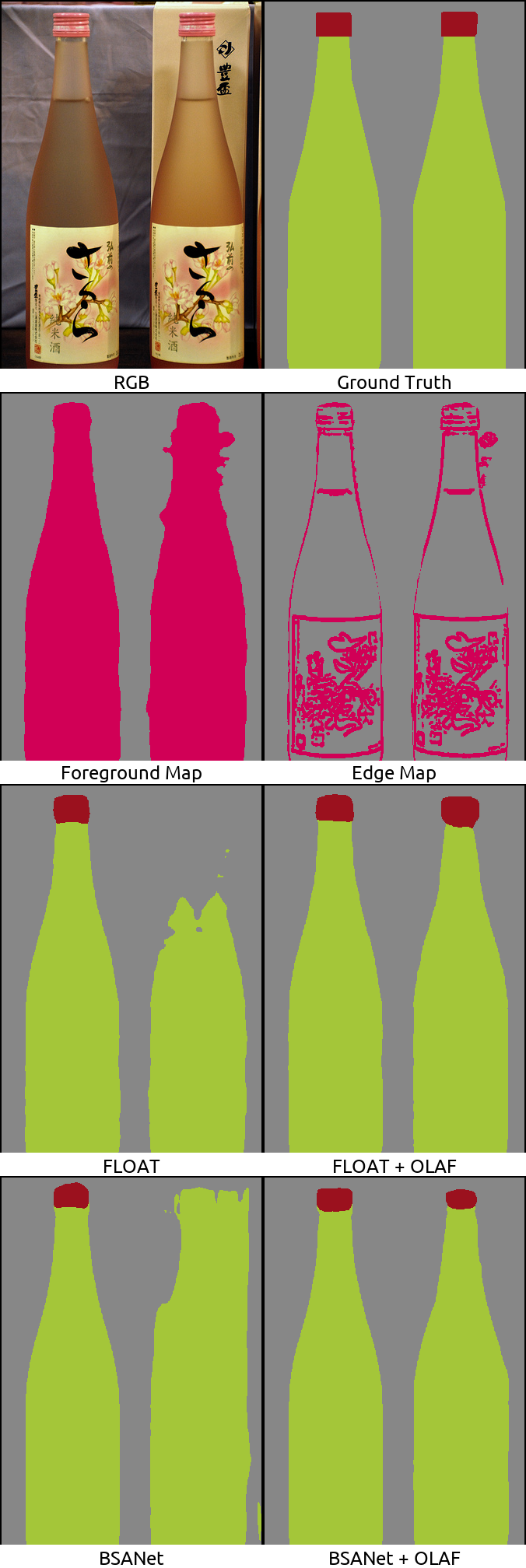}
  \caption{}
\end{figure}

\clearpage

\subsection{Pascal-Part-108}
\begin{figure}[H]
  \centering
  \includegraphics[width=0.9\textwidth, height=0.9\textheight]{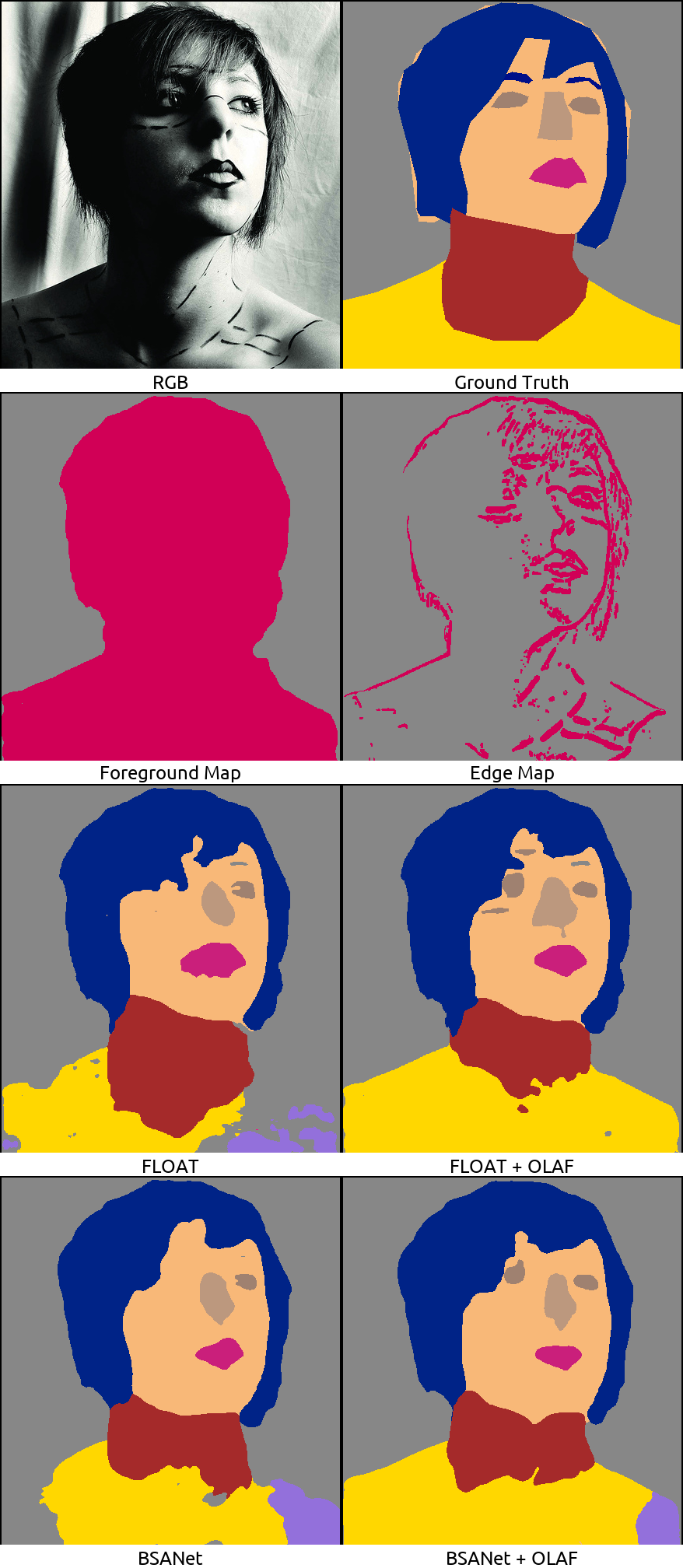}
  \caption{}
\end{figure}

\clearpage

\begin{figure}
  \centering
  \includegraphics[width=\textwidth, height=\textheight]{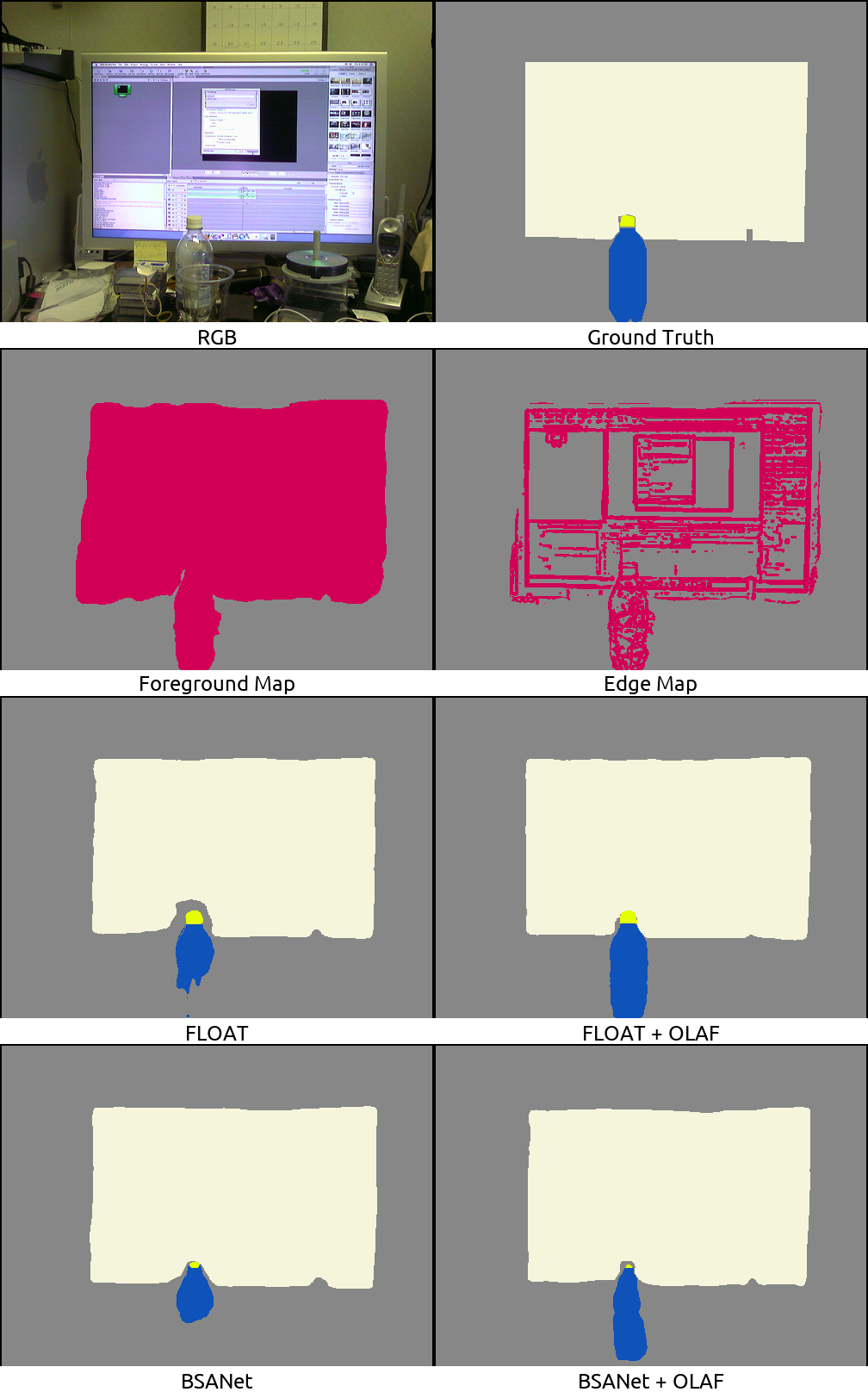}
  \caption{}
\end{figure}

\clearpage

\begin{figure}
  \centering
  \includegraphics[width=\textwidth, height=\textheight]{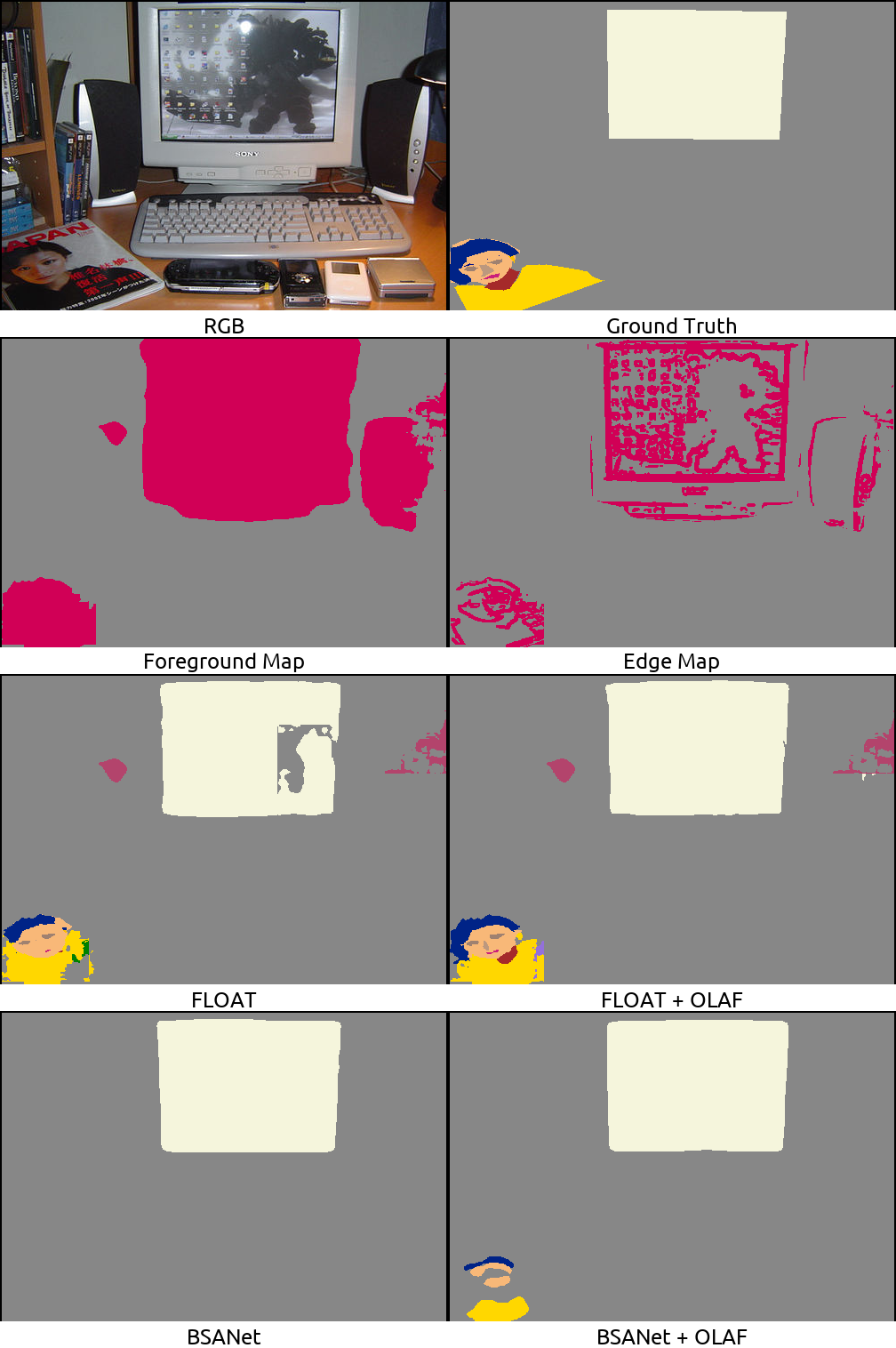}
  \caption{}
\end{figure}

\clearpage

\begin{figure}
  \centering
  \includegraphics[width=\textwidth, height=\textheight]{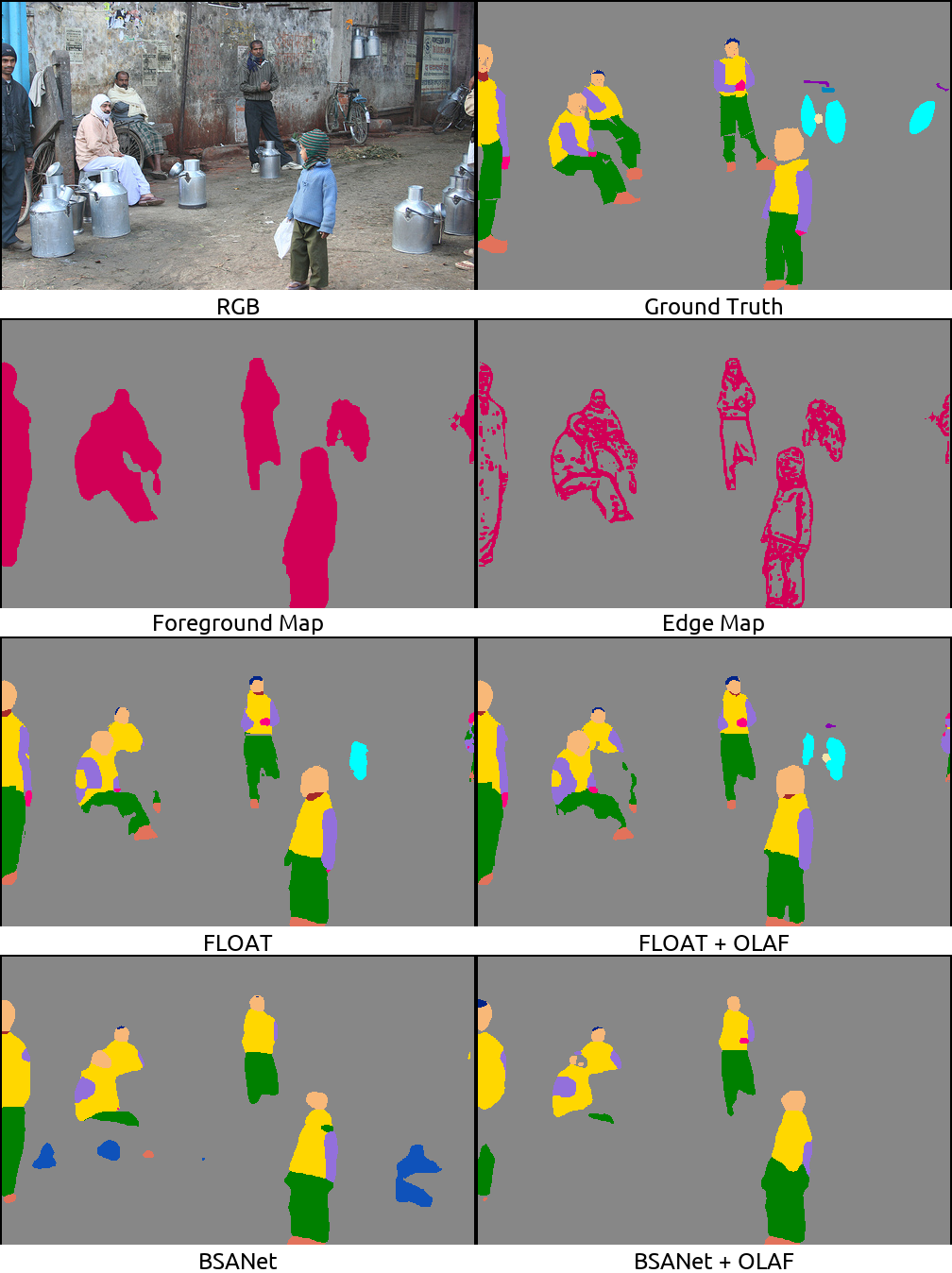}
  \caption{}
\end{figure}

\clearpage

\subsection{Pascal-Part-201}
\begin{figure}[H]
  \centering
  \includegraphics[width=0.9\textwidth, height=0.9\textheight]{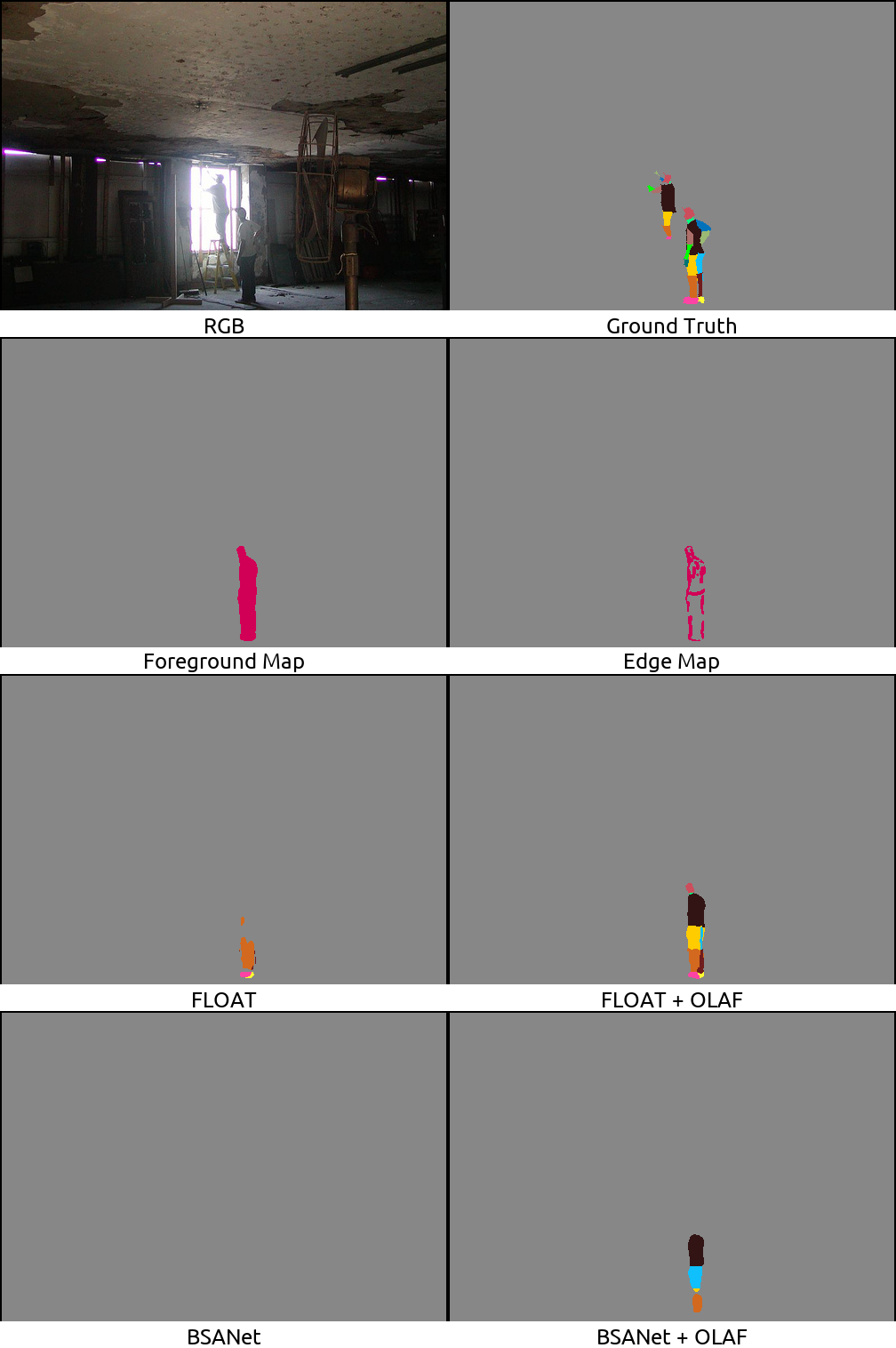}
  \caption{}
\end{figure}

\clearpage

\begin{figure}
  \centering
  \includegraphics[width=\textwidth, height=\textheight]{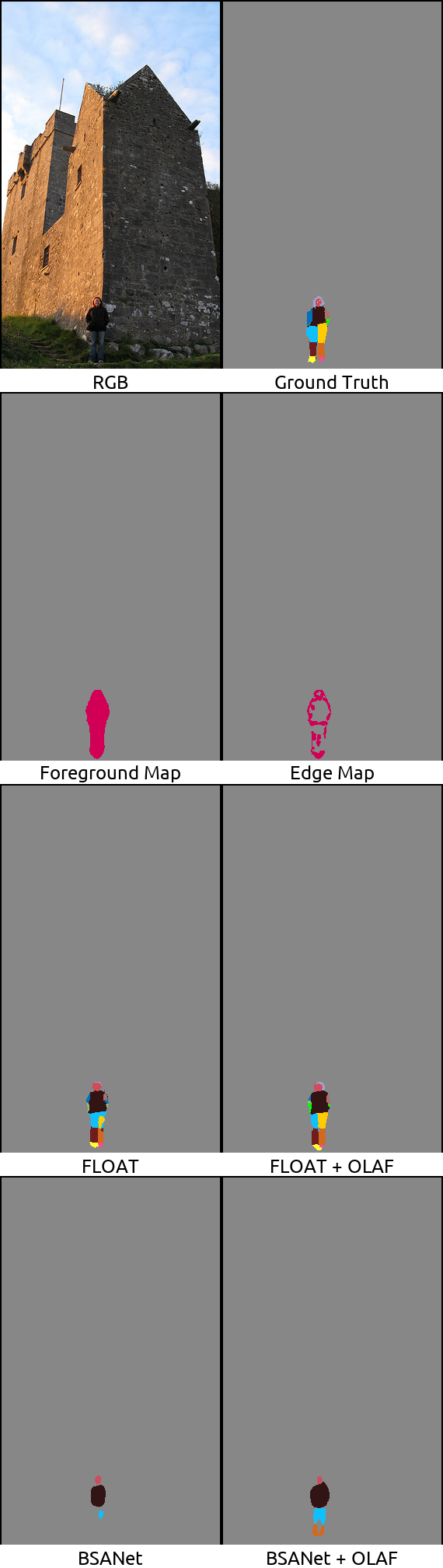}
  \caption{}
\end{figure}

\clearpage

\begin{figure}
  \centering
  \includegraphics[width=0.9\textwidth, height=0.9\textheight]{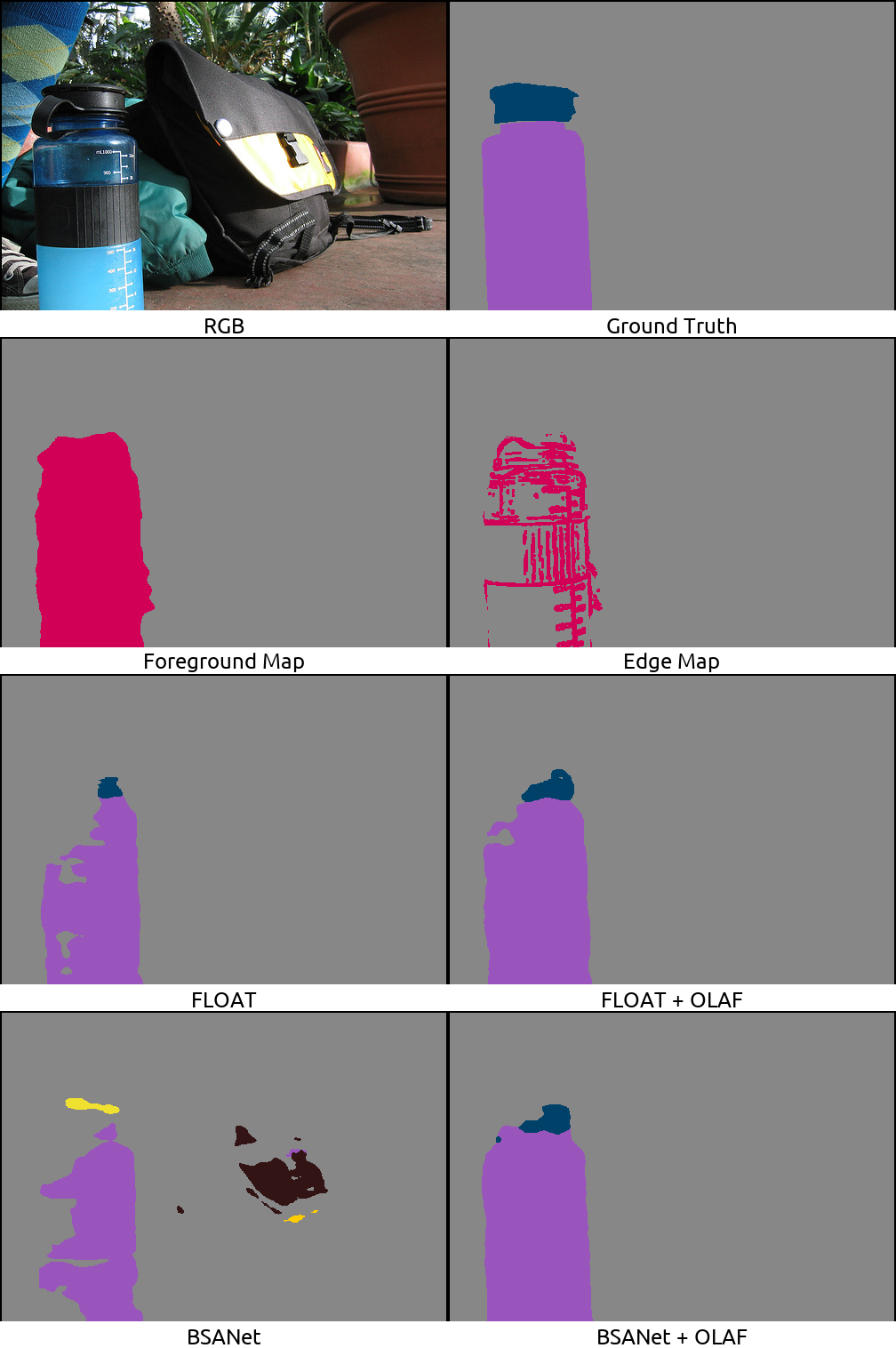}
  \caption{}
\end{figure}
\clearpage

\section{Quantitative Results}
\subsection{Pascal-Part-58 | mIoU}

\begin{table}[H]
\resizebox{\textwidth}{!}{  
\begin{tabular}{lllllllllll}
\toprule
Category & deeplabv3 & deeplabv3+O & BSANet & BSANet+O & GMNet & GMNet+O & FLOAT & FLOAT+O & FLOAT\textsuperscript{\textdagger} & FLOAT\textsuperscript{\textdagger}+O \\
\midrule

Background & 90.1 & 93.6 & 91.6 & 92.1 & 92.7 & 93.3 & 92.9 & 93.6 & 93.1 & 94.3\\
Aeroplane Body & 65.3 & 69.8 & 70.0 & 71.1 & 69.6 & 71.2 & 70.2 & 72.6 & 72.1 & 75.8\\
Aeroplane Engine & 24.9 & 28.4 & 29.1 & 29.5 & 25.7 & 27.1 & 29.7 & 31.4 & 30.5 & 32.5\\
Aeroplane Wing & 33.9 & 36.2 & 38.3 & 38.7 & 34.2 & 35.6 & 37.6 & 40.4 & 39.5 & 46.2\\
Aeroplane Stern & 56.3 & 56.7 & 59.2 & 59.9 & 57.2 & 58.1 & 58.3 & 59.6 & 59.2 & 60.8\\
Aeroplane Wheel & 43.8 & 57.7 & 53.2 & 53.9 & 46.8 & 48.1 & 45.7 & 56.0 & 55.5 & 55.1\\
Bicycle Wheel & 77.8 & 78.3 & 78.0 & 79.1 & 81.3 & 81.4 & 81.2 & 81.6 & 81.2 & 82.9\\
Bicycle Body & 48.4 & 53.1 & 53.4 & 54.4 & 51.5 & 52.9 & 53.0 & 55.6 & 55.1 & 58.2\\
Bird Head & 64.6 & 72.7 & 74.0 & 74.7 & 71.1 & 72.2 & 73.7 & 74.7 & 74.4 & 75.2\\
Bird Wing & 34.1 & 40.5 & 39.7 & 39.9 & 38.6 & 37.2 & 41.6 & 38.0 & 37.0 & 39.2\\
Bird Leg & 28.9 & 33.7 & 34.8 & 34.9 & 28.7 & 29.3 & 30.1 & 32.5 & 31.5 & 33.9\\
Bird Torso & 65.5 & 68.9 & 70.9 & 71.2 & 69.5 & 70.1 & 69.2 & 70.4 & 69.4 & 71.3\\
Boat & 54.4 & 71.3 & 60.2 & 69.9 & 70.0 & 73.2 & 75.3 & 76.7 & 76.3 & 77.9\\
Bottle Cap & 32.7 & 36.0 & 29.8 & 32.7 & 33.9 & 35.2 & 31.9 & 37.6 & 37.2 & 38.1\\
Bottle Body & 68.8 & 76.3 & 68.6 & 74.8 & 77.6 & 78.4 & 73.8 & 79.2 & 78.3 & 80.3\\
Bus Window & 72.7 & 75.6 & 74.8 & 76.7 & 75.4 & 76.3 & 76.7 & 78.6 & 78.2 & 79.1\\
Bus Wheel & 55.3 & 58.3 & 57.1 & 59.2 & 58.1 & 59.6 & 61.0 & 61.0 & 60.1 & 65.2\\
Bus Body & 74.8 & 77.6 & 78.3 & 79.3 & 79.9 & 81.8 & 80.3 & 82.0 & 81.3 & 84.1\\
Car Window & 63.6 & 68.5 & 68.1 & 68.8 & 64.8 & 66.6 & 70.7 & 70.8 & 70.1 & 70.1\\
Car Wheel & 64.8 & 69.9 & 68.5 & 70.2 & 70.3 & 71.2 & 73.7 & 74.2 & 73.3 & 76.9\\
Car Light & 46.2 & 55.1 & 53.7 & 56.2 & 48.4 & 50.5 & 54.6 & 60.2 & 59.5 & 60.3\\
Car Plate & 0.0 & 0.0 & 0.0 & 0.0 & 0.0 & 0.0 & 0.0 & 0.0 & 0.0 & 0.0\\
Car Body & 72.1 & 77.2 & 77.0 & 77.3 & 77.6 & 78.8 & 79.0 & 79.9 & 79.5 & 80.9\\
Cat Head & 80.2 & 81.4 & 83.7 & 84.2 & 83.8 & 83.9 & 85.7 & 85.3 & 84.9 & 86.8\\
Cat Leg & 48.6 & 50.1 & 50.1 & 50.2 & 49.4 & 50.2 & 51.7 & 51.2 & 50.7 & 54.2\\
Cat Tail & 41.2 & 43.0 & 48.8 & 46.2 & 46.0 & 45.2 & 45.5 & 46.3 & 45.8 & 49.2\\
Cat Torso & 70.3 & 70.0 & 72.6 & 72.9 & 73.8 & 74.2 & 73.6 & 74.1 & 73.6 & 75.9\\
Chair & 35.4 & 42.9 & 36.5 & 49.2 & 51.4 & 55.3 & 59.6 & 59.7 & 59.0 & 61.6\\
Cow Head & 74.3 & 75.4 & 76.4 & 77.1 & 80.7 & 80.1 & 80.9 & 80.1 & 79.4 & 83.2\\
Cow Tail & 0.0 & 0.0 & 7.9 & 10.2 & 8.1 & 14.2 & 18.3 & 23.3 & 22.6 & 25.8\\
Cow Leg & 46.1 & 53.0 & 53.4 & 54.7 & 53.5 & 53.8 & 57.0 & 56.0 & 55.2 & 58.2\\
Cow Torso & 67.9 & 72.5 & 73.5 & 73.9 & 77.1 & 76.5 & 76.7 & 77.4 & 76.2 & 79.2\\
Dining Table & 43.0 & 52.2 & 43.7 & 49.2 & 51.3 & 56.2 & 58.2 & 58.3 & 57.4 & 60.2\\
Dog Head & 78.7 & 79.3 & 82.5 & 82.7 & 85.0 & 85.2 & 84.3 & 84.8 & 84.6 & 86.4\\
Dog Leg & 48.1 & 50.5 & 53.8 & 53.1 & 53.8 & 53.9 & 53.8 & 54.4 & 53.4 & 55.8\\
Dog Tail & 27.1 & 30.5 & 31.3 & 32.5 & 31.4 & 32.0 & 37.3 & 35.3 & 34.5 & 38.2\\
Dog Torso & 63.6 & 63.6 & 65.7 & 65.9 & 68.0 & 68.6 & 67.3 & 68.2 & 67.8 & 69.2\\
Horse Head & 74.7 & 72.8 & 76.6 & 75.2 & 73.9 & 73.0 & 81.6 & 81.3 & 80.3 & 84.3\\
Horse Tail & 47.0 & 48.4 & 51.0 & 50.6 & 50.4 & 50.9 & 52.2 & 51.7 & 51.6 & 57.9\\
Horse Leg & 55.2 & 60.5 & 61.6 & 61.8 & 59.3 & 61.2 & 60.3 & 61.7 & 61.4 & 61.9\\
Horse Torso & 71.3 & 76.2 & 74.9 & 76.4 & 73.9 & 75.2 & 77.2 & 78.1 & 77.6 & 78.8\\
Motorbike Wheel & 72.9 & 71.9 & 71.6 & 72.2 & 73.5 & 74.4 & 76.3 & 76.9 & 76.6 & 76.3\\
Motorbike Body & 64.1 & 69.7 & 71.5 & 71.8 & 74.3 & 75.2 & 75.0 & 76.2 & 75.8 & 76.9\\
Person Head & 82.5 & 84.7 & 85.0 & 86.2 & 84.7 & 84.8 & 84.2 & 84.4 & 83.5 & 86.3\\
Person Torso & 65.3 & 68.3 & 68.2 & 68.9 & 67.0 & 67.3 & 68.6 & 69.4 & 68.7 & 70.1\\
Person Lower Arm & 46.9 & 50.6 & 52.0 & 52.1 & 48.6 & 48.9 & 51.8 & 52.3 & 51.5 & 50.3\\
Person Upper Arm & 51.5 & 52.3 & 54.4 & 54.1 & 52.4 & 55.1 & 54.6 & 55.7 & 55.3 & 54.2\\
Person Lower Leg & 38.6 & 41.7 & 43.5 & 43.4 & 40.2 & 41.2 & 42.4 & 42.9 & 42.5 & 43.2\\
Person Upper Leg & 43.8 & 46.3 & 47.4 & 47.1 & 44.5 & 46.3 & 47.4 & 48.4 & 47.9 & 49.2\\
Potted Plant Pot & 47.3 & 57.4 & 53.5 & 57.8 & 56.0 & 58.4 & 50.8 & 60.0 & 59.7 & 66.2\\
Potted Plant Plant & 52.4 & 65.1 & 56.6 & 59.2 & 56.4 & 59.9 & 58.9 & 65.1 & 64.6 & 66.4\\
Sheep Head & 60.9 & 64.3 & 65.4 & 68.3 & 70.8 & 71.1 & 70.6 & 72.2 & 71.8 & 72.9\\
Sheep Leg & 8.6 & 20.0 & 11.7 & 16.2 & 14.3 & 17.2 & 24.4 & 26.9 & 26.1 & 28.3\\
Sheep Torso & 68.3 & 70.8 & 71.6 & 72.4 & 75.6 & 75.3 & 76.0 & 76.7 & 76.0 & 77.8\\
Sofa & 43.2 & 55.6 & 43.1 & 48.0 & 56.1 & 60.5 & 69.1 & 69.1 & 68.4 & 70.8\\
Train & 76.6 & 85.7 & 82.2 & 85.7 & 85.0 & 85.1 & 86.0 & 87.6 & 87.2 & 88.9\\
TV Screen & 69.5 & 80.6 & 73.1 & 77.3 & 77.0 & 78.4 & 72.0 & 81.9 & 81.5 & 85.3\\
TV Frame & 44.4 & 57.7 & 49.8 & 54.6 & 54.1 & 54.3 & 47.3 & 57.9 & 57.1 & 59.2\\

\bottomrule

\end{tabular}
}
\end{table}

\clearpage

\newpage

\subsection{Pascal-Part-58 | $miou_{small}$}
\begin{table}[h]
\begin{tabular}{llllllllll}
\toprule
Object & FLOAT & FLOAT+O \\
\midrule
background &    - &    - \\
  aeroplane & 32.1 & 33.5 \\
    bicycle &   30 & 33.2 \\
       bird & 45.3 & 47.6 \\
       boat & 19.4 & 21.2 \\
     bottle & 41.7 & 45.9 \\
        bus & 43.8 & 47.1 \\
        car & 37.5 & 38.6 \\
        cat & 42.1 & 45.3 \\
      chair & 39.2 & 42.8 \\
        cow & 25.3 & 28.2 \\
diningtable & 32.6 & 36.1 \\
        dog & 29.5 & 32.7 \\
      horse & 41.7 & 45.2 \\
  motorbike & 29.3 & 32.9 \\
     person & 41.3 & 43.2 \\
pottedplant & 54.6 & 55.9 \\
      sheep & 52.8 & 55.5 \\
       sofa & 49.5 & 53.8 \\
      train & 21.1 & 24.4 \\
  tvmonitor & 52.2 & 55.6 \\

\bottomrule
\end{tabular}
\end{table}

\newpage
\subsection{Pascal-Part-108 | mIoU}

\begin{table}[h]
\resizebox{\textwidth}{!}{  

\begin{tabular}{lllllllllll}
\toprule
Category & deeplabv3 & deeplabv3+O & BSANet & BSANet+O & GMNet & GMNet+O & FLOAT & FLOAT+O & FLOAT\textsuperscript{\textdagger} & FLOAT\textsuperscript{\textdagger}+O\\
\midrule
Background & 90.2 & 93.8 & 91.8 & 92.3 & 92.7 & 93.1 & 92.9 & 93.6 & 92.1 & 94.5\\
Aeroplane Body & 60.9 & 70.2 & 69.6 & 71.2 & 61.9 & 66.4 & 70.0 & 71.6 & 70.1 & 72.5\\
Aeroplane Engine & 53.2 & 57.2 & 56.2 & 58.2 & 27.2 & 36.9 & 57.6 & 59.4 & 58.3 & 60.8\\
Aeroplane Wing & 27.9 & 35.8 & 37.2 & 38.4 & 34.3 & 36.2 & 34.3 & 38.5 & 37.4 & 39.6\\
Aeroplane Stern & 24.7 & 29.5 & 29.4 & 29.3 & 57.4 & 58.1 & 27.6 & 30.5 & 29.3 & 31.8\\
Aeroplane Wheel & 40.9 & 56.9 & 51.0 & 52.2 & 51.5 & 52.2 & 45.5 & 45.6 & 44.5 & 46.6\\
Bicycle Wheel & 76.4 & 78.0 & 77.1 & 78.4 & 80.2 & 80.8 & 81.2 & 81.7 & 80.2 & 82.7\\
Bicycle Saddle & 34.1 & 42.2 & 38.3 & 40.2 & 38.0 & 40.2 & 36.9 & 42.2 & 40.6 & 43.7\\
Bicycle Handlebar & 23.3 & 33.1 & 25.2 & 27.7 & 22.4 & 27.1 & 18.8 & 31.8 & 30.5 & 33.4\\
Bicycle Chainwheel & 42.3 & 46.1 & 41.6 & 42.2 & 44.1 & 43.5 & 53.6 & 44.1 & 42.5 & 45.6\\
Bird Head & 51.5 & 61.0 & 66.6 & 67.8 & 65.3 & 67.7 & 70.0 & 70.4 & 69.1 & 71.7\\
Bird Beak & 40.4 & 49.6 & 51.3 & 51.7 & 44.3 & 46.5 & 56.0 & 52.3 & 50.5 & 53.6\\
Bird Torso & 61.7 & 63.4 & 67.3 & 68.3 & 64.8 & 64.8 & 63.4 & 65.6 & 64.5 & 66.9\\
Bird Neck & 27.5 & 34.9 & 34.5 & 35.2 & 28.4 & 29.7 & 34.3 & 35.9 & 34.6 & 37.4\\
Bird Wing & 35.9 & 37.0 & 41.3 & 42.1 & 37.2 & 39.3 & 36.2 & 40.3 & 38.7 & 41.6\\
Bird Leg & 23.5 & 26.3 & 30.8 & 30.9 & 23.8 & 25.1 & 25.5 & 29.6 & 28.4 & 30.9\\
Bird Foot & 13.9 & 15.8 & 18.3 & 19.2 & 17.7 & 17.4 & 17.4 & 18.3 & 16.5 & 19.7\\
Bird Tail & 28.1 & 35.8 & 35.7 & 35.9 & 32.5 & 33.6 & 33.6 & 35.6 & 34.6 & 36.4\\
Boat & 53.7 & 71.8 & 60.7 & 68.2 & 69.2 & 72.2 & 74.8 & 76.7 & 75.7 & 77.6\\
Bottle Cap & 30.4 & 34.8 & 31.0 & 35.2 & 33.4 & 36.3 & 37.6 & 37.1 & 35.3 & 38.2\\
Bottle Body & 63.7 & 74.5 & 71.4 & 74.9 & 78.7 & 79.2 & 72.4 & 79.3 & 77.6 & 80.2\\
Bus Side & 70.1 & 76.0 & 74.9 & 76.3 & 75.7 & 76.6 & 77.1 & 78.6 & 77.8 & 79.6\\
Bus Roof & 7.5 & 10.7 & 6.3 & 8.2 & 13.5 & 14.2 & 17.1 & 15.8 & 14.9 & 17.4\\
Bus Mirror & 2.1 & 7.9 & 8.0 & 10.2 & 6.6 & 7.9 & 0.0 & 13.4 & 12.3 & 14.8\\
Bus Plate & 0.0 & 0.0 & 0.0 & 0.0 & 0.0 & 0.0 & 0.0 & 0.0 & 0.0 & 0.0\\
Bus Door & 40.1 & 42.4 & 47.4 & 44.4 & 38.1 & 40.2 & 40.6 & 41.9 & 40.9 & 43.4\\
Bus Wheel & 53.8 & 58.0 & 56.6 & 59.1 & 56.7 & 58.4 & 59.8 & 61.6 & 60.1 & 62.6\\
Bus Headlight & 25.6 & 37.3 & 31.0 & 34.3 & 30.4 & 36.0 & 44.7 & 49.7 & 48.8 & 50.5\\
Bus Window & 71.8 & 76.6 & 74.7 & 75.1 & 74.6 & 76.9 & 75.7 & 77.7 & 76.4 & 78.8\\
Car Side & 64.0 & 69.4 & 70.2 & 71.2 & 70.5 & 71.9 & 70.8 & 73.0 & 71.7 & 74.6\\
Car Roof & 21.0 & 23.7 & 22.2 & 22.9 & 22.5 & 25.8 & 27.1 & 32.5 & 31.4 & 33.4\\
Car Plate & 0.0 & 0.0 & 0.0 & 0.0 & 0.0 & 0.0 & 0.0 & 0.0 & 0.0 & 0.0\\
Car Door & 40.1 & 43.0 & 42.1 & 43.2 & 42.3 & 43.0 & 46.8 & 43.4 & 42.4 & 44.8\\
Car Wheel & 65.8 & 70.2 & 68.8 & 69.3 & 70.2 & 72.2 & 72.7 & 74.1 & 72.6 & 75.6\\
Car Headlight & 42.9 & 48.0 & 53.8 & 55.5 & 46.4 & 46.9 & 52.5 & 58.4 & 57.2 & 59.8\\
Car Window & 61.0 & 67.8 & 69.1 & 69.1 & 65.0 & 69.4 & 69.4 & 70.6 & 69.6 & 71.4\\
Cat Head & 73.9 & 75.8 & 76.7 & 76.9 & 77.5 & 78.1 & 77.6 & 78.8 & 77.5 & 80.5\\
Cat Eye & 58.8 & 66.9 & 64.8 & 66.1 & 62.8 & 67.7 & 67.8 & 70.7 & 69.6 & 71.7\\
Cat Ear & 65.5 & 67.8 & 67.9 & 68.6 & 67.1 & 68.2 & 67.8 & 69.6 & 68.7 & 70.8\\
Cat Nose & 39.1 & 43.8 & 46.6 & 46.9 & 46.3 & 46.9 & 44.7 & 47.8 & 46.1 & 49.4\\
Cat Torso & 64.2 & 65.1 & 66.9 & 67.3 & 68.7 & 58.8 & 68.0 & 68.9 & 67.3 & 70.0\\
Cat Neck & 22.8 & 23.6 & 22.4 & 24.2 & 24.4 & 25.6 & 26.2 & 28.3 & 26.6 & 29.0\\
Cat Leg & 36.5 & 37.9 & 39.6 & 39.5 & 39.1 & 39.1 & 40.3 & 39.2 & 37.3 & 40.2\\
Cat Paw & 40.6 & 39.5 & 42.0 & 42.3 & 41.7 & 41.4 & 43.2 & 41.9 & 40.4 & 43.5\\
Cat Tail & 40.2 & 42.9 & 44.5 & 45.1 & 45.8 & 45.9 & 43.9 & 46.9 & 45.3 & 48.6\\
Chair & 35.4 & 45.6 & 35.7 & 39.9 & 49.1 & 54.2 & 59.4 & 59.7 & 58.6 & 60.4\\
Cow Head & 51.2 & 61.0 & 62.5 & 62.8 & 63.8 & 63.1 & 64.7 & 63.5 & 62.4 & 64.6\\
Cow Ear & 51.2 & 58.3 & 57.8 & 58.6 & 60.0 & 60.5 & 60.9 & 62.0 & 60.7 & 63.3\\
Cow Muzzle & 61.2 & 68.1 & 72.4 & 72.8 & 74.9 & 74.5 & 70.6 & 74.3 & 72.5 & 75.6\\
Cow Horn & 28.8 & 33.3 & 45.5 & 45.2 & 44.0 & 45.7 & 34.9 & 38.4 & 37.6 & 39.2\\
Cow Torso & 63.4 & 72.2 & 73.5 & 73.2 & 73.2 & 73.4 & 72.6 & 73.6 & 72.5 & 74.4\\
Cow Neck & 9.5 & 16.8 & 15.9 & 18.8 & 20.3 & 23.3 & 26.8 & 29.5 & 28.4 & 30.5\\
Cow Leg & 46.5 & 56.1 & 54.8 & 55.1 & 54.8 & 55.6 & 54.8 & 56.0 & 54.2 & 57.7\\
Cow Tail & 6.5 & 2.6 & 3.1 & 8.8 & 13.6 & 16.8 & 22.9 & 31.1 & 29.8 & 32.6\\
Dining Table & 33.0 & 52.7 & 45.6 & 48.3 & 50.6 & 53.3 & 58.0 & 58.2 & 56.3 & 59.7\\
Dog Head & 60.5 & 60.8 & 64.7 & 64.2 & 64.0 & 63.5 & 63.3 & 63.3 & 61.7 & 64.5\\
Dog Eye & 50.1 & 58.4 & 57.0 & 59.3 & 54.7 & 57.9 & 60.9 & 63.6 & 62.4 & 64.6\\
Dog Ear & 52.0 & 55.0 & 57.8 & 57.9 & 56.8 & 56.9 & 57.4 & 58.0 & 56.7 & 59.7\\
Dog Nose & 63.5 & 66.9 & 69.8 & 70.1 & 66.0 & 67.7 & 66.7 & 70.1 & 68.8 & 71.5\\
Dog Torso & 58.4 & 58.4 & 62.3 & 62.8 & 63.2 & 63.1 & 62.2 & 63.7 & 62.2 & 64.7\\
Dog Neck & 27.1 & 27.0 & 28.0 & 28.3 & 28.1 & 28.3 & 26.5 & 28.1 & 26.6 & 29.3\\
Dog Leg & 39.2 & 41.9 & 43.2 & 44.1 & 43.7 & 43.9 & 43.1 & 44.2 & 42.5 & 45.5\\
Dog Paw & 39.4 & 44.5 & 45.2 & 47.2 & 43.7 & 45.6 & 47.8 & 49.4 & 48.1 & 50.7\\
Dog Tail & 24.7 & 31.2 & 35.0 & 35.1 & 30.8 & 32.2 & 31.0 & 36.0 & 34.3 & 37.4\\
Dog Muzzle & 65.1 & 66.3 & 70.1 & 70.0 & 68.9 & 68.1 & 67.0 & 68.9 & 67.4 & 70.5\\

\bottomrule
\end{tabular}
}
\end{table}

\clearpage

\begin{table}[h]
\resizebox{\textwidth}{!}{  

\begin{tabular}{lllllllllll}
\toprule
Category & deeplabv3 & deeplabv3+O & BSANet & BSANet+O & GMNet & GMNet+O & FLOAT & FLOAT+O & FLOAT\textsuperscript{\textdagger} & FLOAT\textsuperscript{\textdagger}+O\\
\midrule

Horse Head & 54.4 & 62.9 & 59.9 & 60.7 & 55.9 & 59.9 & 62.4 & 64.6 & 63.5 & 65.9\\
Horse Ear & 49.7 & 58.7 & 56.8 & 58.1 & 52.2 & 55.7 & 59.2 & 61.1 & 59.7 & 62.5\\
Horse Muzzle & 61.3 & 69.0 & 66.6 & 68.4 & 62.9 & 64.2 & 65.3 & 69.5 & 68.6 & 70.7\\
Horse Torso & 56.7 & 63.3 & 61.1 & 62.6 & 60.7 & 61.1 & 63.1 & 64.8 & 63.7 & 66.5\\
Horse Neck & 42.1 & 51.7 & 44.8 & 46.5 & 47.2 & 48.6 & 49.3 & 53.3 & 51.9 & 54.8\\
Horse Leg & 54.1 & 60.5 & 59.3 & 60.1 & 56.4 & 57.3 & 58.0 & 60.0 & 58.8 & 61.8\\
Horse Tail & 48.1 & 51.2 & 51.9 & 52.1 & 51.4 & 52.2 & 53.4 & 55.2 & 53.6 & 56.5\\
Horse Hoof & 22.1 & 21.6 & 19.8 & 21.2 & 25.3 & 25.4 & 18.2 & 25.2 & 23.5 & 26.9\\
Motorbike Wheel & 69.6 & 71.3 & 71.6 & 73.2 & 73.6 & 75.7 & 76.4 & 76.8 & 75.6 & 78.2\\
Motorbike Hbar & 0.0 & 0.0 & 0.0 & 0.0 & 0.0 & 0.0 & 0.0 & 0.7 & 0.5 & 1.4\\
Motorbike Saddle & 0.0 & 0.0 & 0.0 & 0.0 & 0.8 & 0.5 & 0.1 & 0.0 & 0.0 & 0.0\\
Motorbike Hlight & 25.8 & 28.1 & 21.3 & 29.0 & 28.5 & 31.1 & 35.0 & 42.9 & 41.7 & 44.1\\
Person Head & 68.2 & 71.5 & 71.3 & 71.9 & 69.3 & 70.6 & 72.6 & 73.6 & 72.3 & 74.2\\
Person Eye & 35.1 & 50.4 & 44.6 & 48.2 & 38.7 & 42.4 & 49.6 & 52.7 & 51.8 & 53.3\\
Person Ear & 37.4 & 49.4 & 46.4 & 47.3 & 41.4 & 45.8 & 47.6 & 52.7 & 51.9 & 53.4\\
Person Nose & 53.0 & 61.9 & 57.4 & 59.1 & 56.7 & 59.5 & 62.4 & 65.2 & 63.5 & 66.2\\
Person Mouth & 48.9 & 56.7 & 53.1 & 55.2 & 51.3 & 55.2 & 58.4 & 61.2 & 59.4 & 62.3\\
Person Hair & 70.8 & 72.1 & 73.2 & 74.3 & 71.8 & 71.1 & 70.9 & 71.6 & 70.1 & 72.4\\
Person Torso & 63.4 & 66.2 & 66.3 & 66.9 & 65.2 & 66.3 & 66.1 & 67.5 & 66.6 & 68.5\\
Person Neck & 49.7 & 52.6 & 53.1 & 53.4 & 51.2 & 52.7 & 54.5 & 54.4 & 53.5 & 55.2\\
Person Arm & 54.7 & 58.7 & 58.4 & 59.5 & 57.4 & 58.6 & 58.3 & 59.5 & 58.6 & 60.6\\
Person Hand & 43.0 & 48.3 & 50.1 & 50.2 & 44.1 & 45.8 & 47.8 & 49.3 & 47.1 & 50.7\\
Person Leg & 50.8 & 52.2 & 53.8 & 53.9 & 53.0 & 53.9 & 53.6 & 55.5 & 54.6 & 56.3\\
Person Foot & 29.8 & 33.1 & 33.0 & 31.1 & 31.3 & 30.7 & 31.8 & 31.9 & 29.4 & 33.3\\
Potted Plant Pot & 41.6 & 58.6 & 52.3 & 55.2 & 56.0 & 58.2 & 50.1 & 59.8 & 58.8 & 61.2\\
Potted Plant Plant & 42.9 & 62.6 & 56.1 & 58.3 & 56.6 & 60.8 & 47.7 & 66.4 & 65.3 & 67.3\\
Sheep Head & 45.6 & 45.1 & 50.2 & 49.2 & 54.0 & 53.3 & 51.6 & 51.5 & 50.5 & 52.7\\
Sheep Ear & 43.2 & 44.7 & 48.9 & 49.9 & 45.3 & 47.1 & 54.8 & 56.2 & 54.5 & 57.3\\
Sheep Muzzle & 58.2 & 57.5 & 66.6 & 67.2 & 64.9 & 65.0 & 65.5 & 65.3 & 63.6 & 66.2\\
Sheep Horn & 3.0 & 0.8 & 5.1 & 9.5 & 5.4 & 11.6 & 31.8 & 53.0 & 51.6 & 54.5\\
Sheep Torso & 62.6 & 65.7 & 66.3 & 68.2 & 68.8 & 69.4 & 69.9 & 70.9 & 69.7 & 72.5\\
Sheep Neck & 26.9 & 27.9 & 29.8 & 29.9 & 30.3 & 31.5 & 36.0 & 35.2 & 33.6 & 36.4\\
Sheep Leg & 8.6 & 17.9 & 21.1 & 22.4 & 11.7 & 19.3 & 23.9 & 25.9 & 24.3 & 27.4\\
Sheep Tail & 6.7 & 7.1 & 6.3 & 6.9 & 9.1 & 10.8 & 15.2 & 17.8 & 16.4 & 19.2\\
Sofa & 39.2 & 56.6 & 43.0 & 45.2 & 53.9 & 58.2 & 68.9 & 69.1 & 67.5 & 70.1\\
Train Head & 5.3 & 3.9 & 6.0 & 4.2 & 4.5 & 4.1 & 4.0 & 4.8 & 3.4 & 6.2\\
Train Head Side & 61.9 & 60.9 & 60.8 & 61.2 & 60.8 & 61.4 & 66.6 & 64.4 & 62.2 & 65.4\\
Train Head Roof & 23.0 & 25.3 & 19.9 & 20.1 & 21.1 & 20.7 & 26.5 & 20.1 & 18.4 & 21.5\\
Train Headlight & 0.0 & 0.0 & 0.0 & 0.0 & 0.0 & 0.0 & 0.0 & 0.0 & 0.0 & 0.0\\
Train Coach & 28.6 & 35.8 & 35.7 & 36.6 & 31.4 & 32.8 & 36.4 & 37.0 & 35.1 & 38.8\\
Train Coach Side & 15.6 & 16.6 & 18.4 & 19.2 & 14.9 & 14.2 & 15.5 & 16.1 & 14.4 & 17.7\\
Train Coach Roof & 10.8 & 18.4 & 6.3 & 7.9 & 18.1 & 17.6 & 7.7 & 20.9 & 19.5 & 22.8\\
TV Screen & 64.8 & 80.5 & 70.4 & 72.1 & 70.7 & 71.4 & 69.6 & 70.4 & 69.2 & 71.9\\

\bottomrule
\end{tabular}
}
\end{table}

\clearpage

\subsection{Pascal-Part-108 | $miou_{small}$}
\vspace{-0.3cm}
\begin{table}[h]
\begin{tabular}{llllllllll}
\toprule
Object & FLOAT & FLOAT+O \\
\midrule
background &    - &    - \\
  aeroplane & 29.1 &   32 \\
    bicycle & 23.5 & 25.2 \\
       bird & 40.4 & 41.1 \\
       boat & 15.8 & 16.9 \\
     bottle & 38.3 & 42.6 \\
        bus & 39.6 & 41.8 \\
        car & 34.8 & 36.9 \\
        cat & 39.1 & 42.6 \\
      chair & 39.2 & 42.4 \\
        cow & 25.1 & 29.6 \\
diningtable & 30.7 & 34.3 \\
        dog & 29.1 & 33.6 \\
      horse & 40.1 & 42.7 \\
  motorbike & 25.6 & 27.2 \\
     person & 38.1 & 42.4 \\
pottedplant & 50.1 & 51.5 \\
      sheep & 45.7 & 47.3 \\
       sofa & 47.3 & 49.6 \\
      train & 20.5 & 22.8 \\
  tvmonitor & 50.1 & 53.2 \\
  
\bottomrule
\end{tabular}
\end{table}

\clearpage
\subsection{Pascal-Part-201 | mIoU}
\begin{table}[h]
\resizebox{\textwidth}{!}{  

\begin{tabular}{llllllllllll}
\toprule
Object & Part & deeplabv3 & deeplabv3+O & BSANet & BSANet+O & GMNet & GMNet+O & FLOAT & FLOAT+O & FLOAT\textsuperscript{\textdagger} & FLOAT\textsuperscript{\textdagger}+O\\
\midrule
Background & nan & 91.0 & 93.6 & 91.2 & 92.9 & 90.8 & 93.1 & 92.5 & 93.3 & 91.2 & 94.2\\
\midrule
Aeroplane & Body & 67.3 & 70.1 & 71.1 & 72.2 & 61.9 & 64.0 & 68.9 & 70.5 & 68.3 & 71.7\\
 & Engine & 27.0 & 27.3 & 30.0 & 30.3 & 19.3 & 21.6 & 28.5 & 28.0 & 25.3 & 29.0\\
 & Left Wing & 3.8 & 18.9 & 10.7 & 14.6 & 3.4 & 9.3 & 28.6 & 32.3 & 30.6 & 33.5\\
 & Right Wing & 19.6 & 22.2 & 20.3 & 23.7 & 11.5 & 15.9 & 25.6 & 30.9 & 29.6 & 32.2\\
 & Stern & 53.3 & 55.6 & 56.7 & 56.9 & 48.8 & 51.0 & 55.3 & 57.8 & 55.7 & 58.6\\
 & Tail & 0.0 & 0.0 & 0.0 & 0.0 & 0.0 & 0.0 & 0.0 & 0.0 & 0.0 & 0.0\\
 & Wheel & 50.4 & 53.1 & 52.9 & 53.4 & 36.0 & 37.5 & 49.9 & 51.6 & 49.7 & 52.4\\
 \midrule
Bicycle & Back Wheel & 63.8 & 65.6 & 63.6 & 66.7 & 56.0 & 60.1 & 67.9 & 70.1 & 68.8 & 71.4\\
 & Chainwheel & 41.2 & 44.5 & 44.9 & 45.6 & 34.2 & 36.9 & 42.0 & 46.5 & 44.6 & 47.5\\
 & Body & 44.6 & 46.1 & 44.9 & 46.8 & 40.9 & 44.8 & 47.3 & 49.7 & 47.3 & 50.9\\
 & Front Wheel & 68.4 & 71.4 & 70.9 & 72.9 & 61.7 & 65.7 & 72.9 & 74.4 & 72.7 & 75.3\\
 & Handlebar & 27.1 & 30.4 & 26.1 & 27.1 & 18.2 & 20.2 & 24.9 & 31.2 & 29.8 & 32.7\\
 & Headlight & 0.0 & 0.0 & 0.0 & 0.0 & 0.0 & 0.0 & 0.0 & 0.0 & 0.0 & 0.0\\
 & Saddle & 41.1 & 42.2 & 41.6 & 42.3 & 21.1 & 23.8 & 43.5 & 44.0 & 42.7 & 45.5\\
 \midrule
Bird & Beak & 53.0 & 56.5 & 57.3 & 57.6 & 37.8 & 40.1 & 49.3 & 57.6 & 55.2 & 58.4\\
 & Head & 66.5 & 65.3 & 66.4 & 67.1 & 55.0 & 56.7 & 66.5 & 69.2 & 67.5 & 70.5\\
 & Left Eye & 26.2 & 30.6 & 27.6 & 32.7 & 17.1 & 23.2 & 57.8 & 58.6 & 56.8 & 59.6\\
 & Left Foot & 5.9 & 8.8 & 12.0 & 12.2 & 0.0 & 1.2 & 9.5 & 11.2 & 9.2 & 12.6\\
 & Left Leg & 5.1 & 9.1 & 9.3 & 10.6 & 0.3 & 2.6 & 15.9 & 14.6 & 12.6 & 15.3\\
 & Left Wing & 4.2 & 8.2 & 11.9 & 13.1 & 8.1 & 11.3 & 29.4 & 32.0 & 30.1 & 33.5\\
 & Neck & 34.0 & 32.5 & 35.8 & 34.7 & 31.8 & 31.7 & 34.4 & 32.7 & 30.7 & 33.3\\
 & Right Eye & 0.0 & 12.3 & 11.6 & 20.2 & 0.9 & 10.5 & 55.2 & 63.1 & 61.8 & 64.6\\
 & Right Foot & 0.0 & 2.7 & 1.2 & 3.8 & 0.0 & 4.6 & 7.4 & 8.2 & 6.7 & 9.6\\
 & Right Leg & 11.1 & 12.6 & 11.1 & 13.5 & 4.6 & 9.0 & 11.2 & 15.7 & 13.6 & 16.5\\
 & Right Wing & 18.7 & 19.2 & 16.3 & 19.9 & 17.8 & 22.4 & 20.3 & 23.0 & 21.8 & 24.6\\
 & Tail & 30.0 & 33.5 & 36.2 & 36.4 & 26.3 & 28.5 & 29.5 & 36.0 & 33.1 & 37.0\\
 & Torso & 60.6 & 61.9 & 65.3 & 66.6 & 61.2 & 63.5 & 61.2 & 64.8 & 62.1 & 65.2\\
 \midrule
Boat & nan & 56.7 & 65.5 & 61.2 & 69.8 & 54.9 & 64.5 & 75.3 & 76.7 & 74.4 & 77.5\\
\midrule
Bottle & Body & 64.6 & 71.7 & 72.5 & 74.7 & 65.5 & 69.7 & 67.6 & 76.3 & 74.3 & 77.4\\
 & Cap & 28.1 & 30.3 & 30.9 & 31.8 & 21.4 & 24.3 & 35.1 & 31.8 & 29.1 & 32.4\\
 \midrule
Bus & Back Plate & 0.0 & 4.1 & 0.0 & 0.0 & 0.0 & 0.0 & 13.0 & 13.0 & 11.3 & 14.5\\
 & Back Side & 49.0 & 40.2 & 44.1 & 33.7 & 47.2 & 37.8 & 43.5 & 23.9 & 21.2 & 25.0\\
 & Door & 40.9 & 45.6 & 46.1 & 47.2 & 31.0 & 33.1 & 38.2 & 45.7 & 43.1 & 46.2\\
 & Front Plate & 26.3 & 33.8 & 42.2 & 46.1 & 0.0 & 5.9 & 45.3 & 53.3 & 51.6 & 54.4\\
 & Front Side & 68.9 & 67.1 & 66.9 & 67.2 & 60.9 & 62.2 & 48.6 & 49.0 & 47.6 & 50.5\\
 & Headlight & 32.6 & 33.5 & 34.8 & 38.5 & 5.9 & 11.6 & 38.8 & 49.2 & 46.1 & 50.1\\
 & Left Mirror & 0.0 & 1.3 & 0.8 & 12.3 & 0.0 & 13.5 & 7.5 & 23.8 & 21.8 & 24.5\\
 & Left Side & 21.4 & 24.9 & 25.1 & 27.9 & 23.2 & 27.0 & 34.6 & 34.0 & 32.8 & 35.6\\
 & Right Mirror & 0.0 & 14.6 & 12.5 & 14.6 & 0.0 & 4.1 & 9.7 & 27.7 & 25.1 & 28.3\\
 & Right Side & 33.9 & 31.7 & 31.5 & 33.7 & 24.2 & 27.4 & 30.0 & 32.4 & 30.1 & 33.8\\
 & Roof & 0.0 & 3.4 & 8.0 & 9.4 & 1.0 & 3.4 & 13.5 & 12.2 & 10.1 & 13.2\\
 & Wheel & 57.1 & 57.2 & 56.2 & 57.8 & 48.8 & 51.4 & 59.3 & 60.4 & 58.2 & 61.8\\
 & Window & 73.5 & 75.9 & 74.8 & 76.8 & 66.6 & 69.6 & 76.5 & 78.6 & 76.0 & 79.3\\
 \midrule
Car & Back Plate & 25.6 & 26.0 & 26.9 & 31.2 & 5.5 & 11.8 & 39.2 & 47.8 & 45.6 & 48.5\\
 & Back Side & 45.0 & 42.1 & 44.5 & 44.5 & 37.0 & 38.0 & 44.6 & 39.7 & 37.7 & 40.3\\
 & Door & 41.4 & 43.5 & 44.1 & 46.6 & 37.2 & 41.7 & 43.6 & 46.6 & 44.4 & 47.3\\
 & Front Plate & 43.0 & 44.2 & 38.1 & 39.1 & 4.7 & 6.7 & 48.6 & 52.0 & 50.7 & 53.6\\
 & Front Side & 66.0 & 65.7 & 65.6 & 66.3 & 60.1 & 62.8 & 56.1 & 56.2 & 54.3 & 57.3\\
 & Headlight & 54.4 & 54.3 & 51.5 & 52.7 & 37.4 & 39.6 & 54.3 & 57.1 & 55.5 & 58.7\\
 & Left Mirror & 12.6 & 15.6 & 14.0 & 16.5 & 0.1 & 3.6 & 21.0 & 34.7 & 32.1 & 35.3\\
 & Left Side & 20.5 & 21.8 & 20.5 & 23.8 & 17.1 & 21.4 & 28.7 & 32.2 & 30.4 & 33.8\\
 & Right Mirror & 0.3 & 9.6 & 7.1 & 11.2 & 0.0 & 5.1 & 17.8 & 31.1 & 29.1 & 32.3\\
 & Right Side & 16.7 & 17.2 & 18.7 & 19.9 & 14.3 & 17.5 & 29.0 & 32.9 & 31.4 & 34.6\\
 & Roof & 17.4 & 22.1 & 27.8 & 27.0 & 11.7 & 11.9 & 22.5 & 27.6 & 25.0 & 28.2\\
 & Wheel & 68.3 & 70.5 & 69.2 & 71.6 & 63.7 & 67.1 & 72.4 & 72.5 & 70.2 & 73.7\\
 & Window & 65.4 & 66.9 & 67.1 & 67.4 & 55.4 & 56.7 & 68.4 & 69.7 & 67.1 & 70.3\\
 \bottomrule

\end{tabular}
}
\end{table}

\clearpage
\begin{table}[h]
\resizebox{\textwidth}{!}{  

\begin{tabular}{p{1cm} p{3cm} p{1.4cm}*{9}{l}}
\toprule
Object & Part & deeplabv3 & deeplabv3+O & BSANet & BSANet+O & GMNet & GMNet+O & FLOAT & FLOAT+O & FLOAT\textsuperscript{\textdagger} & FLOAT\textsuperscript{\textdagger}+O\\
\midrule
          Cat & Head & 75.4 & 76.8 & 76.4 & 77.2 & 72.1 & 73.9 & 77.8 & 78.2 & 76.4 & 79.6\\
 & Left Back Leg & 9.7 & 10.4 & 9.9 & 10.9 & 6.7 & 8.7 & 11.6 & 15.6 & 13.5 & 16.7\\
 & Left Back Paw & 9.3 & 11.6 & 10.8 & 14.4 & 4.0 & 9.6 & 10.7 & 19.8 & 17.1 & 20.3\\
 & Left Ear & 12.6 & 15.1 & 22.3 & 26.6 & 21.7 & 28.0 & 59.7 & 61.5 & 59.3 & 62.5\\
 & Left Eye & 7.9 & 9.7 & 11.6 & 17.9 & 8.5 & 16.8 & 59.2 & 66.6 & 64.0 & 67.2\\
 & Left Front Leg & 11.5 & 14.5 & 15.0 & 20.2 & 15.8 & 23.0 & 25.3 & 28.7 & 26.2 & 29.4\\
 & Left Front Paw & 15.3 & 16.2 & 16.7 & 17.5 & 13.6 & 15.4 & 19.0 & 24.6 & 21.2 & 25.1\\
 & Neck & 21.5 & 22.3 & 18.5 & 20.4 & 19.6 & 23.5 & 23.3 & 23.2 & 20.9 & 24.1\\
 & Nose & 39.7 & 42.8 & 46.3 & 46.8 & 32.9 & 35.4 & 43.1 & 46.4 & 44.0 & 47.2\\
 & Right Back Leg & 1.5 & 6.4 & 10.1 & 14.7 & 4.9 & 11.5 & 16.0 & 17.5 & 15.3 & 18.8\\
 & Right Back Paw & 0.2 & 2.6 & 7.5 & 9.3 & 0.6 & 3.4 & 16.1 & 19.0 & 17.5 & 20.7\\
 & Right Ear & 33.2 & 36.9 & 28.6 & 34.1 & 26.4 & 32.9 & 59.7 & 61.5 & 59.2 & 62.4\\
 & Right Eye & 34.0 & 32.1 & 33.8 & 36.2 & 23.4 & 26.8 & 62.8 & 68.7 & 66.0 & 69.2\\
 & Right Front Leg & 16.2 & 16.3 & 12.1 & 16.6 & 12.0 & 17.5 & 26.5 & 29.0 & 27.4 & 30.6\\
 & Right Front Paw & 17.6 & 18.8 & 12.1 & 15.7 & 12.1 & 17.7 & 21.9 & 28.4 & 26.6 & 29.8\\
 & Tail & 40.6 & 41.2 & 45.4 & 46.3 & 35.3 & 38.2 & 43.0 & 47.7 & 45.2 & 48.4\\
 & Torso & 65.6 & 65.7 & 66.7 & 67.1 & 64.9 & 67.3 & 67.6 & 68.0 & 66.1 & 69.3\\
 \midrule
Chair & nan & 35.6 & 42.1 & 35.4 & 39.8 & 35.5 & 41.9 & 59.6 & 59.7 & 57.3 & 60.5\\
\midrule
Cow & Head & 60.1 & 61.2 & 60.6 & 61.4 & 54.7 & 56.5 & 61.1 & 63.7 & 61.0 & 64.2\\
 & Left Back Lower Leg & 0.8 & 2.4 & 3.3 & 6.6 & 0.5 & 5.8 & 15.3 & 18.4 & 16.2 & 19.4\\
 & Left Back Upper Leg & 13.5 & 14.5 & 16.2 & 17.7 & 14.0 & 16.5 & 19.6 & 25.9 & 24.3 & 27.5\\
 & Left Ear & 1.9 & 13.7 & 24.2 & 28.9 & 8.2 & 13.9 & 53.0 & 57.8 & 55.3 & 58.5\\
 & Left Eye & 0.0 & 0.0 & 0.0 & 0.0 & 0.0 & 0.0 & 41.1 & 40.2 & 38.5 & 41.7\\
 & Left Front Lower Leg & 15.9 & 16.8 & 14.5 & 18.2 & 10.7 & 16.4 & 25.4 & 26.7 & 24.2 & 27.4\\
 & Left Front Upper Leg & 14.4 & 18.2 & 18.6 & 22.4 & 9.6 & 15.4 & 33.2 & 39.7 & 37.5 & 40.6\\
 & Left Horn & 0.0 & 5.9 & 13.3 & 18.6 & 0.0 & 6.3 & 28.3 & 33.2 & 31.4 & 34.6\\
 & Muzzle & 71.0 & 70.3 & 72.1 & 71.1 & 64.7 & 65.7 & 70.4 & 72.4 & 70.7 & 73.7\\
 & Neck & 5.7 & 9.1 & 15.1 & 16.7 & 9.0 & 11.6 & 21.9 & 19.8 & 17.0 & 20.2\\
 & Right Back Lower Leg & 16.3 & 17.6 & 18.4 & 18.9 & 1.7 & 4.2 & 17.4 & 19.8 & 17.5 & 20.6\\
 & Right Back Upper Leg & 5.6 & 9.4 & 12.9 & 14.3 & 5.1 & 8.5 & 22.9 & 27.9 & 25.2 & 29.1\\
 & Right Ear & 27.4 & 26.0 & 28.8 & 37.2 & 24.0 & 34.4 & 56.5 & 60.4 & 58.6 & 61.4\\
 & Right Eye & 1.9 & 8.5 & 11.0 & 16.2 & 0.0 & 6.2 & 38.5 & 37.2 & 34.4 & 38.1\\
 & Right Front Lower Leg & 2.6 & 7.2 & 9.5 & 12.5 & 0.5 & 5.5 & 25.1 & 28.4 & 26.7 & 29.8\\
 & Right Front Upper Leg & 19.2 & 20.1 & 21.8 & 23.8 & 14.4 & 18.4 & 32.7 & 37.2 & 35.1 & 38.3\\
 & Right Horn & 0.0 & 9.6 & 30.7 & 30.2 & 2.4 & 2.9 & 24.1 & 24.8 & 22.3 & 25.5\\
 & Tail & 5.6 & 7.6 & 12.2 & 16.1 & 0.0 & 4.9 & 17.8 & 31.0 & 28.8 & 32.0\\
 & Torso & 70.0 & 71.2 & 73.0 & 73.7 & 64.6 & 67.3 & 70.8 & 73.9 & 72.5 & 75.7\\
 \midrule
Dining Table & nan & 38.6 & 46.9 & 43.6 & 49.2 & 40.3 & 47.9 & 58.2 & 58.3 & 56.6 & 59.6\\
\midrule
Dog & Head & 61.7 & 62.4 & 63.4 & 63.9 & 58.9 & 61.4 & 62.7 & 62.8 & 60.1 & 63.3\\
 & Left Back Leg & 5.0 & 8.2 & 8.6 & 11.6 & 3.2 & 8.2 & 17.1 & 23.1 & 21.4 & 24.6\\
 & Left Back Paw & 6.8 & 8.5 & 6.7 & 11.3 & 0.9 & 7.5 & 12.6 & 17.0 & 15.3 & 18.5\\
 & Left Ear & 22.1 & 22.7 & 19.5 & 25.4 & 18.9 & 26.8 & 56.6 & 55.4 & 53.6 & 56.7\\
 & Left Eye & 21.0 & 25.1 & 21.4 & 26.7 & 12.4 & 19.7 & 54.9 & 59.1 & 57.0 & 60.2\\
 & Left Front Leg & 9.4 & 18.8 & 18.6 & 22.1 & 13.5 & 19.0 & 32.3 & 34.4 & 32.5 & 35.7\\
 & Left Front Paw & 8.2 & 13.3 & 16.6 & 18.9 & 16.0 & 19.3 & 30.2 & 33.7 & 31.0 & 34.2\\
 & Muzzle & 67.7 & 69.2 & 70.3 & 70.1 & 63.2 & 65.0 & 64.9 & 66.1 & 64.3 & 67.7\\
 & Neck & 27.5 & 26.5 & 27.4 & 29.2 & 21.7 & 24.5 & 25.6 & 24.9 & 23.0 & 26.2\\
 & Nose & 64.0 & 67.2 & 69.2 & 68.5 & 55.9 & 57.2 & 66.0 & 67.4 & 65.2 & 68.7\\
 & Right Back Leg & 18.0 & 18.1 & 12.3 & 15.8 & 12.8 & 18.3 & 24.4 & 26.6 & 24.1 & 27.3\\
 & Right Back Paw & 2.5 & 4.4 & 5.5 & 9.2 & 0.5 & 6.2 & 18.1 & 19.2 & 17.3 & 20.7\\
 & Right Ear & 20.6 & 26.7 & 23.5 & 28.3 & 26.3 & 33.1 & 53.0 & 52.1 & 50.0 & 53.2\\
 & Right Eye & 20.7 & 21.7 & 17.5 & 22.9 & 17.3 & 23.7 & 59.8 & 62.2 & 60.2 & 63.7\\
 & Right Front Leg & 20.6 & 21.4 & 14.1 & 18.1 & 17.3 & 22.3 & 33.4 & 35.3 & 33.3 & 36.8\\
 & Right Front Paw & 23.5 & 22.8 & 18.7 & 20.6 & 17.4 & 20.3 & 31.8 & 34.1 & 32.3 & 35.5\\
 & Tail & 31.8 & 32.1 & 35.5 & 36.4 & 27.5 & 30.4 & 33.5 & 39.2 & 37.6 & 40.8\\
 & Torso & 60.3 & 60.4 & 62.3 & 62.9 & 58.8 & 60.4 & 62.0 & 62.9 & 61.0 & 64.2\\

\bottomrule
\end{tabular}
}
\end{table}

\clearpage
\begin{table}[h]
\resizebox{\textwidth}{!}{  

\begin{tabular}{p{1.3cm} p{3cm} p{1.4cm}*{9}{l}}
\toprule
Object & Part & deeplabv3 & deeplabv3+O & BSANet & BSANet+O & GMNet & GMNet+O & FLOAT & FLOAT+O & FLOAT\textsuperscript{\textdagger} & FLOAT\textsuperscript{\textdagger}+O\\
\midrule
Horse & Head & 58.0 & 58.7 & 58.2 & 59.3 & 49.0 & 51.1 & 63.5 & 62.7 & 60.4 & 63.6\\
 & Left Back Hoof & 0.0 & 0.0 & 1.7 & 2.9 & 0.0 & 3.2 & 8.2 & 10.9 & 9.4 & 12.6\\
 & Left Back Lower Leg & 4.4 & 6.2 & 8.0 & 10.7 & 1.6 & 6.3 & 19.6 & 22.3 & 20.6 & 23.8\\
 & Left Back Upper Leg & 0.4 & 7.5 & 10.0 & 17.2 & 13.0 & 22.2 & 24.0 & 36.0 & 34.4 & 37.6\\
 & Left Ear & 7.0 & 15.2 & 13.5 & 19.1 & 8.7 & 15.3 & 50.1 & 47.5 & 45.1 & 48.3\\
 & Left Eye & 0.0 & 0.0 & 0.0 & 0.0 & 0.0 & 0.0 & 39.8 & 37.3 & 35.5 & 38.7\\
 & Left Front Hoof & 0.0 & 3.2 & 3.9 & 4.8 & 0.0 & 2.9 & 2.1 & 8.5 & 6.1 & 9.3\\
 & Left Front Lower Leg & 15.5 & 19.6 & 20.1 & 22.6 & 10.9 & 14.4 & 23.3 & 26.3 & 24.5 & 27.7\\
 & Left Front Upper Leg & 14.2 & 23.1 & 24.9 & 26.3 & 14.3 & 16.7 & 30.1 & 34.8 & 32.3 & 35.5\\
 & Muzzle & 65.0 & 65.8 & 66.4 & 67.1 & 56.0 & 58.7 & 69.7 & 69.4 & 67.6 & 70.8\\
 & Neck & 50.8 & 50.9 & 48.9 & 50.9 & 38.9 & 42.9 & 51.5 & 53.9 & 52.6 & 55.8\\
 & Right Back Hoof & 0.0 & 1.2 & 2.6 & 2.9 & 0.0 & 2.3 & 7.7 & 12.8 & 10.1 & 13.3\\
 & Right Back Lower Leg & 16.1 & 18.7 & 19.6 & 20.8 & 5.8 & 9.0 & 21.2 & 22.6 & 20.0 & 23.2\\
 & Right Back Upper Leg & 22.4 & 24.6 & 23.9 & 25.4 & 12.9 & 16.4 & 28.6 & 31.9 & 30.3 & 33.5\\
 & Right Ear & 29.3 & 29.1 & 25.7 & 29.5 & 26.8 & 32.6 & 49.7 & 51.6 & 49.5 & 52.7\\
 & Right Eye & 17.2 & 17.9 & 19.4 & 20.1 & 0.8 & 2.5 & 52.1 & 25.6 & 23.3 & 26.8\\
 & Right Front Hoof & 0.0 & 1.3 & 2.2 & 2.6 & 0.0 & 2.4 & 2.9 & 4.1 & 2.1 & 5.3\\
 & Right Front Lower Leg & 4.1 & 8.2 & 9.6 & 11.3 & 4.6 & 7.3 & 21.7 & 25.2 & 23.6 & 26.8\\
 & Right Front Upper Leg & 21.5 & 12.4 & 12.6 & 14.2 & 13.2 & 16.8 & 33.3 & 36.1 & 34.2 & 37.7\\
 & Tail & 47.9 & 48.7 & 49.9 & 50.7 & 39.0 & 41.8 & 49.6 & 53.4 & 51.1 & 54.3\\
 & Torso & 61.3 & 62.1 & 61.7 & 63.8 & 56.4 & 60.5 & 65.1 & 66.8 & 64.2 & 67.4\\
 \midrule
Motorbike & Back Wheel & 60.7 & 64.5 & 63.9 & 65.6 & 52.3 & 55.0 & 63.7 & 67.1 & 65.5 & 68.7\\
 & Body & 67.8 & 68.0 & 70.7 & 71.4 & 64.5 & 67.2 & 70.8 & 72.6 & 70.0 & 73.2\\
 & Front Wheel & 68.9 & 71.3 & 72.2 & 73.2 & 62.9 & 65.9 & 71.9 & 75.6 & 73.2 & 76.4\\
 & Handlebar & 0.0 & 0.0 & 0.0 & 0.0 & 0.0 & 0.0 & 0.1 & 0.0 & 0.0 & 0.0\\
 & Headlight & 30.7 & 32.1 & 17.8 & 22.9 & 9.6 & 15.7 & 34.7 & 38.7 & 36.4 & 39.6\\
 & Saddle & 0.0 & 0.0 & 0.0 & 0.0 & 0.0 & 0.0 & 0.0 & 0.0 & 0.0 & 0.0\\
 \midrule
Person & Hair & 73.8 & 73.9 & 74.0 & 74.2 & 68.3 & 69.5 & 72.7 & 72.4 & 70.0 & 73.2\\
 & Head & 70.0 & 70.2 & 70.8 & 71.1 & 63.9 & 66.2 & 71.2 & 71.7 & 69.2 & 72.4\\
 & Left Ear & 19.5 & 19.7 & 16.6 & 19.3 & 6.2 & 9.9 & 45.1 & 50.6 & 48.0 & 51.2\\
 & Left Eye & 4.5 & 16.3 & 12.9 & 17.5 & 3.1 & 9.7 & 42.7 & 47.0 & 45.5 & 48.7\\
 & Left Eyebrow & 0.0 & 0.0 & 3.6 & 9.8 & 0.1 & 8.3 & 17.1 & 29.2 & 27.1 & 30.3\\
 & Left Foot & 18.4 & 19.3 & 16.2 & 18.4 & 11.7 & 14.9 & 17.8 & 19.4 & 17.2 & 20.4\\
 & Left Hand & 8.4 & 13.5 & 15.7 & 18.6 & 10.0 & 14.9 & 33.7 & 36.7 & 34.4 & 37.6\\
 & Left Lower Arm & 19.6 & 20.6 & 18.2 & 22.7 & 10.9 & 16.4 & 37.4 & 40.9 & 39.1 & 42.3\\
 & Left Lower Leg & 16.2 & 19.4 & 18.6 & 20.2 & 16.6 & 20.2 & 23.5 & 27.2 & 25.5 & 28.7\\
 & Left Upper Arm & 21.0 & 21.8 & 19.0 & 22.8 & 16.1 & 21.9 & 47.2 & 51.1 & 49.6 & 52.8\\
 & Left Upper Leg & 15.9 & 16.2 & 10.7 & 18.5 & 13.0 & 21.8 & 30.8 & 35.6 & 33.2 & 36.4\\
 & Mouth & 52.2 & 54.6 & 53.6 & 57.1 & 30.2 & 35.7 & 57.9 & 61.3 & 59.5 & 62.7\\
 & Neck & 51.3 & 50.5 & 52.2 & 53.3 & 42.6 & 45.7 & 52.9 & 52.4 & 50.7 & 53.9\\
 & Nose & 59.2 & 60.2 & 58.2 & 59.6 & 40.2 & 42.6 & 61.1 & 64.7 & 62.0 & 65.2\\
 & Right Ear & 16.3 & 18.8 & 19.8 & 27.0 & 9.8 & 18.0 & 47.8 & 53.3 & 51.2 & 54.4\\
 & Right Eye & 22.8 & 21.3 & 17.9 & 19.2 & 12.8 & 15.1 & 47.3 & 51.6 & 49.3 & 52.5\\
 & Right Eyebrow & 0.0 & 1.4 & 3.6 & 7.6 & 0.4 & 5.4 & 12.2 & 28.2 & 26.4 & 29.6\\
 & Right Foot & 9.3 & 12.6 & 12.4 & 14.9 & 6.1 & 9.6 & 18.4 & 19.1 & 17.1 & 20.3\\
 & Right Hand & 28.7 & 30.4 & 24.0 & 32.8 & 22.9 & 33.7 & 36.0 & 38.0 & 36.7 & 39.9\\
 & Right Lower Arm & 17.9 & 22.9 & 19.6 & 26.3 & 18.5 & 26.2 & 37.2 & 40.4 & 38.1 & 41.3\\
 & Right Lower Leg & 15.6 & 17.0 & 16.6 & 19.4 & 9.3 & 14.1 & 24.8 & 27.1 & 25.3 & 28.5\\
 & Right Upper Arm & 22.0 & 24.1 & 21.4 & 25.1 & 22.1 & 26.8 & 45.9 & 49.9 & 48.1 & 51.3\\
 & Right Upper Leg & 19.7 & 23.5 & 23.8 & 25.0 & 19.7 & 22.9 & 32.9 & 35.5 & 33.4 & 36.6\\
 & Torso & 64.2 & 64.6 & 65.3 & 66.2 & 58.7 & 61.6 & 65.3 & 67.0 & 65.1 & 68.3\\
 \midrule
Potted Plant & Plant & 51.6 & 55.7 & 56.2 & 60.2 & 48.2 & 53.2 & 54.5 & 63.2 & 61.3 & 64.5\\
 & Pot & 50.0 & 56.4 & 53.3 & 58.7 & 40.0 & 46.4 & 49.9 & 58.7 & 56.0 & 59.2\\
 \midrule
Sheep & Head & 49.2 & 50.2 & 45.1 & 49.3 & 42.5 & 47.7 & 47.6 & 53.0 & 51.6 & 54.8\\
 & Left Back Lower Leg & 2.3 & 2.6 & 3.6 & 3.7 & 0.3 & 1.4 & 7.7 & 3.9 & 1.9 & 5.1\\
 & Left Back Upper Leg & 0.0 & 0.0 & 0.0 & 0.0 & 0.0 & 2.0 & 9.7 & 8.9 & 7.0 & 10.2\\
 & Left Ear & 18.3 & 30.7 & 27.2 & 33.6 & 3.4 & 11.8 & 51.9 & 50.0 & 48.2 & 51.4\\
 & Left Eye & 0.0 & 0.0 & 0.0 & 12.1 & 0.0 & 14.1 & 37.6 & 42.8 & 40.1 & 43.3\\
 & Left Front Lower Leg & 0.0 & 0.0 & 0.0 & 7.5 & 0.0 & 8.5 & 15.7 & 17.6 & 15.4 & 18.6\\
 & Left Front Upper Leg & 0.0 & 0.0 & 0.0 & 5.8 & 0.0 & 6.8 & 14.4 & 15.8 & 13.2 & 16.4\\
 & Left Horn & 0.0 & 4.8 & 0.7 & 6.2 & 0.0 & 6.5 & 26.1 & 47.1 & 45.4 & 48.6\\
 & Muzzle & 59.8 & 60.4 & 61.1 & 62.4 & 58.6 & 60.9 & 66.0 & 66.1 & 64.5 & 67.7\\
 & Neck & 24.9 & 25.3 & 23.1 & 27.3 & 17.1 & 22.3 & 32.0 & 33.1 & 31.3 & 34.5\\
 & Right Back Lower Leg & 0.0 & 1.2 & 1.3 & 2.4 & 0.0 & 2.1 & 4.9 & 9.5 & 7.6 & 10.8\\
 & Right Back Upper Leg & 0.2 & 0.9 & 1.6 & 4.6 & 0.0 & 4.0 & 7.1 & 7.3 & 5.2 & 8.4\\
 & Right Ear & 15.5 & 16.5 & 12.9 & 18.5 & 24.1 & 31.7 & 49.5 & 44.2 & 42.3 & 45.5\\
 & Right Eye & 8.5 & 12.8 & 15.8 & 17.2 & 1.2 & 3.6 & 37.0 & 38.6 & 36.5 & 39.7\\
 & Right Front Lower Leg & 0.4 & 1.6 & 0.0 & 5.4 & 0.0 & 6.4 & 12.7 & 15.5 & 13.1 & 16.3\\
 & Right Front Upper Leg & 2.3 & 4.2 & 0.0 & 6.1 & 2.2 & 9.3 & 16.3 & 13.4 & 11.5 & 14.7\\
 & Right Horn & 0.0 & 5.6 & 8.4 & 16.8 & 0.0 & 10.4 & 25.0 & 43.3 & 41.0 & 44.2\\
 & Tail & 6.8 & 10.2 & 5.1 & 8.3 & 0.1 & 5.3 & 12.3 & 17.6 & 15.3 & 18.5\\
 & Torso & 65.1 & 67.6 & 65.5 & 68.6 & 62.4 & 66.5 & 69.2 & 71.0 & 69.5 & 72.7\\

\bottomrule
\end{tabular}
}
\end{table}

\clearpage
\begin{table}[h]
\resizebox{\textwidth}{!}{  

\begin{tabular}{p{1.3cm} p{3cm} p{1.4cm}*{9}{l}}
\toprule
Object & Part & deeplabv3 & deeplabv3+O & BSANet & BSANet+O & GMNet & GMNet+O & FLOAT & FLOAT+O & FLOAT\textsuperscript{\textdagger} & FLOAT\textsuperscript{\textdagger}+O\\
\midrule
Sofa &  & 42.1 & 46.8 & 40.4 & 49.9 & 43.3 & 54.8 & 69.0 & 69.1 & 67.2 & 70.4\\
\midrule
Train & Coach Back Side & 0.0 & 1.7 & 3.8 & 1.1 & 5.9 & 4.2 & 11.1 & 0.1 & 0.0 & 0.3\\
 & Coach Front Side & 0.0 & 0.1 & 0.0 & 0.6 & 0.0 & 0.5 & 0.5 & 0.8 & 0.0 & 1.9\\
 & Coach Left Side & 5.9 & 6.1 & 6.5 & 7.2 & 5.9 & 7.6 & 6.1 & 4.2 & 2.4 & 5.6\\
 & Coach Right Side & 3.4 & 7.2 & 10.2 & 11.3 & 4.8 & 7.9 & 9.1 & 2.9 & 1.4 & 4.6\\
 & Coach Roof & 0.0 & 5.3 & 9.1 & 15.7 & 1.3 & 9.9 & 0.0 & 17.1 & 15.5 & 18.7\\
 & Coach & 30.7 & 33.7 & 35.2 & 35.7 & 36.6 & 39.1 & 28.0 & 36.8 & 32.6 & 37.8\\
 & Head & 4.3 & 4.8 & 9.0 & 9.4 & 5.7 & 8.1 & 4.4 & 5.9 & 4.1 & 7.3\\
 & Head Back Side & 0.0 & 0.2 & 0.0 & 0.2 & 0.0 & 2.2 & 1.3 & 0.0 & 0.0 & 0.0\\
 & Head Front Side & 71.0 & 71.1 & 72.6 & 71.1 & 62.2 & 62.7 & 34.5 & 41.5 & 39.2 & 42.4\\
 & Head Left Side & 19.3 & 20.5 & 16.0 & 18.8 & 22.8 & 27.6 & 27.2 & 22.4 & 20.5 & 23.7\\
 & Head Right Side & 14.3 & 20.5 & 19.8 & 20.5 & 18.4 & 21.1 & 22.2 & 18.9 & 17.1 & 20.3\\
 & Head Roof & 18.7 & 23.7 & 25.2 & 26.4 & 0.9 & 3.1 & 22.2 & 22.9 & 21.5 & 24.7\\
 & Headlight & 23.1 & 29.3 & 24.1 & 29.2 & 5.1 & 12.2 & 29.5 & 39.7 & 37.1 & 40.3\\
 \midrule
TV Monitor & Frame & 46.8 & 50.2 & 47.2 & 52.3 & 40.3 & 47.4 & 44.7 & 56.4 & 54.5 & 57.7\\
 & Screen & 68.5 & 80.1 & 71.5 & 79.2 & 63.9 & 73.6 & 67.4 & 82.2 & 80.1 & 83.3\\ 

\bottomrule
\end{tabular}
}
\end{table}

\clearpage

\subsection{Pascal-Part-201 | $miou_{small}$}

\begin{table}[h]
\begin{tabular}{llllllllll}
\toprule
Object & FLOAT & FLOAT+O \\
\midrule
background & - & -\\
aeroplane & 28.3 & 29.3 \\
    bicycle & 20.6 & 26.6 \\
       bird & 40.1 & 44.1 \\
       boat & 15.7 & 18.7 \\
     bottle & 36.2 & 39.2 \\
        bus & 38.4 & 41.4 \\
        car & 32.8 & 37.8 \\
        cat & 37.3 & 42.3 \\
      chair & 38.5 & 41.5 \\
        cow & 22.6 & 25.6 \\
diningtable & 30.1 & 35.1 \\
        dog & 28.1 & 33.1 \\
      horse & 39.4 & 40.4 \\
  motorbike & 24.2 & 30.2 \\
     person & 37.2 & 38.2 \\
pottedplant & 47.4 & 50.4 \\
      sheep & 43.2 & 46.2 \\
       sofa & 45.2 & 46.2 \\
      train & 16.3 & 17.3 \\
  tvmonitor & 47.3 & 51.3 \\
  
\bottomrule
\end{tabular}
\end{table}
\clearpage
\subsection{PartImageNet | mIoU}

\begin{table}[h]
\begin{tabular}{lllll}
\toprule
Class Id & Deeplabv3+ & Deeplabv3++O & Segformer & Segformer+O\\
\midrule
0 &  5.0 &  3.0 &  7.0 & 14.1 \\
 1 & 41.7 & 42.3 & 44.2 & 45.7 \\
 2 & 39.0 & 40.5 & 41.2 & 43.7 \\
 3 & 27.2 & 33.3 & 32.1 & 43.1 \\
 4 & 59.7 & 57.1 & 56.3 & 62.4 \\
 5 & 56.1 & 50.8 & 50.3 & 60.7 \\
 6 & 45.1 & 38.4 & 44.2 & 49.1 \\
 7 & 39.5 & 36.8 & 35.1 & 49.8 \\
 8 & 55.2 & 45.3 & 40.2 & 56.9 \\
 9 & 65.6 & 62.1 & 65.2 & 67.1 \\
10 & 69.4 & 68.5 & 64.7 & 70.3 \\
11 & 54.2 & 55.9 & 54.8 & 56.2 \\
12 & 48.7 & 57.2 & 55.7 & 58.5 \\
13 & 72.4 & 78.4 & 77.5 & 79.2 \\
14 & 72.3 & 79.4 & 76.9 & 77.1 \\
15 & 53.4 & 62.2 & 60.1 & 64.6 \\
16 & 53.8 & 62.9 & 60.8 & 63.7 \\
17 & 59.4 & 64.9 & 65.9 & 66.0 \\
18 & 65.5 & 74.6 & 76.1 & 77.4 \\
19 & 89.6 & 93.3 & 90.5 & 94.2 \\
20 & 73.7 & 69.2 & 65.6 & 77.2 \\
21 & 71.9 & 74.0 & 75.8 & 78.9 \\
22 & 62.8 & 63.8 & 64.2 & 65.8 \\
23 & 54.2 & 61.5 & 63.4 & 64.6 \\
24 & 95.2 & 96.9 & 96.1 & 96.2 \\
25 & 78.2 & 87.4 & 87.9 & 90.3 \\
26 & 22.6 & 39.0 & 35.8 & 43.2 \\
\bottomrule
\end{tabular}
\end{table}

\clearpage

\begin{table}[h]
\begin{tabular}{lllll}
\toprule
Class Id & Deeplabv3+ & Deeplabv3++O & Segformer & Segformer+O\\
\midrule
27 & 53.2 & 71.0 & 71.4 & 72.6 \\
28 & 41.6 & 53.7 & 53.9 & 55.2 \\
29 & 29.0 & 59.2 & 49.9 & 61.2 \\
30 & 78.5 & 89.0 & 81.9 & 90.1 \\
31 & 91.0 & 92.9 & 90.2 & 92.7 \\
32 & 92.6 & 95.7 & 90.8 & 94.3 \\
33 & 42.5 & 44.5 & 45.2 & 48.3 \\
34 & 57.7 & 67.1 & 64.6 & 65.2 \\
35 & 32.3 & 42.9 & 40.2 & 45.1 \\
36 & 31.0 & 25.4 & 25.1 & 38.9 \\
37 & 48.4 & 40.8 & 40.9 & 50.6 \\
38 & 52.5 & 60.7 & 62.1 & 64.8 \\
39 & 78.9 & 90.1 & 79.3 & 90.3 \\
40 & 98.1 & 98.8 & 98.1 & 98.2 \\
\bottomrule
\end{tabular}
\end{table}

\bibliographystyle{splncs04}
\bibliography{main}